\pdfoutput=1

\documentclass[11pt]{article}

\usepackage{ACL2023}

\newcommand{\reproducibilityNote}[1]{\iffalse #1\fi}

\usepackage{times}
\usepackage{latexsym}

\usepackage[T1]{fontenc}

\usepackage[utf8]{inputenc}

\usepackage{microtype}

\usepackage{inconsolata}

\usepackage{subcaption}

\usepackage{tablefootnote}

\usepackage{paralist}
\usepackage{graphicx}
\usepackage{amsmath}
\usepackage{multirow}
\usepackage{subcaption}
\usepackage{tikz}
\usetikzlibrary{tikzmark}
\usepackage{tipa}
\usepackage{float}
\usepackage{array}
\usepackage{dsfont} 
\usepackage{placeins}
\usepackage{hyperref}

\definecolor{dcyan}{HTML}{00bfbf}
\definecolor{dmagenta}{HTML}{bf00bf}
\definecolor{dyellow}{HTML}{bfbf00}
\definecolor{dgreen}{HTML}{008000}
\definecolor{dblue}{HTML}{0000ff}
\definecolor{dred}{HTML}{ff0000}
\definecolor{dblack}{HTML}{000000}

\definecolor{orange}{RGB}{255, 180, 0}
\definecolor{darkgreen}{RGB}{0, 180, 0}
\definecolor{dlgreen}{RGB}{0, 100, 0}
\definecolor{darkred}{RGB}{180, 0, 0}
\definecolor{lightgreen}{RGB}{100, 200, 100}
\definecolor{lightblue}{RGB}{100, 150, 255}
\definecolor{lightred}{RGB}{255, 125, 125}
\definecolor{darkorange}{RGB}{255, 150, 0}
\definecolor{lightgray}{RGB}{200, 200, 200}
\definecolor{gray}{RGB}{150, 150, 150}
\definecolor{darkgray}{RGB}{100, 100, 100}
\definecolor{braun}{RGB}{120, 100, 80}
\definecolor{lbraun}{RGB}{50, 50, 50}

\definecolor{torange}{RGB}{200, 0, 200}
\definecolor{tpurple}{RGB}{200, 0, 200}
\definecolor{tlightpurple}{RGB}{100, 40, 100}
\definecolor{tblue}{RGB}{100, 150, 255}
\definecolor{tgreen}{RGB}{0, 180, 0}
\definecolor{tred}{RGB}{220, 0, 0}
\definecolor{tyellow}{RGB}{180, 180, 0}

\newcommand{\tgray}[1]{\textcolor{gray}{\textbf{#1}}}

\newcommand{\tbl}[1]{\textcolor{blue}{\textbf{#1}}}
\newcommand{\tred}[1]{\textcolor{red}{\textbf{#1}}}

\newcommand{\tgr}[1]{\textcolor{darkgreen}{\textbf{#1}}}

\newcommand{\sred}[1]{\textcolor{darkorange}{\textbf{#1}}}
\newcommand{\scyan}[1]{\textcolor{cyan}{\textbf{#1}}}

\newcommand{\fong}[1]{{\footnotesize\textcolor{gray}{#1}}}

\newcommand{\tdg}[1]{\textcolor{darkgray}{#1}}

\DeclareFontFamily{U}{rcjhbltx}{}
\DeclareFontShape{U}{rcjhbltx}{m}{n}{<->rcjhbltx}{}
\DeclareSymbolFont{hebrewletters}{U}{rcjhbltx}{m}{n}
\DeclareMathSymbol{\lamed}{\mathord}{hebrewletters}{108}
\DeclareMathSymbol{\samekh}{\mathord}{hebrewletters}{115}

\DeclareMathSymbol{\tav}{\mathord}{hebrewletters}{116}
\DeclareMathSymbol{\tsadi}{\mathord}{hebrewletters}{118}
\DeclareMathSymbol{\vav}{\mathord}{hebrewletters}{119}
\DeclareMathSymbol{\chet}{\mathord}{hebrewletters}{120}

\DeclareMathSymbol{\yod}{\mathord}{hebrewletters}{121}
\DeclareMathSymbol{\zayin}{\mathord}{hebrewletters}{122}
\DeclareMathSymbol{\kaf}{\mathord}{hebrewletters}{107}
\DeclareMathSymbol{\mem}{\mathord}{hebrewletters}{109}
\DeclareMathSymbol{\dalet}{\mathord}{hebrewletters}{100}
\DeclareMathSymbol{\nun}{\mathord}{hebrewletters}{103}
\DeclareMathSymbol{\he}{\mathord}{hebrewletters}{104}
\DeclareMathSymbol{\nun}{\mathord}{hebrewletters}{110}

\DeclareMathSymbol{\pe}{\mathord}{hebrewletters}{112}
\DeclareMathSymbol{\qof}{\mathord}{hebrewletters}{113}
\DeclareMathSymbol{\resh}{\mathord}{hebrewletters}{114}

\DeclareMathSymbol{\ayin}{\mathord}{hebrewletters}{96}
\DeclareMathSymbol{\bet}{\mathord}{hebrewletters}{98}
\DeclareMathSymbol{\tsadisofit}{\mathord}{hebrewletters}{90}

\DeclareMathSymbol{\tet}{\mathord}{hebrewletters}{84}
\DeclareMathSymbol{\pesofit}{\mathord}{hebrewletters}{80}

\DeclareMathSymbol{\kafsofit}{\mathord}{hebrewletters}{75}
\DeclareMathSymbol{\memsofit}{\mathord}{hebrewletters}{77}
\DeclareMathSymbol{\nunsofit}{\mathord}{hebrewletters}{78}

\DeclareMathSymbol{\alef}{\mathord}{hebrewletters}{39}
\DeclareMathSymbol{\gimel}{\mathord}{hebrewletters}{103}
\DeclareMathSymbol{\shin}{\mathord}{hebrewletters}{152}

\newcommand{\sref}[1]{\hyperref[#1]{Section~\ref{#1}}}
\newcommand{\cref}[1]{\hyperref[#1]{Chapter~\ref{#1}}}
\newcommand{\aref}[1]{\hyperref[#1]{Appendix~\ref{#1}}}
\newcommand{\snref}[1]{\hyperref[#1]{Section~\ref{#1}:~\nameref{#1}}}
\newcommand{\anref}[1]{\hyperref[#1]{Appendix~\ref{#1}:~\nameref{#1}}}
\newcommand{\cnref}[1]{\hyperref[#1]{Chapter~\ref{#1}:~\nameref{#1}}}
\newcommand{\fnref}[1]{\hyperref[#1]{Figure~\ref{#1}:~\nameref{#1}}}
\newcommand{\tnref}[1]{\hyperref[#1]{Table~\ref{#1}:~\nameref{#1}}}
\newcommand{\asecref}[1]{\hyperref[#1]{\ref{#1}: \nameref{#1}}}

\title{Systematicity between Forms and Meanings across Languages Supports Efficient Communication}

 \author{Doreen Osmelak\textsuperscript{1} \and Yang Xu\textsuperscript{2} \and Michael Hahn\textsuperscript{1} \and Kate McCurdy\textsuperscript{1} \\
         \textsuperscript{1}Saarland University, Saarland Informatics Campus, Germany \\ 
         \textsuperscript{2}Department of Computer Science, Cognitive Science Program, University of Toronto, Canada\\
         \texttt{\{dosmelak,mhahn,kmccurdy\}@lst.uni-saarland.de, yangxu@cs.toronto.edu}\\}

\begin{document}
\maketitle

\begin{abstract}

Languages vary widely in how meanings map to word forms. These mappings have been found to support efficient communication; however, this theory does not account for systematic relations \emph{within} word forms.
We examine how a restricted set of grammatical meanings (e.g. person, number) are expressed on verbs and pronouns across typologically diverse languages. Consistent with prior work, we find that verb and pronoun forms are shaped by competing communicative pressures for \textit{simplicity} (minimizing the inventory of grammatical distinctions) and \textit{accuracy} (enabling recovery of intended meanings). Crucially, our proposed model uses a novel measure of complexity (inverse of simplicity) based on the \textit{learnability} of meaning-to-form mappings. This innovation captures fine-grained regularities in linguistic form, allowing better discrimination between attested and unattested systems, and establishes a new connection from efficient communication theory to systematicity in natural language.

\end{abstract}

\section{Introduction}
\label{sec:intro}

\begin{figure}[t]
    \centering
    \includegraphics[width=0.8\linewidth]{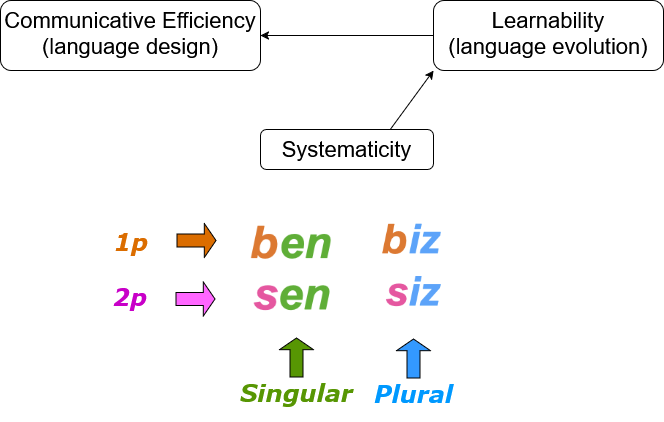}
    
    \caption{Turkish pronouns show \emph{systematic} form-meaning mappings: person is consistently marked by prefixes (e.g., \textit{s-} for second person), number by suffixes. Language evolution research demonstrates that such systematicity supports \emph{learnability}. Our model connects these findings, proposing that learnable, systematic mappings contribute to \emph{communicative efficiency}.
    }

    \label{fig:arabic-example}
\end{figure}

Languages express a vast array of meanings with a limited set of forms. A fundamental goal of natural language research is to understand how these forms map onto meanings, 
and why this mapping differs across languages. For example, standard English uses the pronoun \textit{you} for any second-person addressee, while other languages use distinct forms based on features like gender or number (e.g., \textit{sen} ``you (singular)'' vs.~ \textit{siz} ``you (plural)'' in Turkish; \autoref{fig:arabic-example}). What drives this cross-linguistic variation in form-meaning mappings?

We posit that this variation arises because languages evolve under two competing communicative pressures. On one hand, a general bias toward \textit{simplicity} encourages languages to minimize formal distinctions. This criterion favors the English pronoun system, with only one second-person form, as less cognitively demanding than Turkish, which requires the speaker to distinguish addresses by number. On the other hand, a competing bias toward \textit{accuracy} encourages languages to maintain formal distinctions, 
enabling precise decoding of meaning from form. For instance, a Turkish listener can instantly understand whether the speaker intends to address an individual (\textit{sen}) or a group (\textit{siz}), while an English listener must infer between these two meanings of the word \textit{you}. English and Turkish pronoun systems illustrate two different design solutions to the trade-off between simplicity and accuracy \citep{zaslavsky2021pronouns}.

This trade-off has been explored through two distinct research traditions. An influential line of work on \emph{efficient communication} \citep[e.g.][]{kemp2018semantic,gibson2019efficiency,hahn2020universals} analyzes how natural languages balance simplicity and accuracy in their lexicons. This research finds that lexicons achieve near-optimal simplicity-accuracy trade-offs across semantic domains including color \citep{Zaslavsky2018color}, kinship \citep{kemp2012kinship}, and numerals \citep{xu_2020_numeral}. 
In parallel, research on \textit{language evolution} \citep[e.g.][]{smith_language_2017,culbertson_artificial_2019} investigates how artificial languages evolve in laboratory settings. Through experimental manipulation, this work shows that interacting biases toward simplicity and accuracy drive the emergence of  core ``design features'' of natural language \citep{hockett_origin_1960,christiansen_language_2008,smith_how_2022}.

One such core feature is \textit{systematicity},\footnote{Systematicity enables another core design feature: \textit{compositionality}, the capacity to build complex expressions by combining discrete parts \citep{frege_letter_1914, mccurdy_toward_2024}.} in which discrete parts of linguistic forms reliably indicate specific meanings; consider the Turkish pronouns in \autoref{fig:arabic-example}, where prefixes  
indicate person (e.g. second person pronouns begin with \textit{s-}) and suffixes indicate number. Systematic mappings emerge in artificial languages as a design solution to the competing pressures of simplicity and accuracy \citep{smith_linguistic_2013,kirby_compression_2015,smith_how_2022,smith_communicative_2025}: decomposing complex expressions into parts yields a smaller, simpler lexicon, while recombining these parts enables fine-grained distinctions and accurate decoding. The language evolution literature thus identifies systematic form-meaning mappings as a core mechanism of efficient communication. However, this insight is not yet reflected in research on communicative efficiency in \textit{natural} languages, which 
lacks a general framework for modeling forms' internal structure.

We propose a unified information-theoretic framework to integrate these two research traditions and assess how systematic form-meaning mapping supports efficiency in natural language. Our key contribution is a novel complexity measure based on the learnability of meaning-to-form mappings. This measure captures systematicity in linguistic forms within the theoretical framework of efficient communication. 

We evaluate our framework in two  domains: verbs and pronouns. Verbs offer rich variation in how structured meanings map to forms across languages, and in many languages exhibit overtly systematic form-meaning covariation through inflectional morphology \citep{haspelmath_understanding_2010}. Pronouns provide a natural comparison case to an earlier analysis under the well-known Information Bottleneck model \citep{zaslavsky2021pronouns}, allowing us to directly compare approaches.

We find that our framework successfully represents structured form-meaning mappings in both domains, and consistently discriminates attested from unattested systems. Crucially, our comparison finds that explicitly modeling systematicity yields a more precise account of efficiency in language, better capturing the fine-grained regularities that characterize natural linguistic systems.

 \begin{table}[t]
    \centering
    \begin{tabular}{l|l||l|l|l}
    \multicolumn{2}{c||}{}& singular (sg) & dual (du) & plural (pl) \\\hline\hline
     \multirow{2}{*}{1}
        & m & {\textglotstop}a-\fong{12u3}-u & na-\fong{12u3}-u & na-\fong{12u3}-u \\
        & f & {\textglotstop}a-\fong{12u3}-u & na-\fong{12u3}-u & na-\fong{12u3}-u \\
    \hline
     \multirow{2}{*}{2} 
        & m & ta-\fong{12u3}-u & ta-\fong{12u3}-\={a}ni & ta-\fong{12u3}-\={u}na \\
        & f & ta-\fong{12u3}-\={\i}na & ta-\fong{12u3}-\={a}ni & ta-\fong{12u3}-na \\
    \hline
     \multirow{2}{*}{3}
        & m & ya-\fong{12u3}-u & ya-\fong{12u3}-\={a}ni & ya-\fong{12u3}-\={u}na \\
        & f & ta-\fong{12u3}-u & ta-\fong{12u3}-\={a}ni & ya-\fong{12u3}-na \\
    \end{tabular}
    \caption[Attested Paradigm for Classical Arabic]{Basic imperfective paradigm of a stem I Classical Arabic verb for feature values gender \{\textbf{m}asculine, \textbf{f}eminine\}, person \{1, 2, 3\}, and number \{sg,  du, pl\}.  
    Numbers (e.g. 12u3) represent root consonants. 
    }
    \label{tab:arabic_paradigm_example}
\end{table}

\section{Background}

\label{sec:background:para-syn}
\subsection{Paradigms and Syncretism}

A \textbf{paradigm} comprises systematic relations between word forms which are structured by grammatical categories such as person, gender, number, tense, or case. These relations can hold between distinct words, such as the Turkish pronouns shown in \autoref{fig:arabic-example}, or parts of words, 
such as the prefixes and suffixes shown in \autoref{tab:arabic_paradigm_example}.
Specific feature combinations are realized by the form in the corresponding \textit{cell} of the paradigm table (e.g. first person singular $\rightarrow$ \textit{{\textglotstop}a-}).

\textbf{Syncretism} occurs when two or more cells of a paradigm with distinct grammatical functions share the same surface form. For instance, 
\textit{ta-,-u} realizes multiple feature combinations in Arabic (\autoref{fig:arabic-example}, \autoref{tab:arabic_paradigm_example}). 
Within highly structured paradigmatic meaning spaces, syncretism directly corresponds to the form-meaning partitions relevant to communicative efficiency; for instance,  \citet{zaslavsky2021pronouns} show that pronoun syncretism patterns are optimized for efficient communication. By definition, syncretism reflects only identity relations between forms \citep{haspelmath_understanding_2010}. Our proposed model extends 
beyond syncretism to capture partial overlap in forms, such as the systematic correspondence between the grammatical second person and the prefix \textit{ta-} in Arabic verbs (\autoref{tab:arabic_paradigm_example}).

\label{sec:background:verbs}
\subsection{Systematicity in Verb and Pronoun Forms}

\paragraph{Verbs.} Verbal morphology offers a rich testbed to investigate structured forms and meanings. 
We consider two distinct types of verbal inflection.
Concatenative inflection attaches, i.e.~concatenates, affixes to a word stem; consider for example the English past tense suffix \textit{-ed}, as in \textit{jump, jumped}. Non-concatenative inflection requires changing the word stem; consider for example \textit{run, ran}.
We model Semitic and other Afro-Asiatic verbs, which are characterized by both kinds of morphology.
On the non-concatenative side, Semitic \emph{roots} 
typically comprise 3 consonants 
which combine templatically with vowels to produce stems \citep{semiticlanguages,semlang_2,semiticlanguages_lipinski}. These stems are then additionally combined with concatenative morphology in the form of suffixes and sometimes prefixes. For instance, the forms in \autoref{fig:arabic-example} arise from combining the root verb ``to write'' \textsc{ktb} with the stem templates and affixes shown in \autoref{tab:arabic_paradigm_example}.
We also model verbal morphology in Romance and Germanic languages, where inflection is mainly concatenative.

\label{sec:background:pronouns}
\paragraph{Pronouns.}
While inflectional morphology reflects systematic form-meaning association by definition \citep[][2]{haspelmath_understanding_2010}, pronouns do not generally display such rich formal structure. Systematicity, however, has also been identified at formal levels below morphology, such as specific sounds \citep[e.g.][]{blasi2016sound, pimentel2019meaning, monaghan2014arbitrary}. 
We posit that pronoun forms are also shaped by learnability, and propose modeling their efficiency with an appropriate complexity measure.

\section{Model} \label{sec:model}
We propose an information-theoretic framework to integrate systematicity into efficient communication theory. Our framework builds on a well-established insight: languages evolve under competing pressures for accuracy (enabling precise communication) and simplicity (minimizing cognitive demands). These competing pressures predict that linguistic systems should occupy a particular region of the accuracy-simplicity trade-off space, achieving near-optimal balance between the two.

\begin{figure}
\centering
    \includegraphics[width = 0.9 \linewidth]{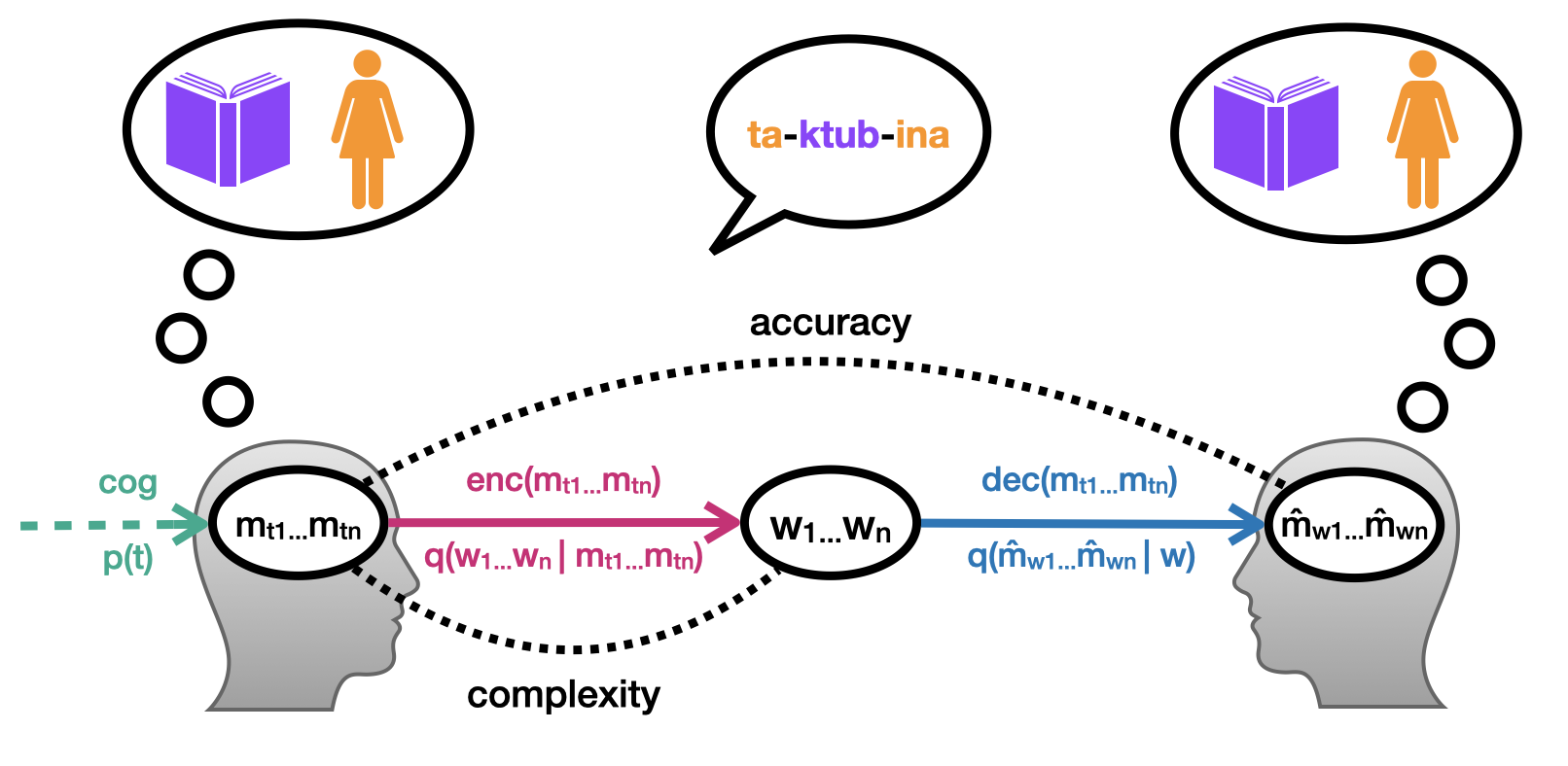}
    \caption[Communication Model]{Communication model, adapted from
    \citet{Zaslavsky2018color,zaslavsky2021pronouns}. Our model encodes the form $w$ as a sequence, and decodes it as an atomic unit.
    }
    \label{fig:communication_model}
\end{figure}

The Information Bottleneck (IB) model \citep{Zaslavsky2018color,Zaslavsky2019SemanticCO,zaslavsky2021pronouns} provides an influential formalization of this trade-off for natural language lexicons. In this model, a speaker samples an object $t$ based on a need distribution $p_\mathsf{cog}(t)$
and communicates it by producing a signal $w$ from an encoding distribution $q_\mathsf{enc}(w|m_t)$. The listener interprets $w$ using Bayesian inference to construct a probability distribution $q_\mathsf{dec}(\hat{m}|w)$ over possible meanings.  The encoding process is under pressure to minimize \emph{complexity}, while decoding is pressured to maximize \emph{accuracy}.
The IB framework evaluates this trade-off by comparing \textit{attested} semantic systems against \textit{counterfactual} alternatives; if attested systems consistently achieve better complexity-accuracy trade-offs than counterfactuals, this indicates they have been shaped for communicative efficiency.

The IB model, however, treats linguistic forms as atomic units with no internal structure. For example, the mappings ``3fs $\rightarrow$ \emph{ta- -u}'' and ``3ms $\rightarrow$ \emph{ya- -u}'' (\autoref{tab:arabic_paradigm_example}) could be equivalently represented as ``3fs $\rightarrow X$'' and ``3ms $\rightarrow Y$.'' This means the model can detect reduced complexity through \emph{syncretism} (using identical forms for multiple meanings) but not through \emph{systematic} form-meaning covariation \emph{within} forms (e.g., the suffix \emph{-u} in the third-person singular). Our framework addresses this gap while preserving the IB model's theoretical foundations and its measure of accuracy (\autoref{fig:communication_model}).

Our key contribution is a novel measure of complexity based on the learnability of meaning-to-form mappings. This measure captures systematicity in linguistic forms through a neural network encoder that learns character-level sequences, allowing us to quantify how structural regularities facilitate learning and reduce cognitive complexity.

\subsection{Complexity} \label{sec:complexity_measure}

\subsubsection{Our model}
We hypothesize that paradigms with more structured mappings will be more learnable.
Following the efficient communication literature, we model this with two components: a need distribution $p_\mathsf{cog}(t)$, and an encoder $q_{enc}(w | m_t)$. 

\paragraph{Need distribution.} $p_\mathsf{cog}(t)$ is a probabilistic weighting over targets $t$ associated with meanings $m_t$, 
reflecting communicative need. We assume that more frequent meanings have greater need, and estimate $p_\mathsf{cog}(t)$ using corpus frequency. This weights meanings according to a learner's exposure to different meanings at different frequencies.

\begin{figure}
    \centering
    \includegraphics[width = 0.8\linewidth]{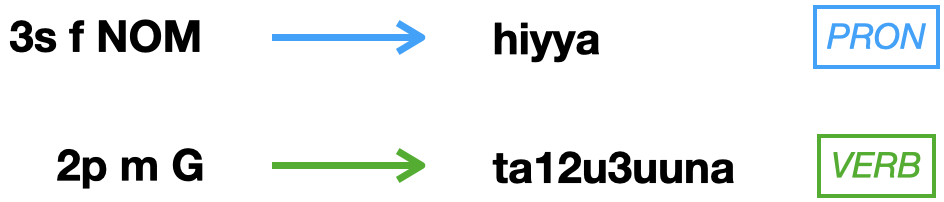}
    \caption{Example input and output used in training.}
    \label{fig:example_input_output}
\end{figure}

\paragraph{Encoder.} $q_{enc}(w | m_t)$  maps from a grammatical meaning $m_t$ to a linguistic form $w$. 
We implement this using a sequence-to-sequence neural network that takes morphosyntactic features as input ($m_t$) and generates the corresponding surface form as output ($w$; \autoref{fig:example_input_output}).\footnote{The encoded form $w$ is domain-specific: for pronouns, $w$ is the entire word; for verbs, $w$ represents only inflectional markers. The model removes whitespaces and reads letter by letter, including special markers for sequence boundaries. For example, "\textit{2p m G}" becomes "\textit{<sos>2pmG<eos>}" as input, with output "\textit{<sos>ta12u3uuna<eos>}".} 
Critically, by encoding forms as character sequences rather than atomic units, our encoder can exploit systematic regularities across forms. We expect training to converge more rapidly for more learnable paradigms. We therefore measure complexity in terms of the encoder's cross-entropy decay during learning.

\paragraph{Complexity measure.} 
Let $q_{enc}^{(T)}(w | m_t)$ be the encoder after $T$ training epochs. We define the cross-entropy training loss (\textbf{CETL}) as:
\begin{align}\label{eq:our-comp-measure}
   - \dfrac{\sum_{T=1}^{T_{max}} \sum_{t} p_\mathsf{cog}(t) \log q_\mathsf{enc}^{(T)}(w_t | m_t)}{T_{max}}
\end{align}
Here $w_t$ is the correct form for meaning $m_t$, and $T_{max}$ is the maximum number of epochs.
The core of this equation is $-\log q_{enc}^{(T)}(w_t | m_t)$, which quantifies how well the encoder predicts the form after $T$ training steps.
By summing over timesteps,  \autoref{eq:our-comp-measure} is smaller when training is faster. CETL is therefore information-theoretic, grounded in learnability, and fully general --- it can be applied broadly across languages and semantic systems.

\paragraph{Implementation.} Our neural network is a LSTM-based sequence-to-sequence (seq2seq) architecture,\footnote{We use LSTMs in keeping with prior neural models of morphology \citep[e.g.][]{wu-etal-2019-morphological, rathi-etal-2021-information, cotterell_arxiv18}.}
with two stacked LSTM layers in the encoder and in the decoder.
We train with batch size 1 and dropout rate 0.5.
Based on preliminary experiments, we set $T_{max}=50$ epochs\footnote{This was sufficient to achieve zero loss on some Semitic verbal paradigms. While not guaranteed sufficient for all variants, crucially, we hold $T_{max}$ constant across paradigms.} and measure total cross-entropy loss weighted by the need distribution.
We train ten separate networks for each original paradigm, and five for each counterfactual paradigm, and average over the results.\footnote{
All code will be made publicly available.
}

\subsubsection{Computational measures of complexity} 
Our learnability-based approach connects to several lines of research on linguistic complexity, though CETL uniquely captures systematicity in form-meaning mappings.

\paragraph{Learnability-based complexity.} Similar neural learnability-based approaches to complexity have been proposed, but none shares our focus on systematicity across forms. Steinert-Threlkeld, Szymanik and colleagues \citep{steinert-threlkeld_learnability_2019,steinert-threlkeld_ease_2020,carcassi_monotone_2021} also measure complexity via RNN learning rates within a communicative efficiency framework.  Johnson and colleagues \citep{complexity_i_e_relwork,johnson_testing_2024} similarly measure morphological complexity via RNN classification accuracy during artificial language learning. These proposed measures, however, rely on classification tasks, and do not attend to forms' internal structure. 
More recently, \citet{denic_recursive_2024} model morphosyntactic complexity in recursive numeral systems, finding an optimal trade-off with lexicon size. This work, like ours, addresses compositional morphological structure; however, it abstracts away from form, instead measuring complexity by average morpheme count. By contrast, CETL detects systematicity across forms (e.g. the shared prefix between “two” and “twenty”) without manual annotation. Compared to other learnability-based measures, our approach is more fine-grained, using a continuous information-theoretic measure which is critically sensitive to output forms.

\paragraph{Information-theoretic complexity.} Other information-theoretic measures of morphological complexity have been proposed, but most do not fit our specific application. 
\citet{complexity_i_e_def} define \emph{enumerative complexity} (\emph{e-complexity}; essentially the number of formal distinctions) and \emph{integrative complexity} (\emph{i-complexity}; the degree to which forms predict each other). They argue that natural languages vary substantially in e-complexity, but tend toward low i-complexity.
Variants of this account have been operationalized with seq2seq models \citep{cotterell2019complexity,johnson2020,complexity_i_e_relwork}. 
There is a conceptual link to our work, which uses seq2seq learners to model predictive relations between inflected forms. However, i-complexity estimates predictability based on variation \emph{across} paradigms (e.g.,~over different inflection classes). Similarly, \citet{wu-etal-2019-morphological} quantify the irregularity of inflected forms based on their predictability given the rest of the language. By contrast, we abstract over lexemes and estimate each paradigm's complexity \emph{independently} from the rest of the language.
\citet{rathi-etal-2021-information} introduce \emph{informational fusion}, which measures the extent to which a form cannot be predicted by the rest of the paradigm. This measure would be applicable in our domains but has a high computational cost, requiring re-fitting seq2seq models for each form in a paradigm.\footnote{Similar concerns apply to related evolutionary models with iterated learning \citep[e.g.][]{ackerman_no_2015,round_natural_2025}. } We prioritize broad coverage, making CETL the most efficient approach.

\paragraph{IB Complexity.} The Information Bottleneck model also uses an information-theoretic formulation of complexity based on need distribution and encoder. For a semantic system with meanings $M$ and forms $W$, the IB model defines complexity as the mutual information between intended meanings $m_t \in M_t$ and word forms $w \in W$: 
\begin{equation}
\label{eq:Ienc}
\begin{split}
    \sum\limits_{\substack{t\in\mathcal{U}\\w\in\mathcal{W}}}
    p_\mathsf{cog}(t) 
    q_\mathsf{enc}(w|m_t)
    \log \dfrac{q_\mathsf{enc}(w|m_t)}{q_\mathsf{enc}(w)}
\end{split}
\end{equation}
where 
$q_\mathsf{enc}(w) = 
    \sum_{t~\in~\mathcal{U}} p_\mathsf{cog}(t) 
    q_\mathsf{enc}(w|m_t)$.
This computes the bits required for the encoder $q_\mathsf{enc}(w|m_t)$ to communicate all meanings, weighted by the need distribution. However, because the IB encoder treats forms as discrete units, this measure of complexity is reduced by syncretism but not by systematic meaning-form covariation.\footnote{\citet{bruneaubongard_assessing_2025} identify other unintuitive properties of IB complexity, such as insensitivity to synonyms.} CETL addresses this gap by encoding $w$ as a character sequence, allowing systematic regularities across forms and meanings to reduce complexity through faster learning.

\subsection{Accuracy} \label{sec:accuracy_measure}

We adopt the IB model's widely-used, domain-general accuracy measure. Under this framework, a listener uses Bayesian inference to construct a probability distribution $q_\mathsf{dec}(\hat{m}|w)$ over possible meanings given the received form $w$ (\autoref{fig:communication_model}).
Accuracy is defined as the similarity between the speaker's intended meaning distribution $m_t$ and the listener's reconstructed distribution $\hat{m}_w$:  
\begin{equation}
\label{eq:Idec}
    \textsf{acc} = -\mathds{E}_\mathsf{dec} [D_{KL}[\hat{m}_w||m_t]] ~,
\end{equation}
where $D_{KL}$ is the Kullback-Leibler (KL) divergence.\footnote{\citet{zaslavsky2021pronouns} add a constant to make this quantity nonnegative, which changes results by a constant offset.} Following the literature, we assume that speakers always produce the correct form for a target meaning, so $m_t$ is a one-hot vector.

\begin{table*}[h!] 
\setlength\tabcolsep{1pt}

    \begin{subtable}[t]{0.32\textwidth}
        \centering
        \begin{tabular}{ll||l|l}
         \multicolumn{2}{c|}{}& sg & pl\\\hline\hline
         \multirow{1}{*}{1}
            &  & 'a- & na- \\
        \hline
         \multirow{2}{*}{2} 
            & m & \underline{ta-} & ta- -u \\
            & f & \sred{\underline{ta-}} &  ta- -u \\
        \hline
         \multirow{2}{*}{3}
            & m & ya- & ya- -u \\
            & f & \sred{ta- -i} & ya- -u  \\
        \end{tabular}
    \caption{A \textbf{structural} permutation over the person category (\texttt{PERS}) that \textbf{increases} naturalness: the form \textit{ta-} now has only value \texttt{2} for grammatical person, whereas before it covered values \texttt{2,3}.\label{tab:ex_more_natural}
    }
    
    \end{subtable}
    \hfill
    \begin{subtable}[t]{0.32\textwidth}
        \centering
        \begin{tabular}{ll||l|l}
         \multicolumn{2}{c|}{}& sg & pl\\\hline\hline
         \multirow{1}{*}{1}
            &  & 'a- & na- \\
        \hline
         \multirow{2}{*}{2} 
            & m & ta- & ta- -u \\
            & f & \sred{\underline{ya- -u}} &  ta- -u \\
        \hline
         \multirow{2}{*}{3}
            & m & ya- & \underline{ya- -u} \\
            & f & ta- &  \sred{ta- -i}  \\
        \end{tabular}
    \caption{A \textbf{structural} 
    permutation over multiple categories (\texttt{PERS}, \texttt{NUM}) that \textbf{decreases} naturalness: the form \textit{ya- -u} now has values \texttt{2,3} for grammatical person, and \texttt{sg}, \texttt{pl} for number, whereas before it covered only the feature values \texttt{3}, \texttt{pl}.}    \label{tab:ex_less_natural}
    \end{subtable}
    \hfill
    \begin{subtable}[t]{0.32\textwidth}
        \centering
        \begin{tabular}{ll||l|l}
         \multicolumn{2}{c|}{}& sg & pl\\\hline\hline
         \multirow{1}{*}{1}
            &  & 'a- & na- \\
        \hline
         \multirow{2}{*}{2} 
            & m & \scyan{ya- -u} & ta- -u \\
            & f & ta- -i &  ta- -u \\
        \hline
         \multirow{2}{*}{3}
            & m & ya- & \scyan{ta-} \\
            & f & ta- &  \scyan{ta-}  \\
        \end{tabular}
    \caption{A \textbf{form-only} permutation over multiple categories (\texttt{PERS}, \texttt{NUM}, \texttt{GEN}) which changes form realizations but does not affect paradigm structure, i.e. the distribution of syncretic forms; compare to \autoref{tab:ex_less_natural}.
    }
    \label{tab:ex_surface}
    \end{subtable}

    \caption{Example permutations applied to Dialectal Arabic conjugation.
    Identical (i.e. syncretic) forms whose naturalness is affected by the permutation are \underline{underlined}. 
    Structural permutations affect syncretic patterns, but form-only permutations do not. Altered forms are highlighted in \sred{orange} (structural) or 
    \scyan{cyan} (form-only).
    }\label{tab:example_permutations}
\end{table*}

To specify $\hat{m}_w$, we require the need distribution $ p_\mathsf{cog}(t)$ and an underlying representation of the semantic domain, reflecting the listener's knowledge. Following \citet{zaslavsky2021pronouns}, we encode grammatical meanings as categorical features \citep[cf.][]{macwhinney_word_2015}, and specify their similarity as a weighted Hamming distance $d(u,t)$\footnote{They include a free parameter $\gamma$; we set $\gamma=1$.}:
\begin{equation}
m_t(u) \propto \exp(-d(u,t)).
\end{equation}
However, where \citet{zaslavsky2021pronouns} use binary features, we employ a general categorical encoding scheme.\footnote{We compare against their feature encoding in \autoref{app:direct_comparison} and find that our results are robust to the change.}
For feature representations $u,t$, we define $d(u,t)$ as the number of dimensions $i$ where $u_i \neq t_i$. This groups combinatorial meanings with shared features; for example, “first-person singular past” is closer to the meaning “first-person plural past” than “third-person singular future.”
Importantly, while this IB accuracy measure posits complex structure in semantic \emph{meaning}, it still represents the given form $w$ as a discrete atomic unit.

\section{Evaluation}

We formulate two complementary hypotheses about how systematicity affects communicative efficiency in natural language paradigms. To test these hypotheses, we compare attested paradigms against systematically generated counterfactuals. We use this approach to evaluate our model on verb and pronoun data, and compare to the IB framework.

\paragraph{Hypothesis 1 (Efficiency).} Attested paradigms achieve better complexity-accuracy trade-offs than counterfactual alternatives. This follows standard practice in the IB literature \citep[e.g.][]{kemp2012kinship,zaslavsky2021pronouns} and tests whether natural language paradigms are optimized for communicative pressures.

\paragraph{Hypothesis 2 (Naturalness).} Among counterfactual paradigms, those with more natural syncretism patterns (where syncretic forms share similar meanings) are more learnable \citep{saldana_more_2022} and thus have lower complexity. This tests whether our model captures not only form-internal, but also paradigm-level structure.

\subsection{Generating counterfactual paradigms} \label{sec:methodology:perms}

We generate counterfactuals by permuting existing paradigms, creating alternatives that vary in two key ways: their syncretism patterns (relevant to systematicity \emph{and} naturalness) and their specific form-meaning mappings (relevant to systematicity alone). All counterfactuals are compared to their source paradigm as baseline.

\textbf{Structural permutations} swap the contents of paradigm cells, altering syncretism patterns.  For example, we might permute the person feature while holding other distinctions constant, swapping ``2sg f → \textit{ta- -i}'' with ``3sg f → \textit{ta-}'' (\autoref{tab:ex_more_natural}). We can permute single features (e.g.,$[2 \Rightarrow 3]$) or multiple features simultaneously (e.g.,  $[2 \Rightarrow 3]~[sg \Rightarrow pl]$; \autoref{tab:ex_less_natural}). These permutations change which meanings are expressed by syncretic forms, affecting the paradigm's naturalness score (defined in \ref{sec:naturalness_measure}).

\begin{figure*}[t]
    \centering
    \includegraphics[width=1\linewidth]{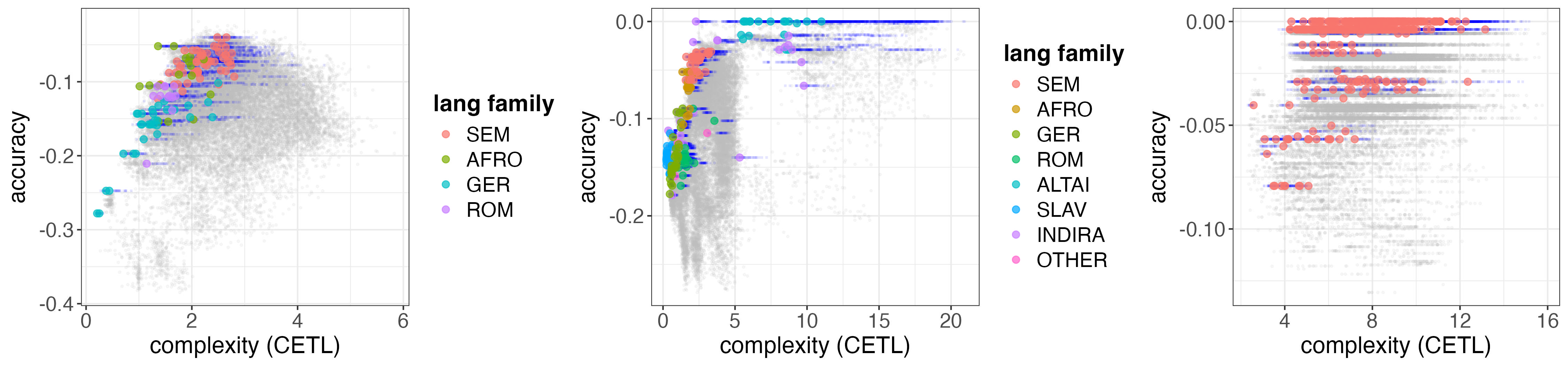}
    \caption{Complexity and accuracy in our model. Across domains, real, attested paradigms (colored, one dot per language) are more efficient than nearly all counterfactual structural (gray) and surface (blue) permutations.
    }
    \reproducibilityNote{
        - Run efficiency_plots.R.
        - The plot is in <OUTPUT_DIR>/EFF_ALL_CETL.pdf. (EFF_ALL_CETL_PERM_SHUF.pdf for plot with surface vs structural separated)
        - The output directory can be set in the first lines (OUTPUT_DIR <- "[...]")
  }
    \label{fig:efficiency}
\end{figure*}

\textbf{Form-only permutations} swap surface forms while maintaining the original syncretism patterns. For example, we might exchange the form realizing ``2sg m'' with the form realizing ``3sg [m, f],'' which retains the original syncretic structure of the paradigm (\autoref{tab:ex_surface}). By definition, these permutations do not affect naturalness scores. Critically, they also cannot affect complexity in the IB model, which treats forms as atomic units. However, they can affect our complexity measure if the permutation disrupts systematic regularities (e.g., breaking a consistent prefix pattern). Form-only permutations thus provide a key test of whether our model captures systematicity beyond syncretism.

\subsection{Measuring naturalness} \label{sec:naturalness_measure}\label{def:naturalness}

Our naturalness hypothesis builds on an insight already present in the IB accuracy measure (\ref{sec:accuracy_measure}): grammatical meanings with shared features are more similar. Systematicity suggests that this principle should extend to forms. In particular, \textit{syncretic} forms ---which share all of their formal features --- should also share many aspects of grammatical meaning. For example, the syncretic form \textit{na- -u} in \autoref{tab:arabic_paradigm_example} (top two rightmost cells) expresses meanings that all share $person={1}$. By contrast, an "unnatural" syncretism might use the same form for meanings with no shared features. 

\citet{saldana_more_2022} develop this intuition as a \textbf{naturalness} gradient in syncretism patterns. Inspired by their analysis, we define the \emph{unnaturalness score} at the paradigm level:
\begin{equation}
\label{eq:unnat_score}
   \mathsf{unnat}_\pi = \sum\limits_{c \in \mathsf{SynClass_\pi}} \sum\limits_{f \in \mathsf{FeatCat_\pi}} (\#c_f -1)
\end{equation}
where $c_f$ are all feature values that exist in the syncretism class $c$ for the feature category $f$.
This counts the number of feature values that differ within each syncretic form. A low unnaturalness score indicates natural syncretism patterns where syncretic forms express similar meanings. We compute unnaturalness scores for counterfactual paradigms relative to their attested baselines.

\reproducibilityNote{
To create the results files:
a) run create_permutations.py
b) run compute_naturalness.py
c) run model_runner.py
d) run create_ib_files.py
e) run compute_hit_fail_rates.py
}

\subsection{Data}

\autoref{tab:overview_dataset} summarizes the languages and features used in our evaluation. Domains 1 and 2b use resources we collected ourselves; see \autoref{data_sources} for sources and citations.

\begin{table}[t]
\scriptsize

\setlength{\tabcolsep}{3pt}
    \centering
    \begin{tabular}{r|lrrl}
         & subsets  & \#languages & \#total permutations  & feature categories\\
       \hline
        \textsc{ppd} &  &  561 & 93845 
            & PERS, NUM\\
        \hline
        \textsc{pron} 
        & \textsc{Pr\_Sem} & 45 &16437
         &  \multirow{8}{*}{\begin{tabular}{l}PERS, NUM,\\ GEN, CASE\end{tabular}}\\
        & \textsc{Pr\_Afro} & 19 &6677\\
        & \textsc{Pr\_Ger} & 30 &10649\\
        & \textsc{Pr\_Rom} & 17 &3605\\
        & \textsc{Pr\_Slav} & 20 &6063\\
        & \textsc{Pr\_IndoIran} & 18 &3616 \\      
        & \textsc{Pr\_Altaic} & 19 &3765 \\
        & \textsc{Pr\_Other} & 9 &2264  \\
        \hline
         \textsc{verb}  
         & \textsc{Verb\_Sem} & 56& 19700
         &  \multirow{4}{*}{\begin{tabular}{l}PERS, NUM,\\ GEN, TEN\end{tabular}}\\
         & \textsc{Verb\_Afro} & 13 & 5773\\
         & \textsc{Verb\_Ger} & 32 & 8875\\
         & \textsc{Verb\_Rom} & 12 & 2245\\
    \end{tabular}

    \caption{Summary of the datasets. For details of family-specific features within each category, see \autoref{app:paradigm_representation}.
    } 
    \label{tab:overview_dataset}
\end{table}

\paragraph{Domain 1: Verbal Morphology (\textsc{verb}).} 
We test our model on verbal inflection in Semitic, non-Semitic Afro-Asiatic, Germanic, and Romance languages, including both concatenative and non-concatenative morphology. \autoref{tab:arabic_paradigm_example} illustrates our approach to form and feature representation; see \autoref{app:paradigm_representation} for more details.

\paragraph{Domain 2a: Pronouns (\textsc{ppd}).}
To directly compare with \citet{zaslavsky2021pronouns}, we use the Pronoun Paradigms Database \citep{parabank-pronouns}, which contains hundreds of languages across families. We follow their simplified feature scheme (\autoref{tab:overview_dataset}, top row), which uses only masculine forms and omits dual number.

\begin{figure*}
    \centering
    \includegraphics[width=0.95\linewidth]{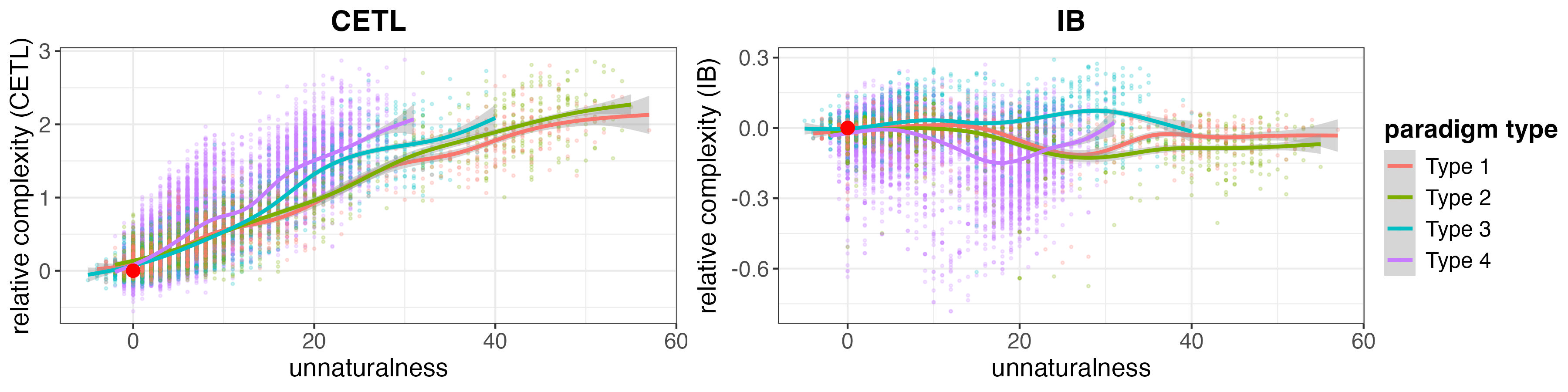}
    \caption{ Complexity plotted against unnaturalness for Afro-Asiatic verbs by model, CETL (left) and IB (right).
    We group results by paradigm type (reflecting different grammatical categories; cf. App. \ref{app:feat_repr_APP}).
    CETL positively correlates with unnaturalness, meaning more natural paradigms have lower complexity, while IB shows no correlation.
    }
    \label{fig:complexity_vs_naturalness}
    \reproducibilityNote{
        - Run naturalness_plots.R.
        - The plot is in <OUTPUT_DIR>/NAT_VERB_SemAfro_langfams.pdf.
        - The output directory can be set in the first lines (OUTPUT_DIR <- "[...]")
  }
\end{figure*}

\paragraph{Domain 2b: Detailed Pronouns (\textsc{pron}).}
We also test on full pronoun paradigms for selected language families (Semitic, Germanic), including case and gender. Unlike Domain 2a, the feature set varies by language to capture language-specific distinctions.

\paragraph{Need Distribution.}
We estimate $p_{cog}(t)$ by combining the need distribution from \citet{zaslavsky2021pronouns} with additional corpus frequency estimates for gender and dual/plural distinctions, assuming uniform distributions otherwise.
Results are robust to this choice (\autoref{app:modeling_details}).

\section{Results}

We first evaluate whether our model supports the efficiency and naturalness hypotheses 
(\ref{sec:learnability}), then compare its performance to the IB approach (\ref{sec:comparison}). For additional detailed results, see  \autoref{app:detailed_results}.

\begin{table*}[th!]
\centering
\small
\setlength{\tabcolsep}{3pt}
    
    \begin{subtable}[t]{0.49\linewidth}
    \centering
    \begin{tabular}{c|rr|rr|rr|r}
    & \multicolumn{2}{c|}{$C_M$ (\%)} & \multicolumn{2}{c|}{$I_M$ (\%)} & \multicolumn{2}{c|}{$\mathsf{Perf_M}$}\\
    &  $\mathsf{CETL}$ & $\mathsf{IB}$ & $\mathsf{CETL}$ & $\mathsf{IB}$ & $\mathsf{CETL}$ & $\mathsf{IB}$ & support\\
    \hline
    \textsc{ppd} & \tgr{72.37} & 4.82 & 2.19 & \tred{0.73} & \tbl{70.19} & 4.10 & \tgray{44112}\\
   \textsc{pron} & \tgr{90.12} & 48.71 & \tred{1.48} & 4.53 & \tbl{88.64} & 44.18 & \tgray{38623}\\
    \textsc{verb} & \tgr{76.32} & 41.03 & \tred{3.77} & 4.08 & \tbl{72.55} & 36.95 & \tgray{32825}\\
    \end{tabular}
    \caption{Structural Permutations}
    \label{tab:hitfailrates_PERM}
    \end{subtable}
    \begin{subtable}[t]{0.49\linewidth}
     \centering
    \begin{tabular}{c|rr|rr|rr|r}
    & \multicolumn{2}{c|}{$C_M$ (\%)} & \multicolumn{2}{c|}{$I_M$ (\%)} & \multicolumn{2}{c|}{$\mathsf{Perf_M}$}\\
    &  $\mathsf{CETL}$ & $\mathsf{IB}$ & $\mathsf{CETL}$ & $\mathsf{IB}$ & $\mathsf{CETL}$ & $\mathsf{IB}$ & support\\
    \hline
   \textsc{ppd} & \tgr{65.81} & 0.00 & \tred{34.19} & 100 & \tbl{31.63} & -100 & \tgray{51061}\\
   \textsc{pron} & \tgr{89.21} & 0.00 & \tred{10.79} & 100 & \tbl{78.42} & -100 & \tgray{10697}\\
    \textsc{verb} & \tgr{78.23} & 0.00 & \tred{21.77} & 100 & \tbl{56.47} & -100 & \tgray{3781}\\
    \end{tabular}
    \caption{Form-only Permutations}
    \label{tab:hitfailrates_SHUF}
    \end{subtable}
    \reproducibilityNote{
        - Run hitfail_tables.R.
        - The tables are printed to the console.
  }
    \caption{Performance for structural (left) and surface permutations (right). 
    CETL outperforms IB across the board.}
    
    \label{tab:hitfailrates}
\end{table*}

\subsection{Testing efficiency and naturalness}
\label{sec:learnability}

\paragraph{Efficiency.} 
\autoref{fig:efficiency} shows that attested verb and pronoun paradigms are more efficient than nearly all counterfactual permutations: they show lower CETL and higher accuracy, with statistical significance across all domains (details in Appendix \ref{app:add_res:EFF}). This confirms Hypothesis 1: natural language paradigms occupy a near-optimal region of the complexity-accuracy trade-off space, consistent with communicative efficiency theory.

\paragraph{Naturalness.}
\autoref{fig:complexity_vs_naturalness} (left panel) shows a 
positive correlation ($\rho = 0.5745$; \textit{p} $< 2.2e-16$) 
between CETL and unnaturalness for Afro-Asiatic verbs. 
Since the unnaturalness scores depend upon paradigm size, we calculate averaged per-language correlations for each domain, finding
a 
lower correlation of 0.36 for \textsc{ppd}, and strong correlations of 0.82 and 0.88 for \textsc{pron} and \textsc{verb} respectively.
 This confirms Hypothesis 2: paradigms with more natural syncretism patterns are more learnable, and thus have lower complexity.

Together, these results validate our core hypothesis: systematicity in form-meaning mappings facilitates learning and thereby reduces complexity.

\subsection{Comparison to the IB Model}
\label{sec:comparison}

Our CETL measure is sensitive to surface form structure, which we expect to increase discriminative capacity relative to the IB model's complexity measure. Specifically, we expect CETL to correctly identify attested paradigms as more efficient than counterfactual permutations, and show a stronger correlation than IB to paradigm naturalness.

\paragraph{Evaluation.}
We measure each model's ability to correctly identify attested paradigms as more efficient than counterfactuals. For each counterfactual permutation, we classify each estimate as either
$\mathsf{Correct (C)}$: worse than the attested paradigm in both accuracy and complexity, or worse in one measure while equal in the other --- or $\mathsf{Incorrect (I)}$: better than or equal to the attested paradigm in both measures.
We define the performance of a model $M$ as $\mathsf{Perf}_M = \mathsf{C}_M - \mathsf{I}_M$ (i.e., the difference between correct and incorrect identifications). We consider CETL to outperform the IB model when it performs better by at least 5\% of the permutations for a given language.

\paragraph{Results.} 
\autoref{tab:hitfailrates_SHUF} shows that CETL correctly identifies 65\% to nearly 90\% of form-only permutations as less efficient than attested paradigms, while the IB model cannot distinguish these permutations at all (by design, since they have identical syncretism patterns). This demonstrates that CETL captures systematicity beyond mere syncretism.
\autoref{tab:hitfailrates_PERM} shows that CETL consistently outperforms the IB model in identifying attested paradigms as more efficient than structural permutations.
Finally, \autoref{fig:complexity_vs_naturalness} shows that CETL (left) strongly correlates with naturalness, with correlations exceeding 0.8 for PRON and VERB, while the IB model (right) shows no correlation for either domain. 
This confirms that CETL successfully captures fine-grained regularities in syncretism patterns that the IB model cannot detect.
We conclude that our learnability-based complexity measure better captures systematic form-meaning mappings, enabling better discrimination between attested and counterfactual paradigms.

\section{Conclusion}\label{sec:discussion}

Languages vary widely in form-meaning mappings. 
We propose a unified framework that integrates systematicity between meaning and form  into efficient communication theory,
and evaluate our model 
on hundreds of languages from diverse families. Our results show that: (1) attested paradigms are more efficient than counterfactuals; (2) paradigms with more natural syncretism patterns are more learnable; and (3) our model outperforms the IB approach in discriminating attested from counterfactual paradigms.
This work connects efficient communication theory with language evolution research on systematicity, showing how communicative pressures shape language.

\section*{Limitations}

Our analysis focuses exclusively on discrete, paradigmatically structured domains—specifically verbal inflection and pronouns. The applicability of CETL to other semantic domains remains uncertain. In particular, continuous domains like color present a fundamental challenge: they lack the discrete categorical structure required by our sequence-to-sequence architecture.

While CETL captures fine-grained systematicity that the IB measure misses, we have not established whether it is universally preferable or whether trade-offs exist. For instance, we do not replicate the fine-grained feature weight optimization of \citet{zaslavsky2021pronouns}, leaving open whether CETL supports such analyses as effectively. The two measures may prove complementary, with different applications suited to each.

Lastly, unlike some prior work in the efficient communication literature, we do not provide an explicit Pareto frontier of maximally efficient paradigms. Doing so would require specifying all possible counterfactual paradigms, including all forms compatible with the phonology and phonotactics of each language. Our permutation-based approach generates structured counterfactuals but does not exhaustively sample the space of possibilities, limiting our ability to make strong optimality claims about attested systems.

\section*{Ethics Statement}
This paper concerns foundational research on the structure of language. We do not foresee immediate ethical implications.

\section*{Acknowledgements}
Funded by the Deutsche Forschungsgemeinschaft (DFG, German Research Foundation) – Project-ID 232722074 – SFB 1102. We are grateful to Richard Futrell, Mora Maldonado, Carmen Saldana, and Noga Zaslavsky for helpful discussion.

\bibliography{anthology,custom}
\bibliographystyle{acl_natbib}

\appendix

\section{Representation Details}
\label{app:paradigm_representation}

Here, we describe our approach to paradigm representation paradigms in detail, covering both form and meaning (feature) representations.

\subsection{Representing Verbal Paradigms Across Language Families}

\paragraph{Philosophy: Single Paradigms, No Variability between Verbs}
An important design choice in our study is to specifically focus on individual paradigms, without regard for variability across different verb classes or irregular verbs (except for the weak-strong distinction in Germanic -- which translates to two fundamentally different ways of marking tense), to isolate the effect of the regularity of the paradigm-level form-meaning mapping.
That is, we only include one or two (as described below) paradigms per language, deliberately excluding irregular verbs, morphological interaction with lexical stems, etc. When we include more than one paradigm for a single language, we model them fully separately.
The motivation for this choice is that it allows us to \emph{entirely control} for the effect of several proposed prominent models of complexity in morphological paradigms, including i-complexity \citep{complexity_i_e_def} and others \citep{cotterell2019complexity, wu-etal-2019-morphological}: Our paradigms and all their counterfactual variants have the same (essentially zero) complexity under those models. 
We evaluate robustness to this choice in \aref{app:classes}.

Below, we describe in detail how we represented verbal paradigms for training the Seq2Seq models in different languages.

\paragraph{Afroasiatic Languages}
In many Afroasiatic languages (including the Semitic ones), verbs use a non-concatenative morphology based on \emph{root consonants}. Therefore, we represent Semitic roots as \textit{1-2-3}, where each number represents one root consonant, and fill in the vowel patterns, and suffixes. Doubled root consonants are represented as doubled numbers. For example Arabic "\textit{(ana) a\textbf{kt}a\textbf{b}u}" would be represented as \textit{a12a3u} and "\textit{(huwwa) \textbf{k}u\textbf{tt}i\textbf{b}a}" as \textit{1u22i3a}.

Cushitic verbs use prefix conjugation (PC) and suffix conjugation (SC) as two separate conjugation classes. We include two different paradigms for Cushitic languages: one for PC and one for SC.

For those Afroasiatic languages in which verbs do not exhibit noncatenative morphology, we represent the stem as \textit{123}, irregardless of it's length and consonant-verb makeup. 
Those languages usually have tonal elements, which we represent with \textit{H}, \textit{L}, \textit{F} for a high, low or falling tone.
For example Yemsa (Omotic) "\textit{\textbf{zag}ín}" would be represented as \textit{123iHn}.

Semitic roots usually structurally form several different derived verbal stems conveying different derived meanings (e.g.~reflexive, causative, intensive). 
We focus on the so called G-stem or Form-I-Stem\footnote{Often also called the fa3al-stem in Arabic linguistics, and the qal or pa'al stem in Hebrew linguistics.}, and choose only one specific voweling pattern, in case several vowel combinations exist (e.g. \textit{-a-a-}, \textit{-a-i-} etc.~in Arabic).
We disregard any derived stems as well as any further irregularities (such as weak root consonants) in the paradigms.

\paragraph{Germanic and Romance Languages}
Some Germanic verbs display a form of non-concatenative morphology (\autoref{sec:background:verbs}) called \textit{ablaut}, where the main vowel of the verb, the \emph{stem vowel}, is changed in certain forms of the conjugation. 
Therefore, we represent all Germanic stems with their stem vowel as \textit{1v2}, where \textit{v} represents the stem vowel\footnote{In some languages and conjugation classes, this vowel consists of two consecutive vowels.}.
For example, the German verb forms "\textit{(ich) empf\textbf{i}ng}" and "\textit{(du) empf\textbf{ä}ngst}" would be represented as \textit{1i2} and \textit{1ä2st}.
We include two different paradigms per Germanic language: one \emph{strong verb} paradigm, where the past conjugation is based on ablaut, and one \emph{weak verb} paradigm where the past conjugation is based on affixed dentals.

For Romance languages, we only use paradigms without internal changes, and represent all paradigm forms as pure suffixes. Romance languages usually have three basic conjugation classes (\textit{-a-}, \textit{-e-}, \textit{-i-}). We only use one of the conjugations per language, since the overall structure of the paradigms is mostly not affected by the specific choice of the class.

We disregard any further kinds of irregular verbs, including stem vowel shift or diphthongization in the Romance paradigms.

$~$\\
Some languages have more than one realization for certain feature combinations, irregardless of flexion classes. Especially in cases where this effects the syncretism pattern of the paradigm (e.g.~in Serbian certain short accusative forms can be syncretized with the genitive forms, or have distinct realizations) or might have an effect on the learnability of the paradigm (e.g.~in Ge'ez the 3rd person plural can have a very similar form to the 3rd person singular, or have an unrelated form), we use distinct paradigms.

\paragraph{Representations of Forms}
The exact representation of the phonemes in the languages depend on the specific language.
For Semitic and Afroasiatic languages we use transliterations that convey a phonemic representation of the language.
Long vowels and geminated consonants are represented as double letters.
Tonal elements are represented with capital letters put after the affected vowel.

For Germanic and Romance languages, we only depart from the official orthography in cases where phonemes are inherently different from writing, and might affect the syncretism patterns and/or learnability. This is the case for example in French where the verb forms "\textit{(je) mang\textbf{e}}", "\textit{(tu) mang\textbf{es}}", "\textit{(ils) mang\textbf{ent}}" are all pronounced the same.
For languages marking stress orthographically (including Latin) with accents we keep this in the representation.
For Germanic we chose verbal classes (i.e.~which specific vowel combinations are used) such that interference between orthography and phonemes is minimal. 

For pronouns we use the official orthography for languages originally written in Latin script, and transliterations representing the phonemes for languages originally written in other scripts.

\subsection{Representing Features}
\label{app:feat_repr_APP}

\paragraph{Feature Representation}
As described in \autoref{sec:model}, the speaker's mental representation of a meaning for a target referent $t$ is a probability distribution $m_t(u)$ over $u \in \mathcal{U}$.
\citet{zaslavsky2021pronouns} model $m_t(u)$ as assigning probability to different $u$ depending on the overlap in feature representations to $t$:
\begin{equation}\label{eq:mt-density-exponential}
m_t(u) \propto \exp(-\gamma\cdot d(u,t)).
\end{equation}
To define those feature encodings, each meaning is mapped to a vector in a multi-dimensional discrete feature-space.
The original conceptual space from \citet{zaslavsky2021pronouns} uses a custom binary encoding based on (1) the communicative roles \texttt{a}, \texttt{o}, \texttt{s}, (2) binary encoding of a three-way number distintion.
We find this encoding not scalable to the larger meaning we require for verbs, and thus resort to a general categorical encoding to sidestep the need for arbitrary binarization.
For a pair of feature representations $u,t$, we then define $d(u,t)$ as the number of dimensions $i$ where $u_i \neq t_i$.

\paragraph{Dimensions of Paradigms}

\label{app:def_stuff}

Here, we describe the details of the paradigm dimensions in each of the three domains.
\autoref{tab:appendix_overview_dataset} shows an overview of the features used for each data set.

\paragraph{\textsc{ppd}}
For the basic pronouns, each meaning has the form 
$m = \langle number, person \rangle$ where 
\begin{compactdesc}
\item $number \in\{singular,plural\}$, and 
\item $person \in \{1,2,3,12\}$.\footnote{Here, $12$ represents inclusive plural.}
\end{compactdesc}
For the \textsc{ppd} experiments, all language paradigms use all possible values, irregardless of whether the distinction is actually existing in the language.

\paragraph{\textsc{verb}}
For the verbs, we define \\
$m = \langle number, person , gender, tense \rangle$ where 
\begin{compactdesc}
    \item $number \in$ \{singular, dual, plural\},
    \item $person \in$ \{1, 2, 3\}, 
    \item $gender \in$ \{masculine, feminine\}, and 
    \item $tense \in$ \{perfective, imperfective,
                subjunctive, jussive, converb\}.
\end{compactdesc}
For the \textsc{verb} experiment each language uses the number, person, and gender values that exist in the language family, irregardless of weather the specific language utilized them (e.g.~the German paradigm uses dual, even though dual does not exist in modern Germanic languages anymore, but was existing in Proto-Germanic).
Tense values depend on the specific language.

\autoref{tab:verb_paradigm_classes} lists the languages used for each paradigm type for the \textsc{verb} paradigms.

\paragraph{\textsc{pron}}
For detailed pronoun paradigms, we define 
\[m = \langle number, person, gender, case \rangle \]
and add a couple of possible values for the features depending on whether the respective language uses them.
For the $person$ we add 3 different forms of formality for the 2nd person (\textit{2informal}, \textit{2formal}, \textit{2highformal}),
3 different types of remoteness for the 3rd person (\textit{3proximal}, \textit{3distal}, \textit{3remote}) and a \textit{neutral\_reflexive} form as used in Spanish or Russian. 
Some languages might have a combination of formality and remoteness in the 3rd person, for which we add further possibilities (e.g.~\textit{3formalproximal}, \textit{3formalremote} etc.). 
For the $gender$ we add a $neutral$ gender as used in Germanic and an \textit{animate} vs.~\textit{inanimate} distinction as used in some Slavic languages.
We use a wide range of cases depending on the specific language (\textit{nominative}, \textit{genitive}, \textit{dative}, \textit{accusative}, \textit{locative}, \textit{ablative} etc.).

\autoref{tab:pron_paradigm_classes} lists the languages used for each paradigm type for the \textsc{pron} paradigms.

\begin{table*}[ht]
\small
    \centering
    \begin{tabular}{r|ll}
         & subsets  & feature sets\\
       \hline
        \textsc{pron} & 
            & \texttt{PERS} = \{1,2,3,12\}; \\
         && \texttt{NUM} = \{s,p\}\\
        \hline
        \textsc{pr} 
        & \textsc{Pr\_Sem}
         &  \texttt{PERS} = \{1,2,3\}; \\
         &&\texttt{NUM} = \{s,d,p\}; \\
         &&\texttt{GEN} = \{m,f\}; \\
         && \texttt{CASE} = \{isolated, suffixal\}\\
        & \textsc{Pr\_Afro} 
         &  \texttt{PERS} = \{1,2,3\}; \\
         &&\texttt{NUM} = \{s,d,p\}; \\
         &&\texttt{GEN} = \{m,f\}; \\
         && \texttt{CASE} = \{isolated, suffixal\}\\
        & \textsc{Pr\_Ger}
         &  \texttt{PERS} = \{1,2,3,2v,2h\}; \\
         &&\texttt{NUM} = \{s,d,p\}; \\
         &&\texttt{GEN} = \{m,f,n\}; \\
         && \texttt{CASE} = \{nom, gen, dat, akk\}\\
        & \textsc{Pr\_Rom} 
         &  \texttt{PERS} = \{1,2,3,r,2v,2h\}; \\
         &&\texttt{NUM} = \{s,p\}; \\
         &&\texttt{GEN} = \{m,f,n\};\\
         &&  \texttt{CASE} = \{nom, gen, dat, akk, disj, con, akk-l, dat-l\}\\
        & \textsc{Pr\_Slav} 
         &  \texttt{PERS} = \{1,2,3,r\}; \\
         &&\texttt{NUM} = \{s,d,p\}; \\
         &&\texttt{GEN} = \{m,f,n,v,i\}; \\
         && \texttt{CASE} = \{nom, gen, dat, akk, instr, loc, gen-s, dat-s, akk-s\}\\
        & \textsc{Pr\_IndoIran} 
         &  \texttt{PERS} = \{1,2,3,2v,2h,2j,3e,3x,3w,12, 3ev,3eh,3ej,3xv,3xh,3xj,3wv,3wh,3wj\}; \\
         &&\texttt{NUM} = \{s,d,p\}; \\
         &&\texttt{GEN} = \{c,m,f,n\};  \\
         &&\texttt{CASE} = \{nom, obl, gen, akk, dat, loc, instr, abl, erg, indir, obj\}\\         
        & \textsc{Pr\_Altaic} 
         &  \texttt{PERS} = \{1,2,3,2v,2h,12\}; \\
         &&\texttt{NUM} = \{s,p\}; \\
         &&\texttt{GEN} = \{c\};  \\
         &&\texttt{CASE} = \{nom, gen, dat, akk, abl,instr, loc, alla, equa, simi, comm, prol, dir\}\\
        & \textsc{Pr\_Other} & \\
        \hline
         \textsc{verb}  
         & \textsc{Verb\_Sem} 
         &  \texttt{PERS} = \{1,2,3\}; \\
         &&\texttt{NUM} = \{s,d,p\}; \\
         &&\texttt{GEN} = \{m,f\}; \\
         &&\texttt{TEN} = \{perf, imperf, subj, juss, conv\}\\
         & \textsc{Verb\_Afro} 
         &  \texttt{PERS} = \{1,2,3\}; \\
         &&\texttt{NUM} = \{s,d,p\}; \\
         &&\texttt{GEN} = \{m,f\}; \\
         &&\texttt{TEN} = \{perf, imperf, subj\}\\
         & \textsc{Verb\_Ger} 
         &  \texttt{PERS} = \{1,2,3\}; \\
         &&\texttt{NUM} = \{s,d,p\}; \\
         &&\texttt{GEN} = \{m,f\}; \\
         &&\texttt{TEN} = \{pres, pret\}\\
         & \textsc{Verb\_Rom} 
         &  \texttt{PERS} = \{1,2,3\}; \\
         &&\texttt{NUM} = \{s,p\}; \\
         &&\texttt{GEN} = \{m,f\}; \\
         &&\texttt{TEN} = \{pres, imperf, perf, fut\}\\
    \end{tabular}
    \caption{Summary of the features used in each data set. Abbreviations are explained in \autoref{tab:abbreviation_tables}}
    \label{tab:appendix_overview_dataset}
\end{table*}

\begin{table*}[h!]
    \centering
    \small
    \begin{tabular}{c|c}
    & \texttt{PERS}\\
    \hline
        1 & 1st person\\
        2 & 2nd person\\
        3 & 3rd person\\
        \hline
        12 & 1st person inclusive\\
        \hline
        2v & 2nd person familiar\\
        2h & 2nd person formal\\
        2j & 2n person very formal\\
        \hline
        r & neutral-reflexive\\
        \hline
        3e & 3rd person proximal\\
        3x & 3rd person distal\\
        3w & 3rd person remote\\
        \multicolumn{2}{c}{}\\
    \end{tabular}
    $~~$
    \begin{tabular}{c|c}
    & \texttt{PERS}\\
    \hline
        3ev & 3rd person proximal familiar\\
        3eh & 3rd person proximal formal\\
        3ej & 3rd person proximal very formal\\
        3xv & 3rd person distal familiar\\
        3xh & 3rd person distal formal\\
        3xj & 3rd person distal very formal\\
        3wv & 3rd person remote familiar\\
        3wh & 3rd person remote formal\\
        3wj & 3rd person remote very formal\\
        \multicolumn{2}{c}{}\\
        \multicolumn{2}{c}{}\\
        \multicolumn{2}{c}{}\\
    \end{tabular}
    $~~~~~$
    \begin{tabular}{c|c}
    & \texttt{NUM}\\
    \hline
        s & singular\\
        d & dual\\
        p & plural\\
        \multicolumn{2}{c}{}\\
        & GEN\\
    \hline
        m & masculine\\
        f & feminine\\
        n & neutral\\
    \hline
        v & virile / animate\\
        i & non-virile / inanimate\\
        \hline
        c & common\\
        \multicolumn{2}{c}{}\\
    \end{tabular}
    
    \begin{tabular}{c|c}
    & \texttt{TEN}\\
    \hline
        perf / V\tablefootnote{V, G, S, J, C are the representations we use in the model} & perfective\\
        imperf / G & imperfective\\
        subj / S & subjunctive\\
        juss / J & jussive\\
        conv / C & converb\\
        \hline
        pres / G & present\\
        pret / V & preterite\\
        \hline
        fut / S & future\\
        \multicolumn{2}{c}{}\\
    \end{tabular}
    $~~~~~$
    \begin{tabular}{c|c}
    & \texttt{CASE}\\
    \hline
        nom & nominative\\
        gen & genitive\\
        dat & dative\\
        akk & accusative\\
        disj & disjunctive\\
        instr & instrumental\\
        loc & locative\\
        abl & ablative\\
        con & con-version\tablefootnote{special version used with the pronoun con/cum/com in many Romanic languages.}\\
    \end{tabular}
    $~~$
    \begin{tabular}{c|c}
    & \texttt{CASE}\\
    \hline
        akk-l & long accusative\\
        dat-l & long dative\\
        gen-s & short genitive\\
        dat-s & short dative\\
        akk-s & short accusative\\
        obl & oblique\\
        erg & ergative\\
        indir & indirect\\
        obj & object\\
    \end{tabular}
    $~~$
    \begin{tabular}{c|c}
    & \texttt{CASE}\\
    \hline
        alla & allative\\
        equa & equative\\
        simi & similative\\
        comm & comitative\\
        prol & prolative\\
        dir & directive\\
        \multicolumn{2}{c}{}\\
        \multicolumn{2}{c}{}\\
        \multicolumn{2}{c}{}\\
    \end{tabular}
    \caption{Feature Abbreviations}
    \label{tab:abbreviation_tables}
\end{table*}

\section{Modeling Details} \label{app:modeling_details}

\subsection{Details for Estimating Need Probabilities}\label{app:need-probabilities}

\begin{figure}
    \centering
    \includegraphics[width=0.8\linewidth]{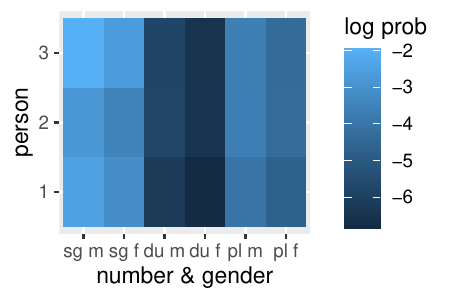}
    \caption{Estimated need probability for person-number-gender combinations
    , obtained by combining need probabilities estimated by \citet{zaslavsky2021pronouns} with corpus counts for gender and the dual/plural distinction.
    These collapse the inclusive/exclusive distinction; we keep that distinction (with relative frequencies from \citet{zaslavsky-etal-2021-rate}) in those paradigms where it is relevant.
    See \autoref{app:need-probabilities} for details.
    }
    \label{fig:heatmap}
    \reproducibilityNote{
        - Run frequency_heatmap.R.
        - The plot is in <OUTPUT_DIR>/FREQ_heatmap.pdf.
        - The output directory can be set in the first lines (OUTPUT_DIR <- "[...]")
        - frequencies input file has to contain tab-separated features and lie in <INPUT_DIR>/calculated_weights_0_DEF.tsv
        - The input directory can be set in the first lines (INPUT_DIR <- "[...]")
    }
\end{figure}

Efficient communication models require a communicative need distribution, i.e. the prior term $p_{cog}(t)$ in \autoref{eq:Ienc}.
For simple pronoun paradigms (\textsc{ppd}), we used the need distribution that \citet{zaslavsky2021pronouns} had estimated from various corpora. 
For the more complex pronoun (\textsc{pron}) and verb (\textsc{verb}) paradigms, we started from these distributions, treating these as marginal distributions of each combination of a person and binary number (singular vs dual/plural).
We then obtained relative frequencies of dual versus plural and masculine vs feminine from the NUDAR Treebank \citep{taji-etal-2017-universal}\footnote{\url{https://github.com/UniversalDependencies/UD_Arabic-NYUAD}}, a large treebank for a language (Standard Arabic) that marks all distinctions relevant to the Semitic paradigms experiment.
We collapsed the frequencies of those forms that are not distinguished in the language, and assumed equal distribution for further features (such as tense and case).
Resulting probabilities for \textsc{verb} are shown in \autoref{tab:calc_prob_table} and \autoref{fig:heatmap}.

\begin{table}[h]
\scriptsize
    \centering
    \begin{tabular}{l|r}
        1s m G & 16.1149\\
        1s ~f~ G & 8.0371\\
        2s m G & 10.8485\\
        2s ~f~ G & 5.4090\\
        3s m G & 25.7411\\
        3s ~f~ G & 12.8247\\
    \end{tabular} 
    \begin{tabular}{l|r}
        1p m G & 3.168\\
        1p ~f~ G & 1.5789\\
        2p m G & 4.7571\\
        2p ~f~ G & 2.3734\\
        3p m G & 4.6043\\
        3p ~f~ G & 2.2919\\
    \end{tabular} 
    \begin{tabular}{l|r}
        1d m G & 0.3769\\
        1d ~f~ G & 0.1935\\
        2d m G & 0.5704\\
        2d ~f~ G & 0.2852\\
        3d m G & 0.5501\\
        3d ~f~ G & 0.2750\\
    \end{tabular} \\
    $~$\\$~$\\
    \begin{tabular}{l|r}
        1s m V & 16.1149\\
        1s ~f~ V & 8.0371\\
        2s m V & 10.8485\\
        2s ~f~ V & 5.4090\\
        3s m V & 25.7411\\
        3s ~f~ V & 12.8247\\
    \end{tabular} 
    \begin{tabular}{l|r}
        1p m V & 3.168\\
        1p ~f~ V & 1.5789\\
        2p m V & 4.7571\\
        2p ~f~ V & 2.3734\\
        3p m V & 4.6043\\
        3p ~f~ V & 2.2919\\
    \end{tabular} 
    \begin{tabular}{l|r}
        1d m V & 0.3769\\
        1d ~f~ V & 0.1935\\
        2d m V & 0.5704\\
        2d ~f~ V & 0.2852\\
        3d m V & 0.5501\\
        3d ~f~ V & 0.2750\\
    \end{tabular}
    \caption{Calculated Feature Probabilities. G stands for imperfective, V for perfective.}
    \label{tab:calc_prob_table}
\end{table}

\paragraph{Discussion}

An alternative would have been to obtain counts for all entries in the paradigm table; however, we opted against this as the written genres (e.g., newspaper text) represented in the available treebank data show overwhelming biases (e.g., towards third-person forms) unlikely to be representative of spoken language.
An ideal solution would be to obtain counts for large conversational corpora of the relevant languages; however, such data is not available with suitable annotation at the required scale.
We show the robustness of our results to other distributions in \autoref{app:robustness}, where we evaluated with a uniform and three different randomly generated need distributions.

\subsection{Robustness towards Frequency Distributions}
\label{app:robustness}
\begin{figure*}[hbt]
    \centering
    
    \begin{subfigure}{0.49\textwidth}
        \caption{Uniform}
        \includegraphics[width=0.9\linewidth]{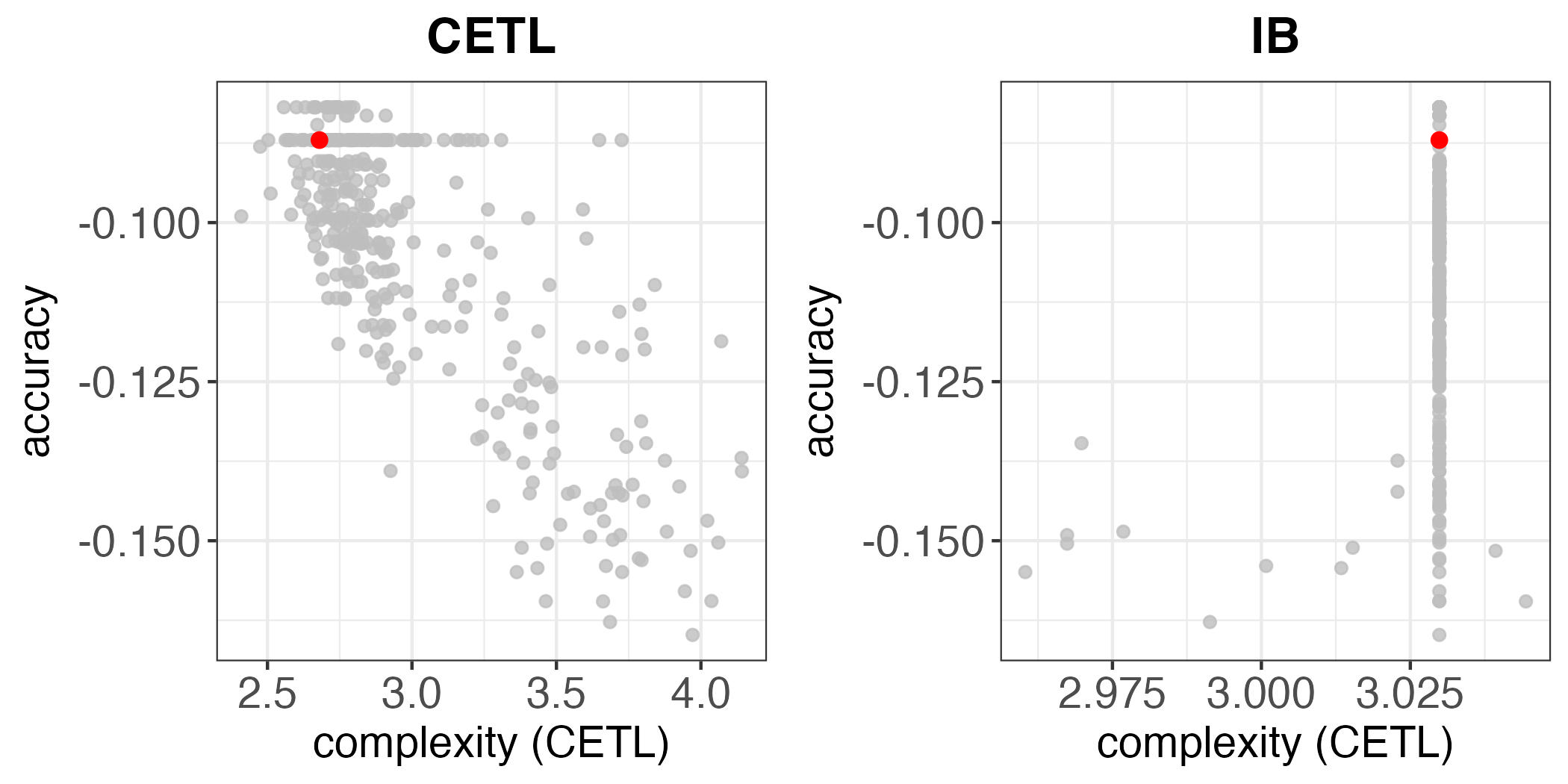}
    \end{subfigure}
    \begin{subfigure}{0.49\textwidth}
        \caption{Random1}
        \includegraphics[width=0.9\linewidth]{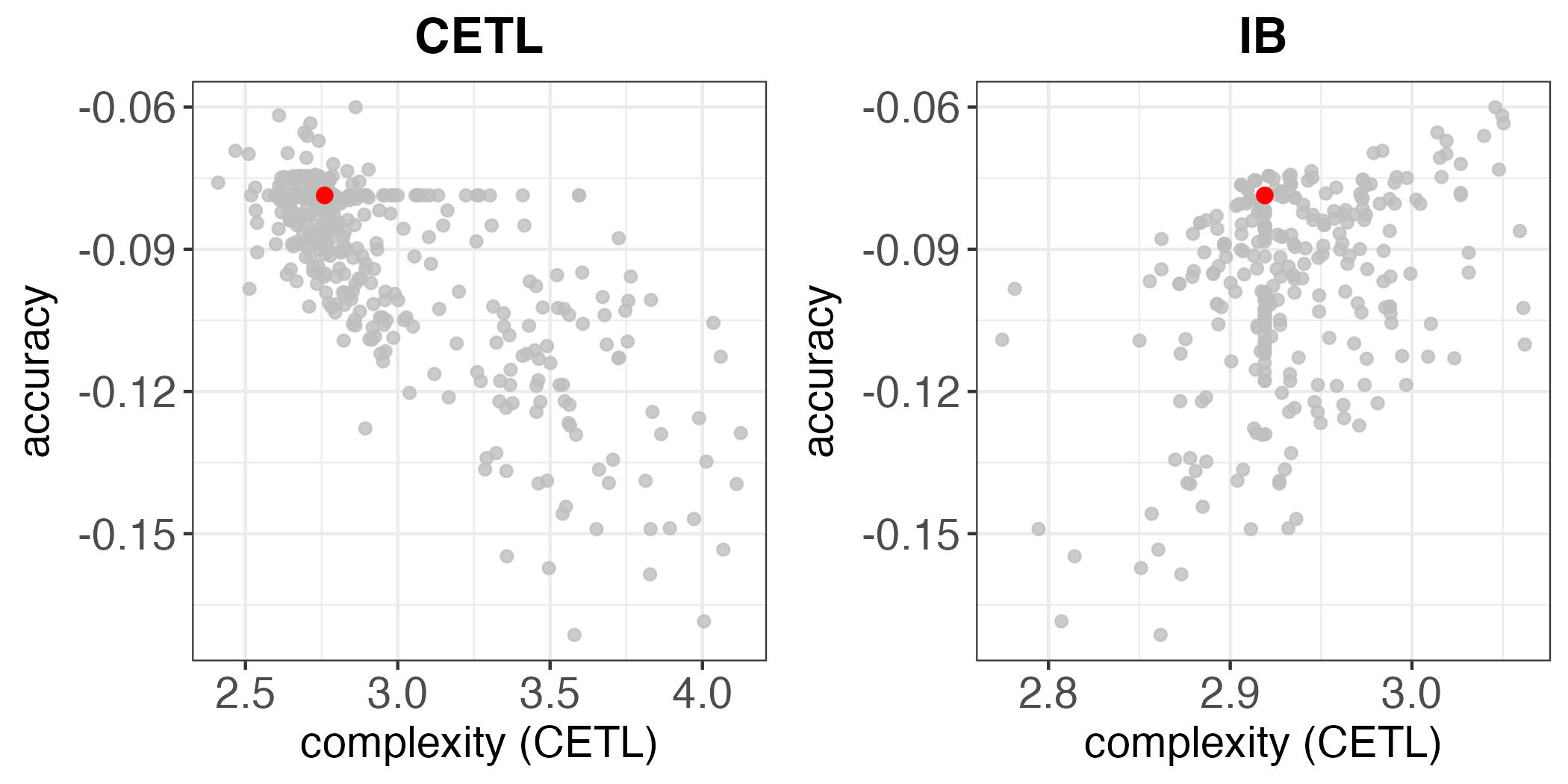}
    \end{subfigure}
    
    \begin{subfigure}{0.49\textwidth}
        \caption{Random2}
        \includegraphics[width=0.9\linewidth]{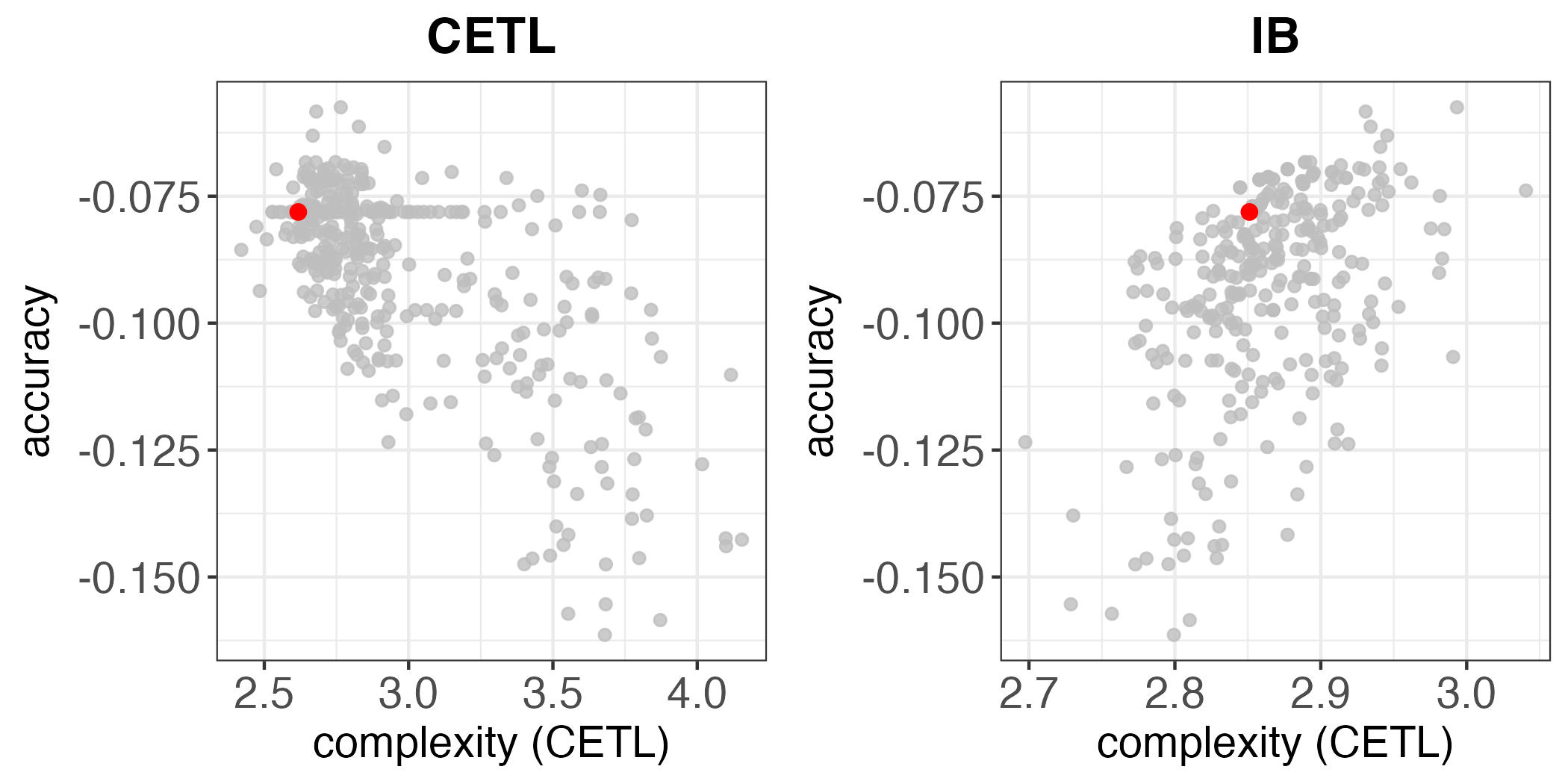}
    \end{subfigure}
    \begin{subfigure}{0.49\textwidth}
        \caption{Random3}
        \includegraphics[width=0.9\linewidth]{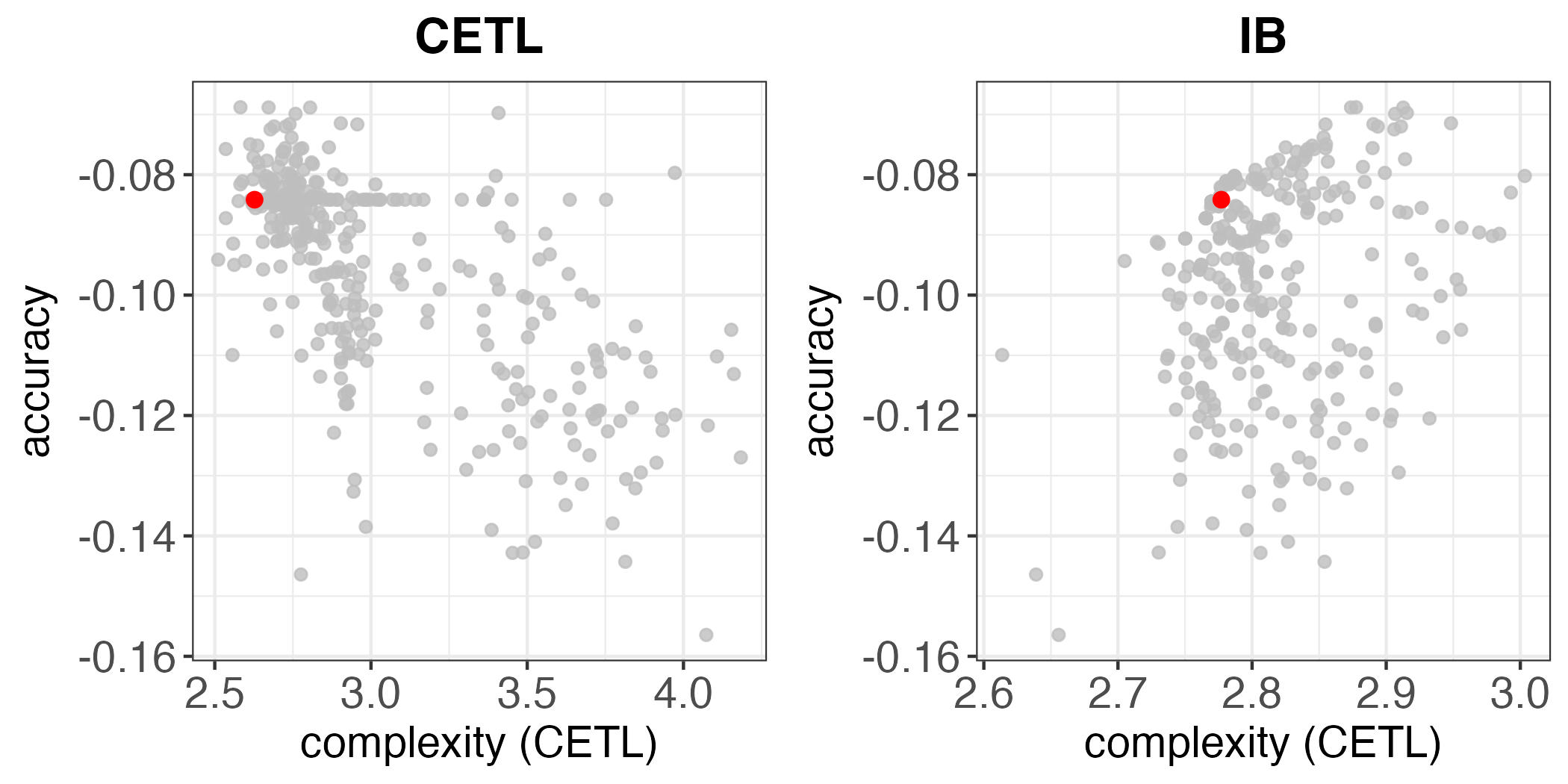}
    \end{subfigure}
    \caption{Results for \aref{app:robustness}: Efficiency of permutations: Accuracy plotted (inverted) against complexity measures for a uniform distribution and three random distributions of feature frequencies.}
    \label{fig:rob_EFF}
 \reproducibilityNote{
        - Run robustness.R.
        - The plot is in <OUTPUT_DIR>/EFF_rob_ALL.pdf.
        - The output directory can be set in the first lines (OUTPUT_DIR <- "[...]")
    }
\end{figure*}

\begin{figure*}[hbt]
    \centering
    \begin{subfigure}{1\textwidth}
    \centering
        \caption{Uniform}
        \includegraphics[width=0.9\linewidth]{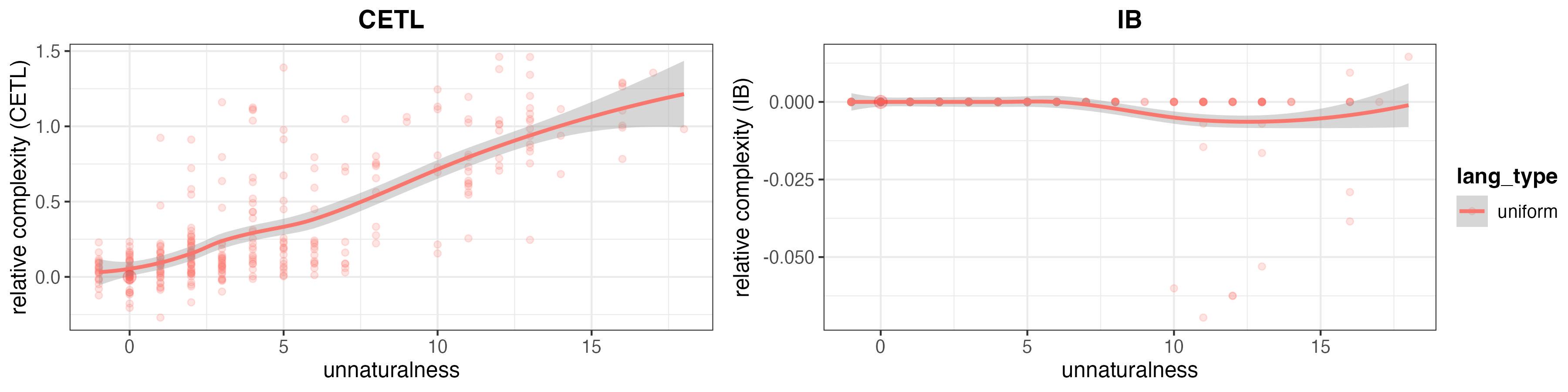}
    \end{subfigure}
    \begin{subfigure}{1\textwidth}
    \centering
        \caption{Random1}
        \includegraphics[width=0.9\linewidth]{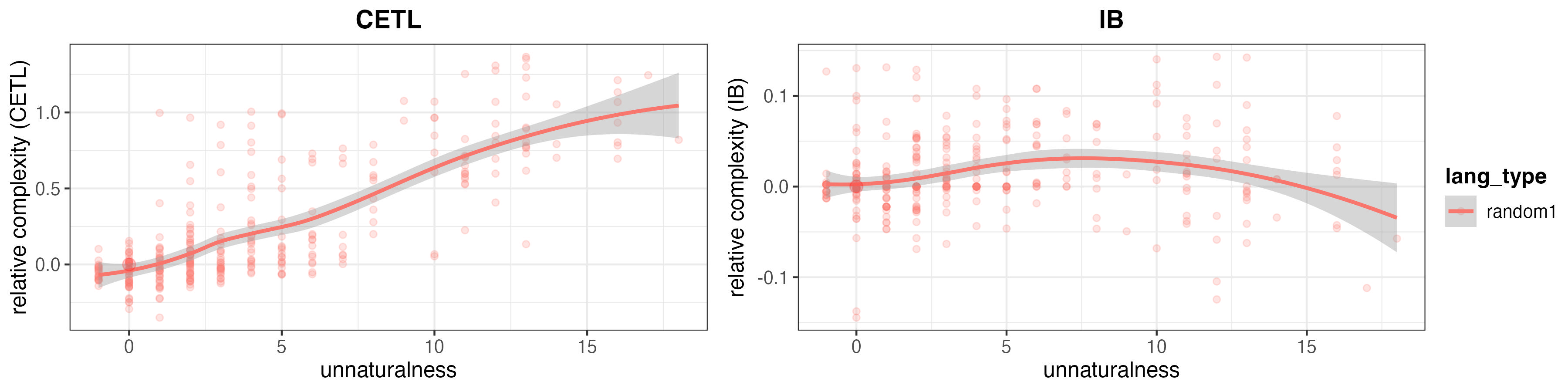}
    \end{subfigure}
    \begin{subfigure}{1\textwidth}
    \centering
        \caption{Random2}
        \includegraphics[width=0.9\linewidth]{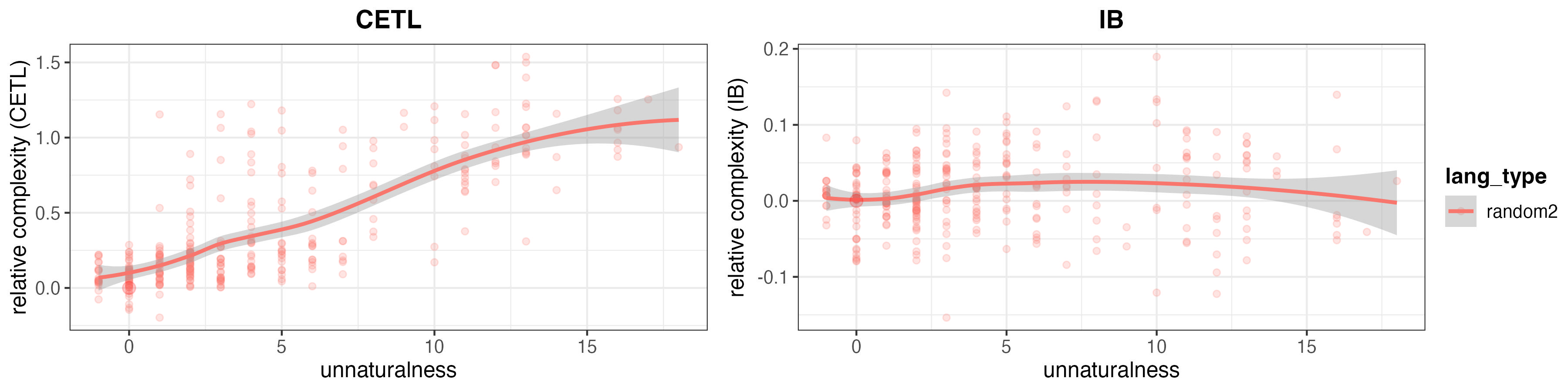}
    \end{subfigure}
    
    \begin{subfigure}{1\textwidth}
    \centering
        \caption{Random3}
        \includegraphics[width=0.9\linewidth]{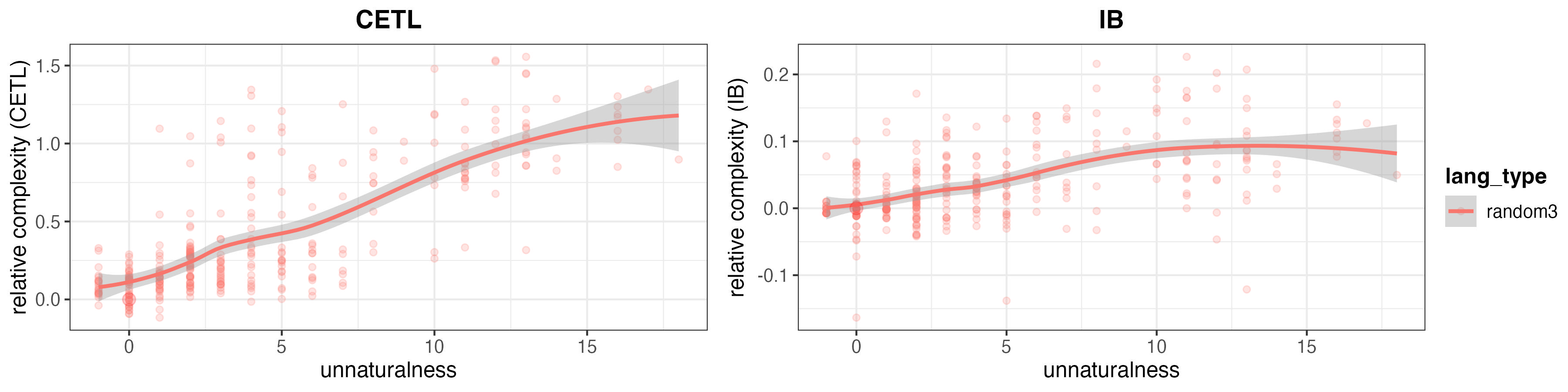}
    \end{subfigure}
    \caption{Results for \aref{app:robustness}: Complexity of permutations of different naturalness: Complexity measures plotted against unnaturalness for a uniform distribution and three random distributions of feature frequencies.}
    \label{fig:rob_NAT}
 \reproducibilityNote{
        - Run robustness.R.
        - The plot is in <OUTPUT_DIR>/NAT_rob_ALL.pdf.
        - The output directory can be set in the first lines (OUTPUT_DIR <- "[...]")
    }
\end{figure*}

We create a uniform distribution and 3 different random distributions of frequencies, and test them on the Classical Arabic verbal paradigm, to verify that our results are robust to the choice of frequency distributions.
\autoref{fig:rob_EFF} shows the efficiency plots, \autoref{fig:rob_NAT} and \autoref{tab:correlation_unifrand} show the correlation results. We see a stably strong correlation for CETL across the different frequency distributions.
We note that the need distribution may still play a substantial role in efficient coding (e.g. in accounting for the detailed typological patterns studied by \citet{zaslavsky2021pronouns}), but these analyses show that our results are robust to the choice of the need distribution.

\autoref{tab:hitfailrates_robustness} lists the hit and fail rates. The need distribution does have an effect on the performance of the models. For one of the random distributions the overall performance is slightly better for MI, for the other distributions, as well as the corpus-based distribution explained in the main part of this paper, CETL performs much better than MI.

\begin{table*}[h]
    \centering
    \small
    \begin{tabular}{c||cc|c}
         &\multicolumn{2}{c|}{correlation (avg)} &support\\
         & \texttt{CETL} & \texttt{IB} &  \\
         \hline\hline
      uniform &0.78& -0.26&\tdg{318}\\
random1 &0.76  &0.13&\tdg{318}\\
random2 &0.75 &-0.05&\tdg{318}\\
random3 &0.74  &0.54&\tdg{318}\\
    \end{tabular}
    \caption{Results for \aref{app:robustness}: Correlation between complexity and unnaturalness for a uniform distribution and three random distributions.}
    \label{tab:correlation_unifrand}
  \reproducibilityNote{
        - Run robustness.R.
        - The table is printed to the console.
  }
\end{table*}

\begin{table*}[t]
\centering
\small
\setlength{\tabcolsep}{3pt}
    
    \begin{subtable}[t]{0.49\linewidth}
    \centering
    \begin{tabular}{c|rr|rr|rr|r}
    & \multicolumn{2}{c|}{$C_M$ (\%)} & \multicolumn{2}{c|}{$I_M$ (\%)} & \multicolumn{2}{c|}{$\mathsf{Perf_M}$}\\
   Exp &  $\mathsf{CETL}$ & $\mathsf{IB}$ & $\mathsf{CETL}$ & $\mathsf{IB}$ & $\mathsf{CETL}$ & $\mathsf{IB}$ & support\\
    \hline
    Unif   &  \tgr{82.50} & 76.25   &   \tred{6.56} &  20.31  &  \tbl{75.94} & 55.94  &   \tgray{320}\\
    Rand1  &  \tgr{59.43} & 49.06   &  15.09 &   \tred{4.09}  &  44.34 & \tbl{44.97}  &   \tgray{318}\\
    Rand2  &  \tgr{68.77} & 32.81   &   \tred{1.89} &   3.15  &  \tbl{66.88} & 29.65  &   \tgray{317}\\
    Rand3  &  \tgr{75.16} & 47.77   &   2.23 &   \tred{1.59}  &  \tbl{72.93} & 46.18  &   \tgray{314}\\
    \end{tabular}
    \caption{Structural Permutations}
    \label{tab:hitfailrates_PERM}
    \end{subtable}
    \begin{subtable}[t]{0.49\linewidth}
     \centering
    \begin{tabular}{c|rr|rr|rr|r}
    & \multicolumn{2}{c|}{$C_M$ (\%)} & \multicolumn{2}{c|}{$I_M$ (\%)} & \multicolumn{2}{c|}{$\mathsf{Perf_M}$}\\
   Exp &  $\mathsf{CETL}$ & $\mathsf{IB}$ & $\mathsf{CETL}$ & $\mathsf{IB}$ & $\mathsf{CETL}$ & $\mathsf{IB}$ & support\\
    \hline
    Unif   &  \tgr{81.13}    &  0   &  \tred{18.87}   &  100  &  \tbl{62.26}  & -100   &   \tgray{53}\\
    Rand1  &  \tgr{54.72}    &  0   &  \tred{45.28}   &  100  &   \tbl{9.43}  & -100   &   \tgray{53}\\
    Rand2  &  \tgr{86.54}    &  0   &  \tred{13.46}   &  100  &  \tbl{73.08}  & -100   &   \tgray{52}\\
    Rand3  &  \tgr{96.49}    &  0   &   \tred{3.51}   &  100  &  \tbl{92.98}  & -100   &   \tgray{57}\\
    \end{tabular}
    \caption{Form-only Permutations}
    \label{tab:hitfailrates_SHUF}
    \end{subtable}
    \reproducibilityNote{
        - Run robustness.R.
        - The tables are printed to the console.
  }
    \caption{Results for \aref{app:robustness}: Hit and Fail rates for CETL vs.~MI for structural permutations (left) and form-only permutations (right) for a uniform and three random distributions of feature frequencies. 
    }
    \label{tab:hitfailrates_robustness}
\end{table*}

\subsection{Comparison using Feature Representation and Weighting from \citet{zaslavsky2021pronouns}}
\label{app:direct_comparison}

Recall from \aref{app:feat_repr_APP} that meanings of referents $t$ are formalized as probability distributions:
\begin{equation}\label{eq:exponential-repeat}
m_t(u) \propto \exp(-\gamma\cdot d(u,t)).
\end{equation}
where $d(u,t)$ is a weighted Hamming distance based on feature encodings of the individual references.
As described in that section, \citet{zaslavsky2021pronouns} defined $d(\cdot,\cdot)$ in terms of a weighted 5-bit vector feature representation with specific weights, whereas we  use a more generic discrete feature vector space for our analyses that easily accommodates further features (e.g., gender, tense)  even when they are not binary.

\begin{figure*}
    \centering
    \begin{subfigure}{1\textwidth}
    \centering
        \caption{$\gamma = 1$}
        \includegraphics[width=0.75\linewidth]{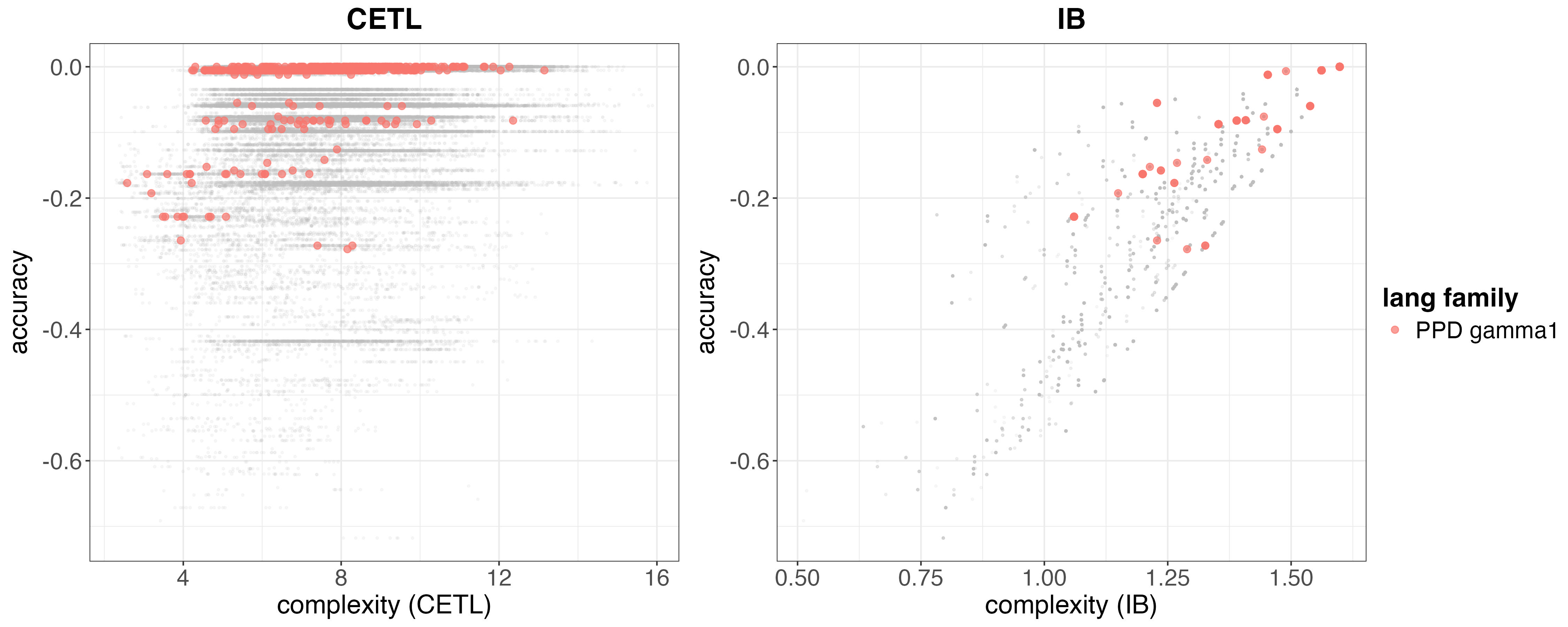}
    \end{subfigure}
    \begin{subfigure}{1\textwidth}
    \centering
        \caption{$\gamma = 2$}
        \includegraphics[width=0.75\linewidth]{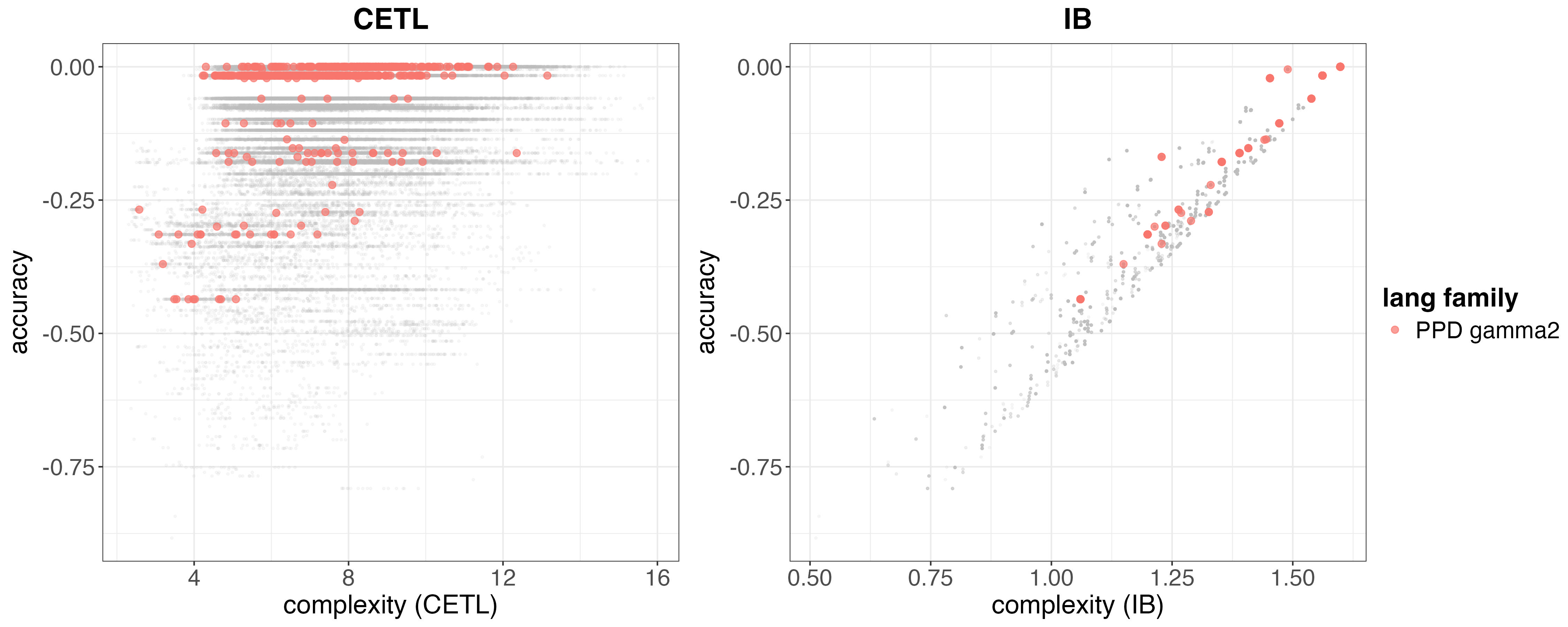}
    \end{subfigure}
    \begin{subfigure}{1\textwidth}
    \centering
        \caption{$\gamma = 5$}
        \includegraphics[width=0.75\linewidth]{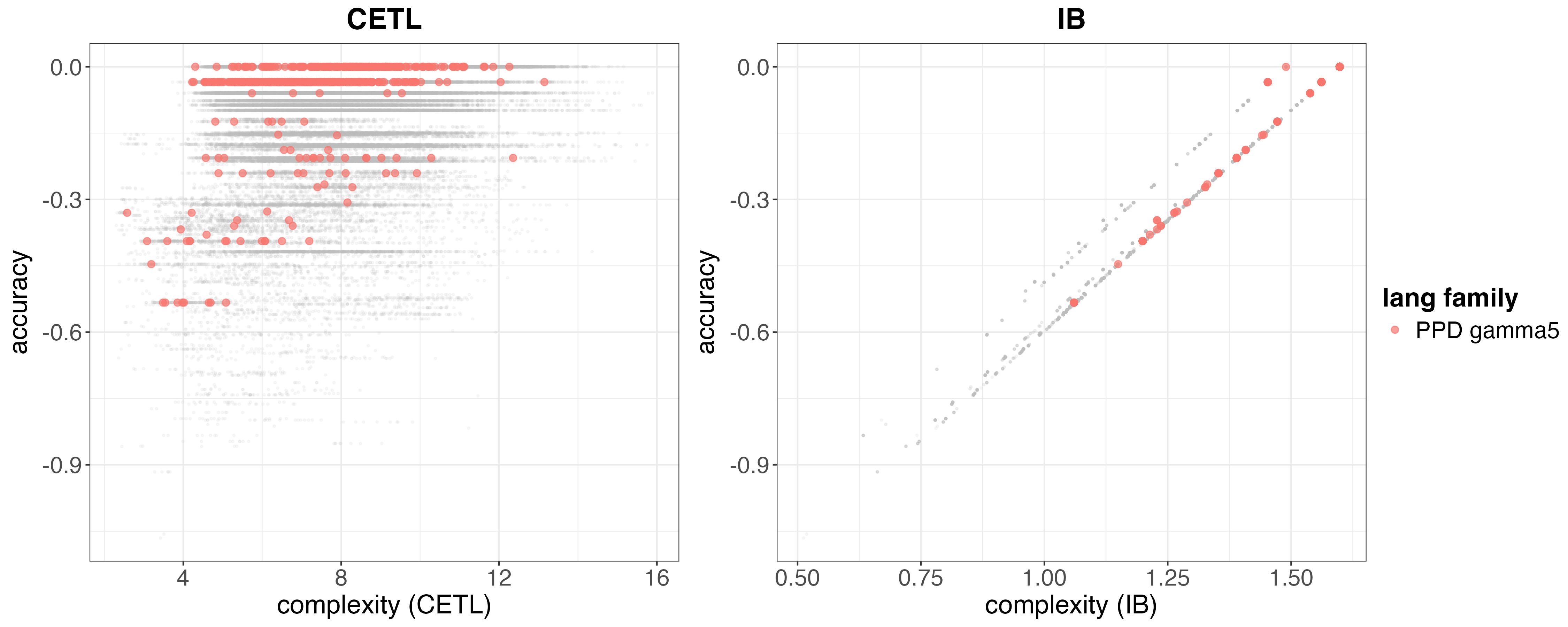}
    \end{subfigure}
    \caption{Efficiency of permutations. Accuracy plotted against complexity measures, using \citet{zaslavsky2021pronouns}'s feature representation, with $\gamma=1$ (a), $\gamma=2$ (b) and $\gamma=5$ (c). Here, we used the feature representations from \citet{zaslavsky2021pronouns}, with their feature weights. The model has a free parameter $\gamma$; we show results for three values.}
    \label{fig:direct_comparisonAPP}
     \reproducibilityNote{
        - Run direct_comparison.R.
        - The plot is in <OUTPUT_DIR>/EFF_PRONCogSci21gamma1.pdf, EFF_PRONCogSci21gamma2.pdf, EFF_PRONCogSci21gamma5.pdf.
        - The output directory can be set in the first lines (OUTPUT_DIR <- "[...]")
    }
\end{figure*}

\begin{figure*}
    \begin{subfigure}{0.49\textwidth}
        \centering
        \includegraphics[width=0.9\linewidth]{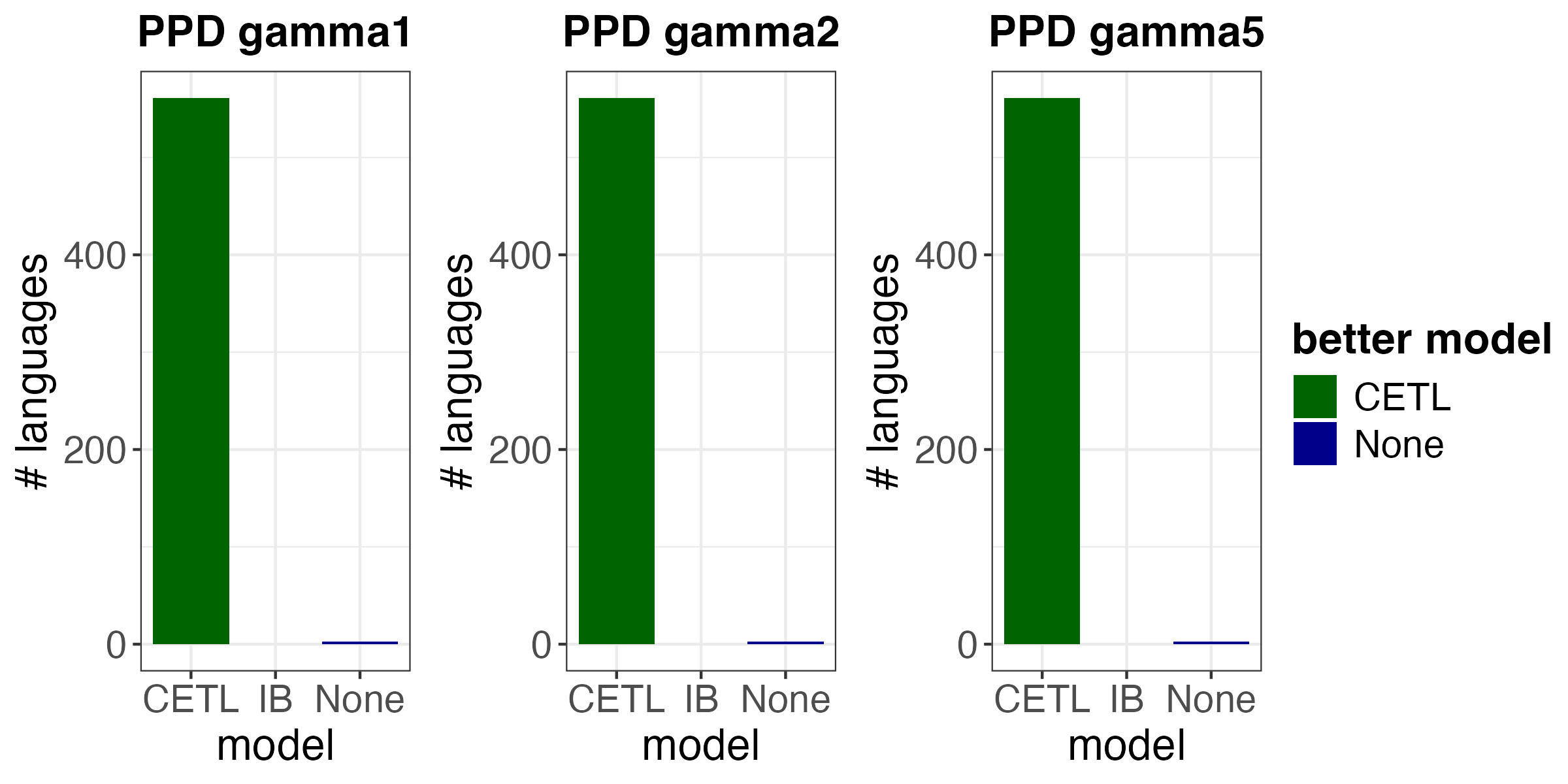}
        \caption{Overall}
    \end{subfigure}
    \begin{subfigure}{0.49\textwidth}
        \centering
        \includegraphics[width=0.9\linewidth]{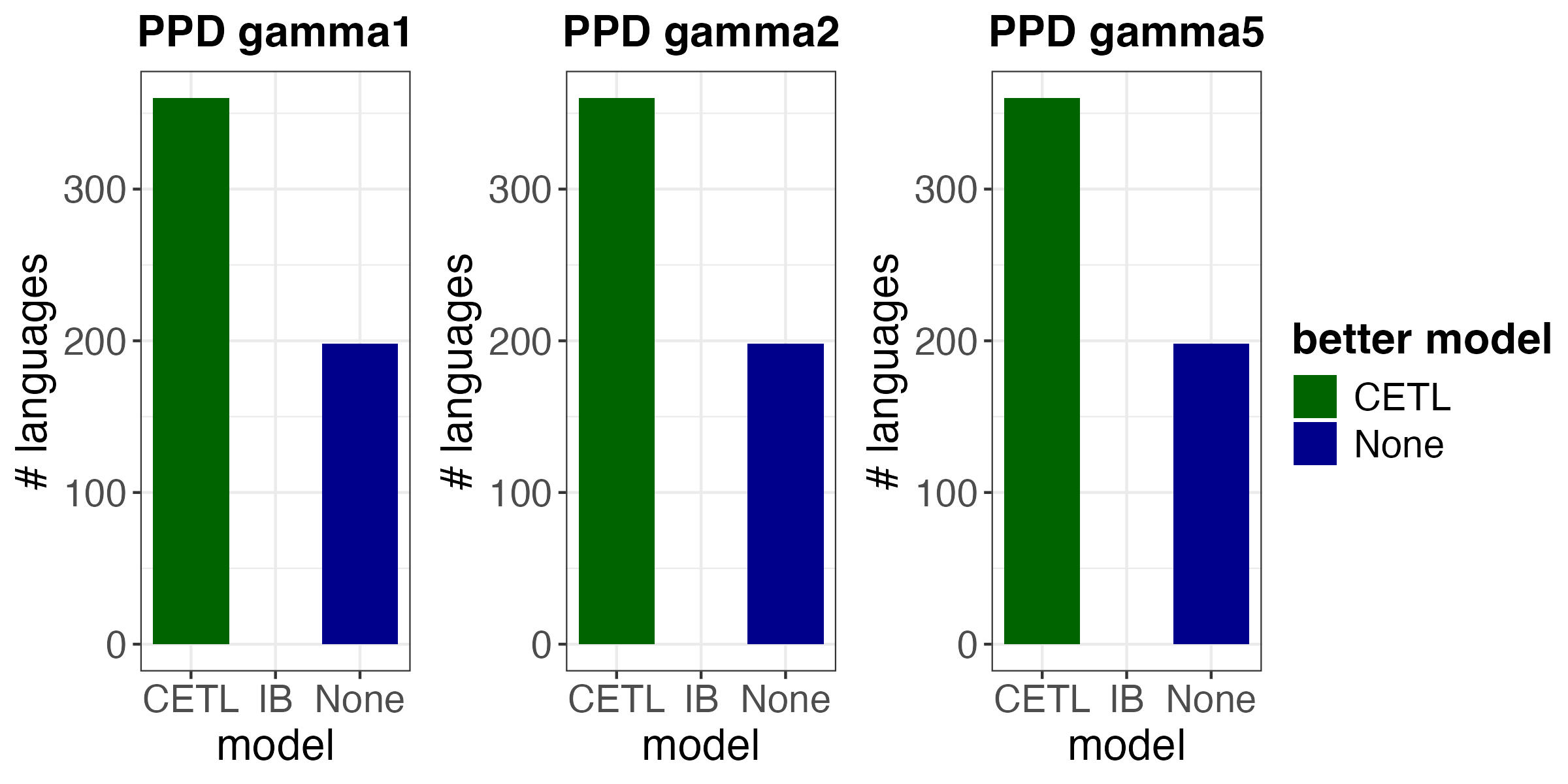}
        \caption{Structural Permutations}
    \end{subfigure}
    \caption{Performance of the CETL vs.~the IB model using \citet{zaslavsky2021pronouns}'s feature representation: Amount of languages for which each model outperforms the other by at least 5\% of the total permutations.}
    \label{fig:direct_comparisonAPP-perf}
    \reproducibilityNote{
        - Run direct_comparison.R.
        - The plot is in <OUTPUT_DIR>/HIT_FAIL_CogSci21.pdf,HIT_FAIL_CogSci21_PERM.pdf.
        - The output directory can be set in the first lines (OUTPUT_DIR <- "[...]")
    }
\end{figure*}

Here, in the \textsc{ppd} domain, we compare the CETL- and MI-based models directly using the 5-bit-vector feature representation and feature weights used by  \citet{zaslavsky2021pronouns}. 
We used three different values for the free parameter $\gamma$ (\autoref{eq:exponential-repeat}), and compute the efficiencies for our \textsc{ppd} dataset (\autoref{fig:direct_comparisonAPP} and \autoref{fig:direct_comparisonAPP-perf}).
Overall, even in this setup, CETL provides stronger performance, even on structural permutations.
This shows that, while the specific feature weights used by \citet{zaslavsky2021pronouns} may be important in accounting for specific typological patterns, the improved discrimination between real and counterfactual paradigms provided by the CETL-based model is general and robust to the weighting of different features.

\subsection{Robustness towards Paradigm Classes}
\label{app:classes}

\begin{figure*}[hbt]
    \centering
    \begin{subfigure}{\textwidth}
        \centering 
        
        \includegraphics[width=0.9\linewidth]{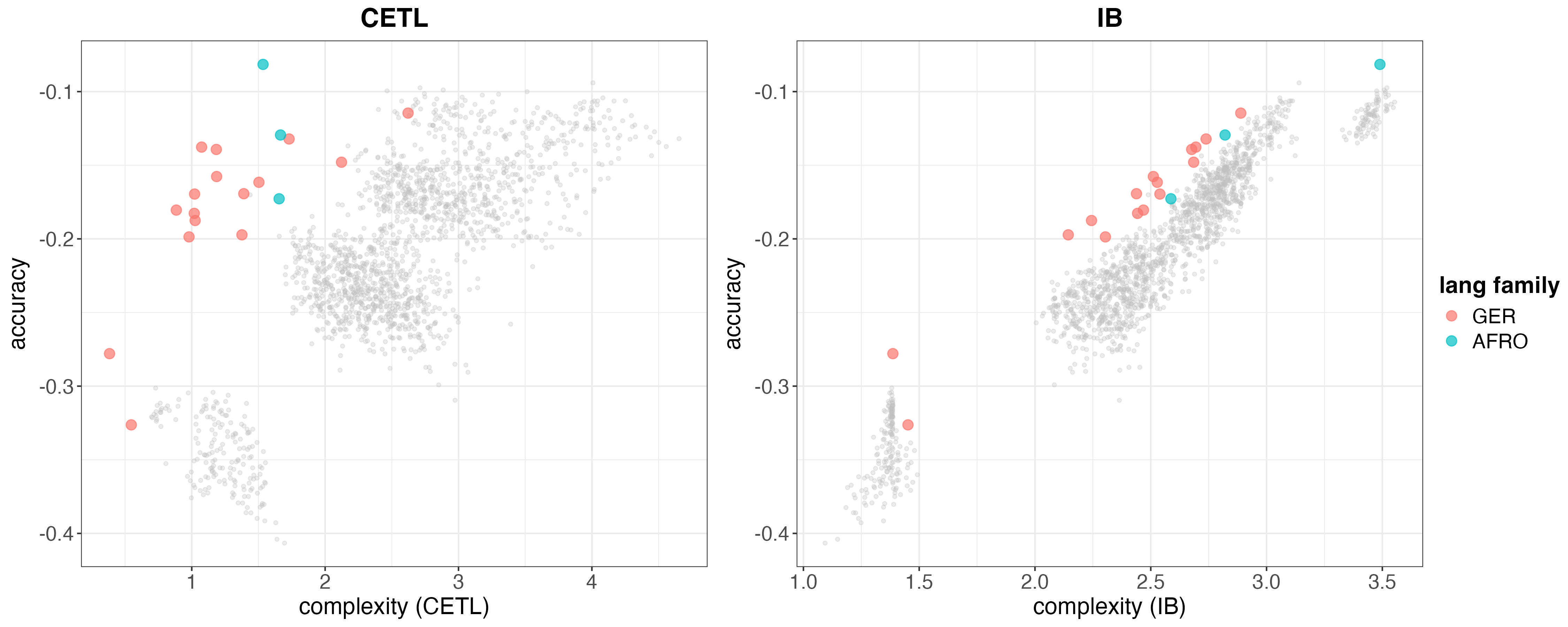}
        \caption{Inside conjugation classes}
    \end{subfigure}
    \begin{subfigure}{\textwidth}
        \centering 
    \includegraphics[width=0.9\linewidth]{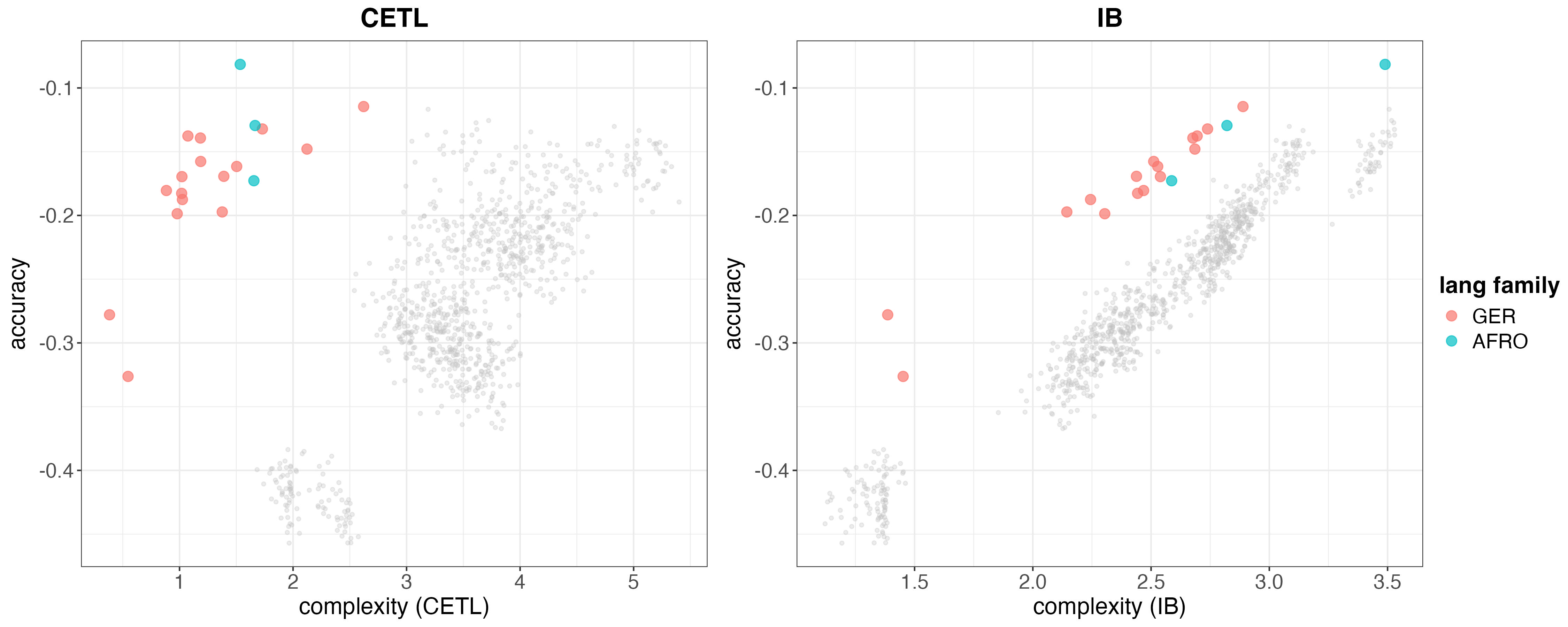}
     \caption{Across conjugation classes}
    \end{subfigure}
    \caption{Results for \aref{app:classes}: Efficiency of permutations. Accuracy plotted against complexity measures for permutations across different conjugation classes (b) and inside permutation classes only (a). Results are shown for Germanic (red) and Cushitic (blue, a subfamily of Afro-Asiatic).}
    \label{fig:classes_EFF}
\end{figure*}

\begin{figure*}[hbt]
    \centering
     \begin{subfigure}{\textwidth}
        \centering \includegraphics[width=0.9\linewidth]{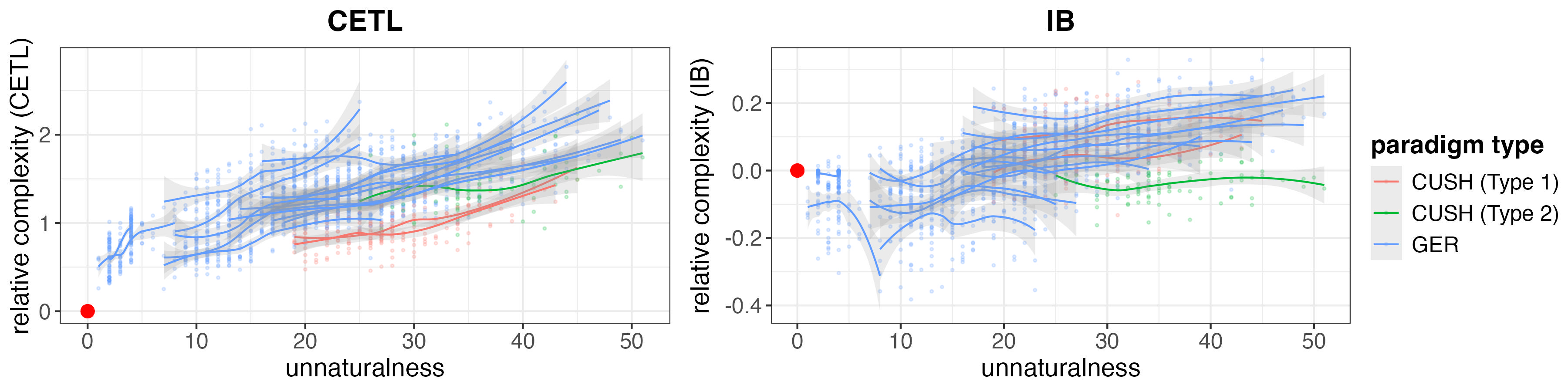}
        \caption{Inside conjugation classes}
    \end{subfigure}
    \begin{subfigure}{\textwidth}
        \centering 
        \includegraphics[width=0.9\linewidth]{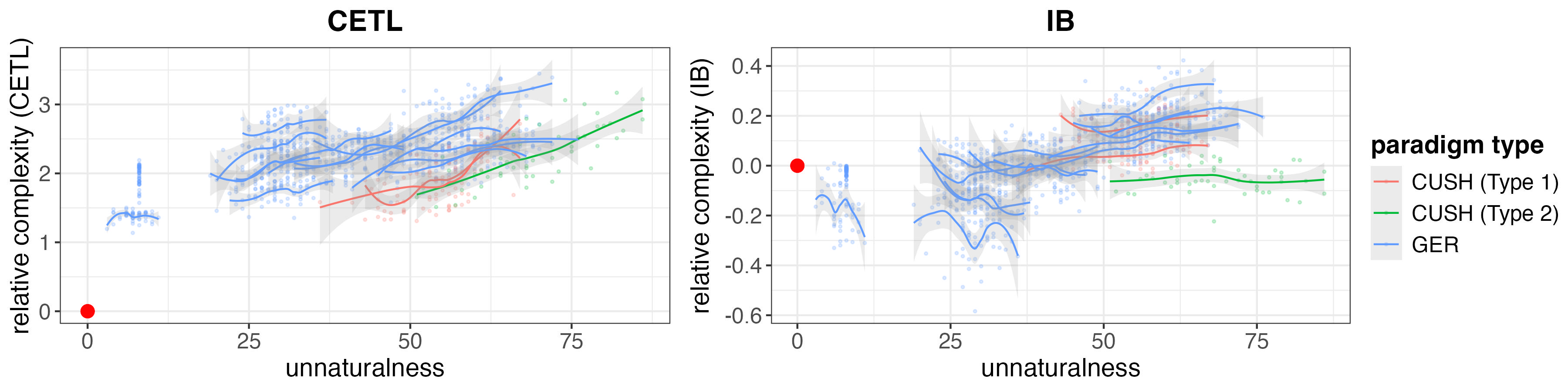}
        \caption{Across conjugation classes}
    \end{subfigure}
    \caption{Results for \aref{app:classes}: Complexity of permutations of different naturalness: Complexity measures plotted against unnaturalness for permutations across different conjugation classes (b) and inside permutation classes only (a).}
    \label{fig:classes_NAT}
\end{figure*}

As discussed in \aref{app:paradigm_representation}, our analyses do not consider variability between different verbs, in order to control for the effect of previously-proposed approaches such as i-complexity \citep{complexity_i_e_def}.
Here, we consider an alternative modeling choice, where we jointly model different paradigms in a language, and show that it leads to qualitatively equivalent results.
We use two language families in which verbs tend to fall into two classes, with very distinct morphology: Germanic, where verbs have a strong and weak conjugation class, and Cushitic, where verbs have a prefix-based and a suffix-based conjugation class.
We extend the meaning space by a fifth dimension \emph{class} for conjugation classes.
$m = \langle number, person, gender, tense, class \rangle$, where $class \in \{weak, strong, pc, sc\}$.
We create counterfactuals which permute inside the conjugation classes only, and ones which permute across the conjugation classes.
\autoref{fig:classes_EFF} shows the efficiency plots. \autoref{fig:classes_NAT} and \autoref{tab:correlation_classes} shows the correlation results.
Attested paradigms are substantially more efficient than most counterfactuals, even when permutations are applied only within conjugation classes.

\begin{figure*}[hbt]
    \centering
    \begin{subfigure}{\textwidth}
        \centering 
        
        \includegraphics[width=0.9\linewidth]{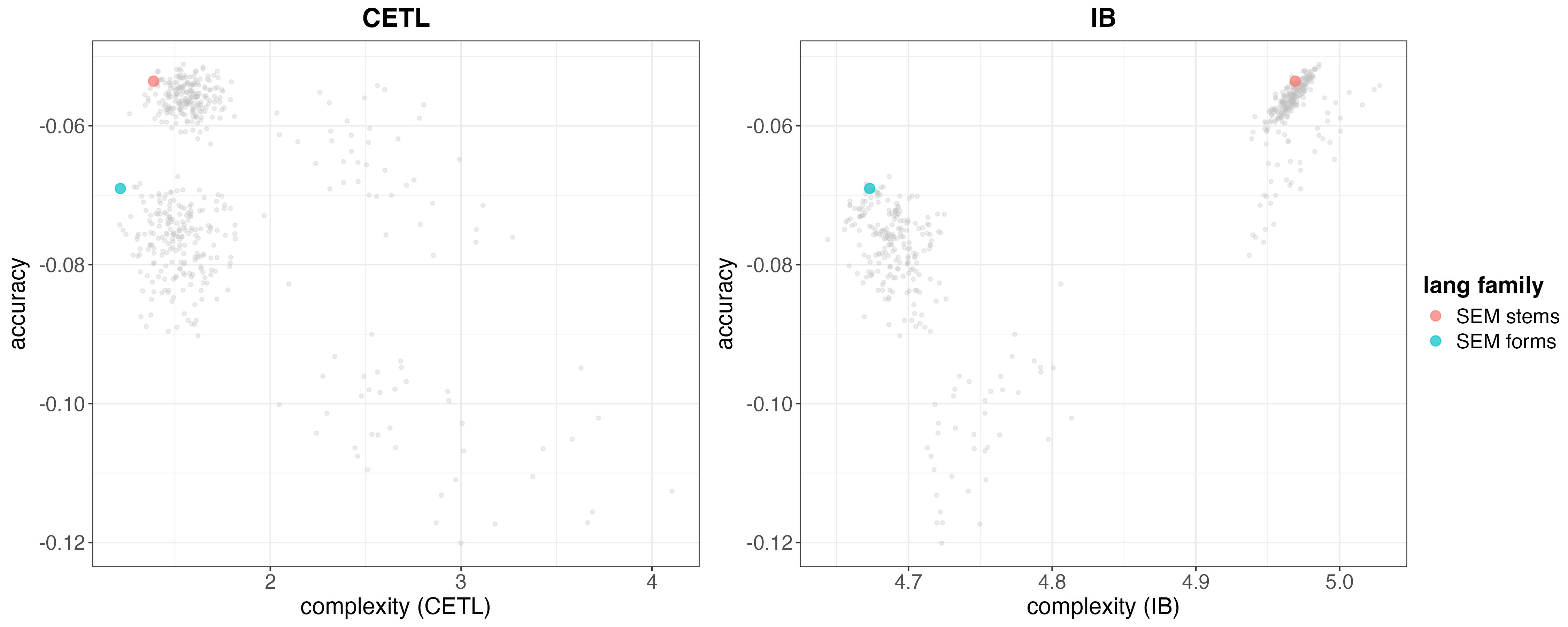}
        \caption{Inside conjugation classes}
    \end{subfigure}
    \begin{subfigure}{\textwidth}
        \centering 
    \includegraphics[width=0.9\linewidth]{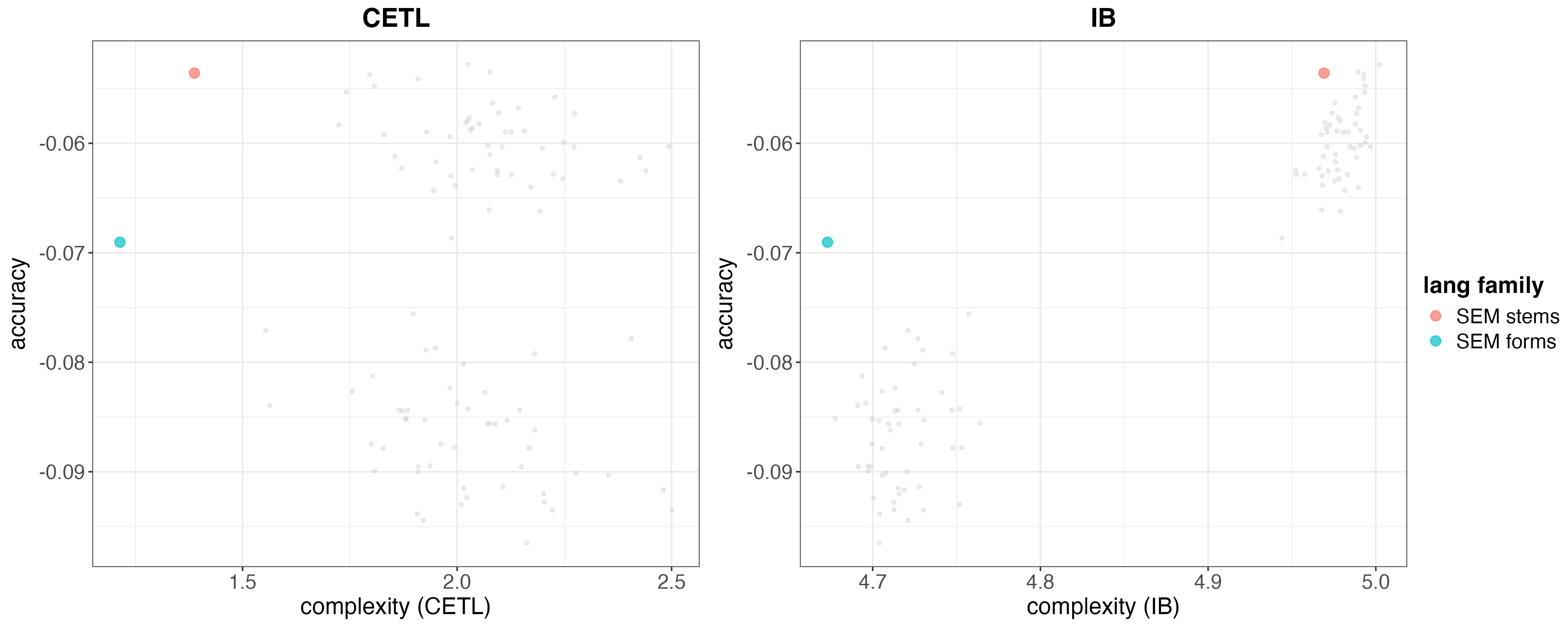}
     \caption{Across conjugation classes}
    \end{subfigure}
    \caption{Results for \aref{app:classes}: Efficiency of permutations. Accuracy plotted against complexity measures for permutations across different conjugation classes (b) and inside permutation classes only (a). Results are shown for Arabic stems (red) and Arabic forms (blue).}
    \label{fig:classesSEM_EFF}
\end{figure*}

\begin{figure*}[hbt]
    \centering
     \begin{subfigure}{\textwidth}
        \centering \includegraphics[width=0.9\linewidth]{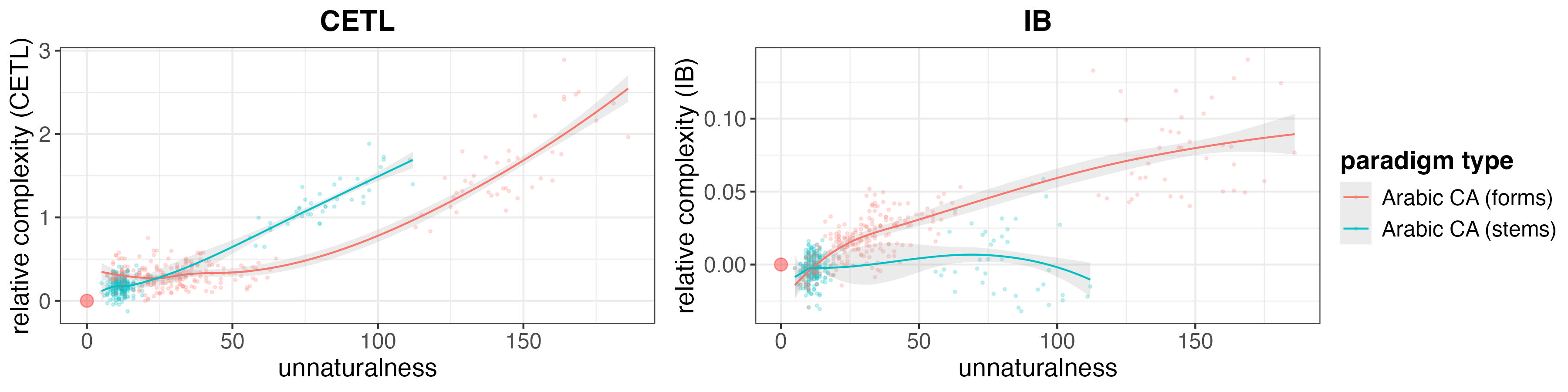}
        \caption{Inside conjugation classes}
    \end{subfigure}
    \begin{subfigure}{\textwidth}
        \centering 
        \includegraphics[width=0.9\linewidth]{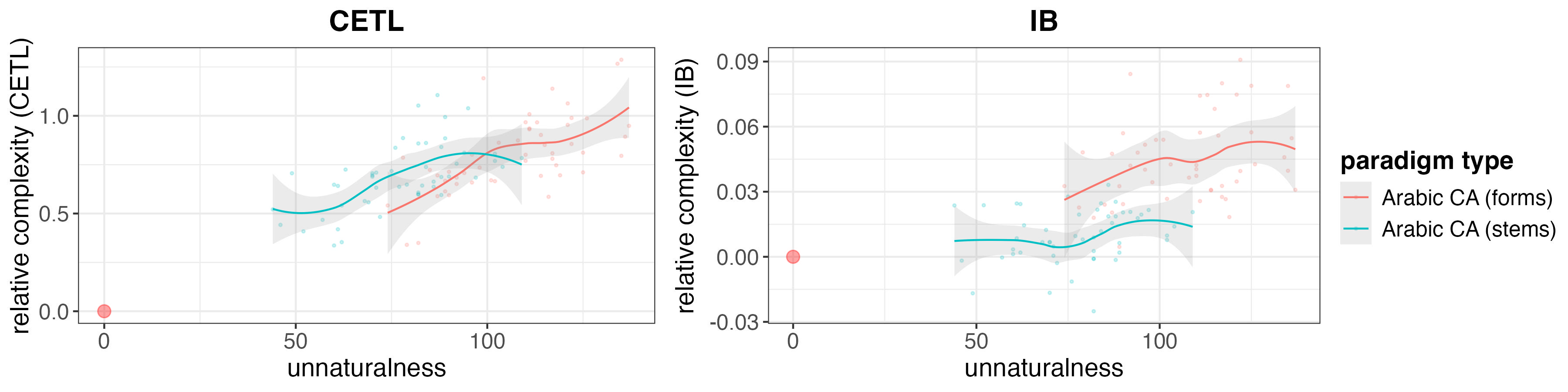}
        \caption{Across conjugation classes}
    \end{subfigure}
    \caption{Results for \aref{app:classes}: Complexity of permutations of different naturalness: Complexity measures plotted against unnaturalness for permutations across different conjugation classes (b) and inside permutation classes only (a). Results are shown for Arabic stems (red) and Arabic forms (blue).}
    \label{fig:classesSEM_NAT}
\end{figure*}

We further create similar paradigms for Classical Arabic.
Semitic verbs are classified into classes based on (i) whether their root consonants contain a glide (\textit{w} or \textit{y}; ``weak verbs'') or no glide (``strong verbs''), and (ii) whether they are directly inflected from a root (``Stem I''), or the result of a derivational process (``Stems II, III, $\dots$, X''). 
We create two different paradigms for Classical Arabic, 
one containing the 10 main stems as conjugation classes, and one containing one strong and different weak forms of Stem I as classes.
\autoref{fig:classesSEM_EFF} shows the efficiency plots. \autoref{fig:classesSEM_NAT} and \autoref{tab:correlation_classesSEM} shows the correlation results.
For both cases, the attested paradigms are more efficient than most counterfactuals.

\begin{table*}[h]
    \centering
    \small
    \begin{tabular}{c||cc|c}
         &\multicolumn{2}{c|}{correlation (avg)} &support\\
         & \texttt{CETL} & \texttt{IB} &  \\
         \hline\hline
CLASS &0.66 & 0.32&\tdg{1620}\\
XROSS &0.49 & 0.09&\tdg{900}\\
    \end{tabular}
    \caption{Results for \aref{app:classes} - Germanic \& Cushitic: Correlation between complexity and unnaturalness for permutations inside the conjugation classes (CLASS) and across the conjugation classes (XROSS).}
    \label{tab:correlation_classes}
  \reproducibilityNote{
        - Run classes.R.
        - The table is printed to the console.
  }
\end{table*}

\begin{table*}[h]
    \centering
    \small
    \begin{tabular}{c||cc|c}
         &\multicolumn{2}{c|}{correlation (avg)} &support\\
         & \texttt{CETL} & \texttt{IB} &  \\
         \hline\hline
CLASS &0.93 & 0.51&\tdg{500}\\
XROSS &0.63 & 0.28&\tdg{100}\\
    \end{tabular}
    \caption{Results for \aref{app:classes} - Arabic stems and forms: Correlation between complexity and unnaturalness for permutations inside the conjugation classes (CLASS) and across the conjugation classes (XROSS).}
    \label{tab:correlation_classesSEM}
  \reproducibilityNote{
        - Run classes.R.
        - The table is printed to the console.
  }
\end{table*}

\subsection{Relation to Informational Fusion}\label{app:information-fusion}

Here, we discuss the relation between CETL and another information-theoretic metric for morphological paradigms, Informational Fusion \citep{rathi-etal-2021-information}.
The Informational Fusion of a pair of features (e.g., 2nd person \& plural) is defined as the cross-entropy that a seq2seq model trained to produce all paradigm cells not involving this pair experiences on predicting the cells involving this pair.
Like CETL, but unlike the measures of \cite{complexity_i_e_def, cotterell2019complexity, wu-etal-2019-morphological}, Informational Fusion is defined even on a single paradigm (a single pronoun paradigm, or an affix set applicable across verbs).
Informational Fusion is related to CETL in that both measures quantify the difficulty of learning a paradigm; the difference being that CETL assumes that the whole paradigm is learned progressively, whereas Informational Fusion assumes that all forms excluding a specific feature pair are learned to convergence before encountering that pair.
One important difference between the two measures is that Informational Fusion requires retraining seq2seq models for each feature combination. Given the large number of languages and counterfactual  paradigms, computing Informational Fusion would not have been feasible in our study.

\section{Additional Results} \label{app:detailed_results}

\subsection{Efficiency Plots and Significance Tests}
\label{app:add_res:EFF} 
\begin{figure*}
    \centering
    \begin{subfigure}{\textwidth}
        \includegraphics[width=0.9\linewidth]{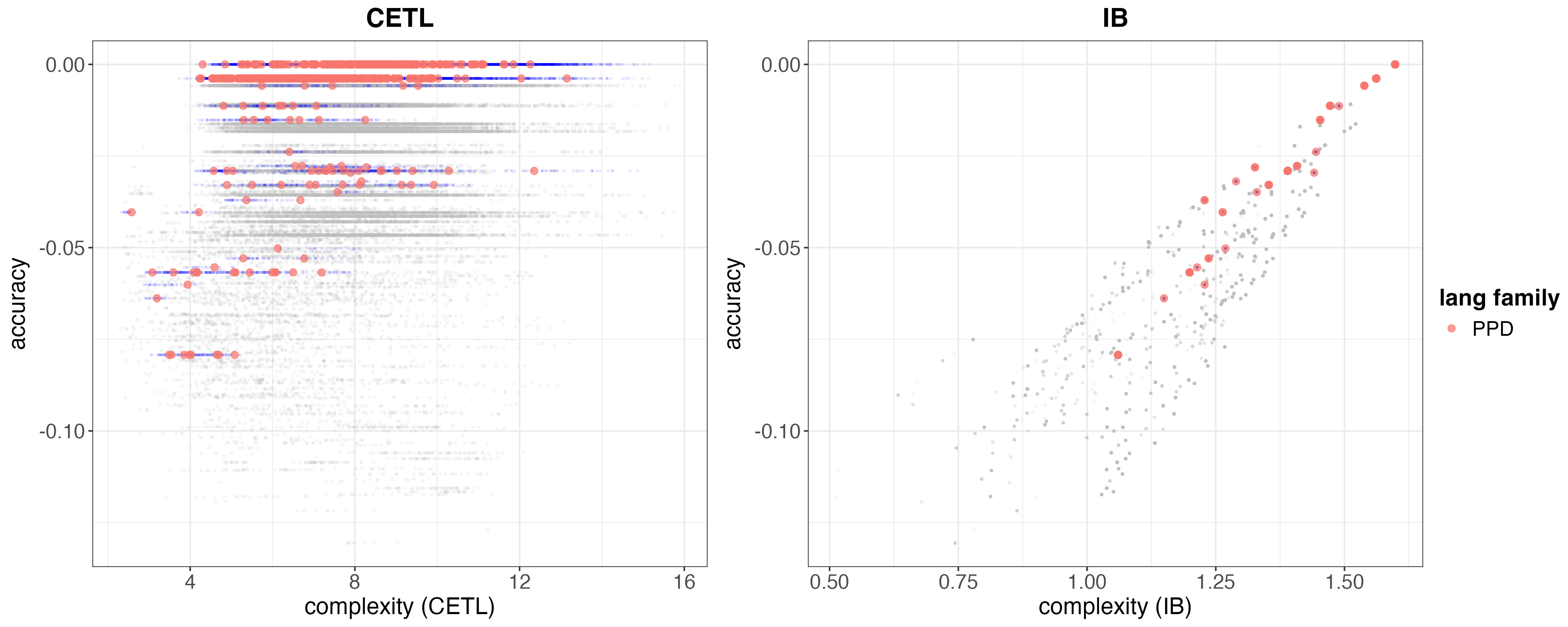}
        \caption{PPD}
    \end{subfigure}
    \begin{subfigure}{\textwidth}
        \includegraphics[width=0.9\linewidth]{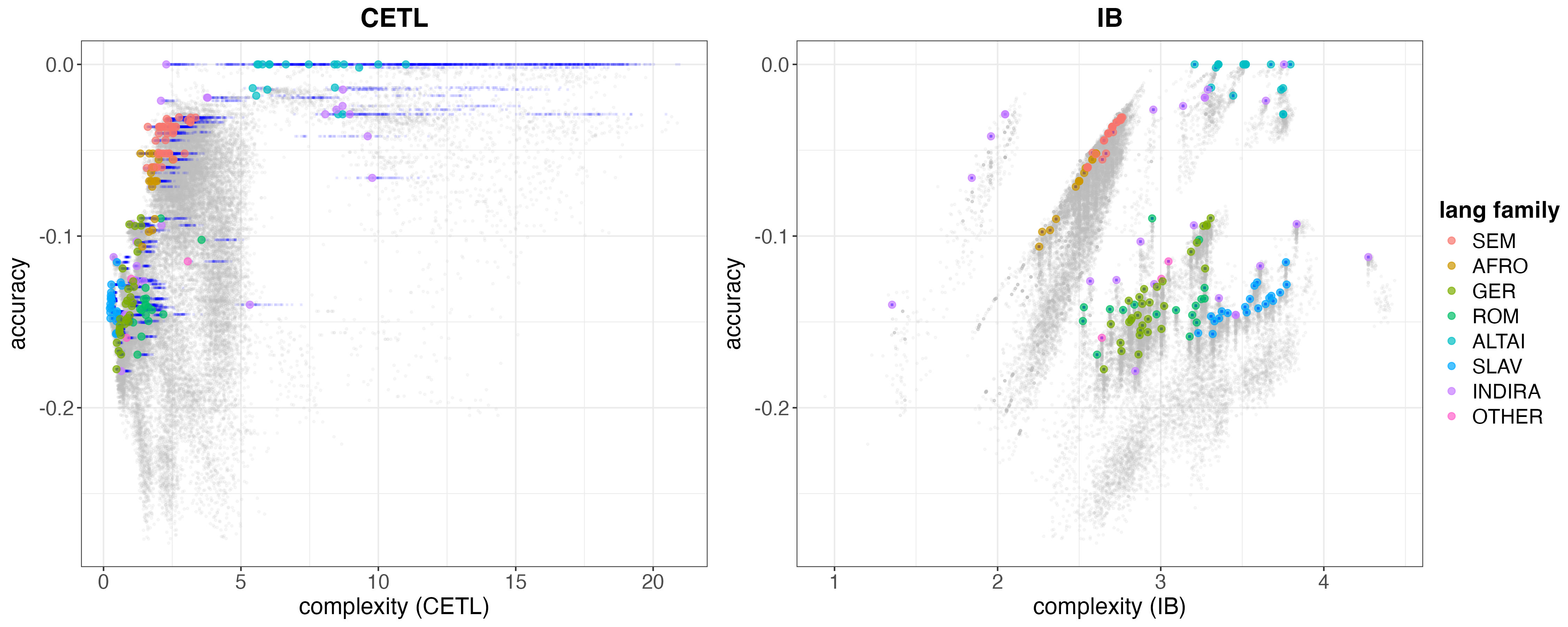}
        \caption{PRON}
    \end{subfigure}
    \begin{subfigure}{\textwidth}
        \includegraphics[width=0.9\linewidth]{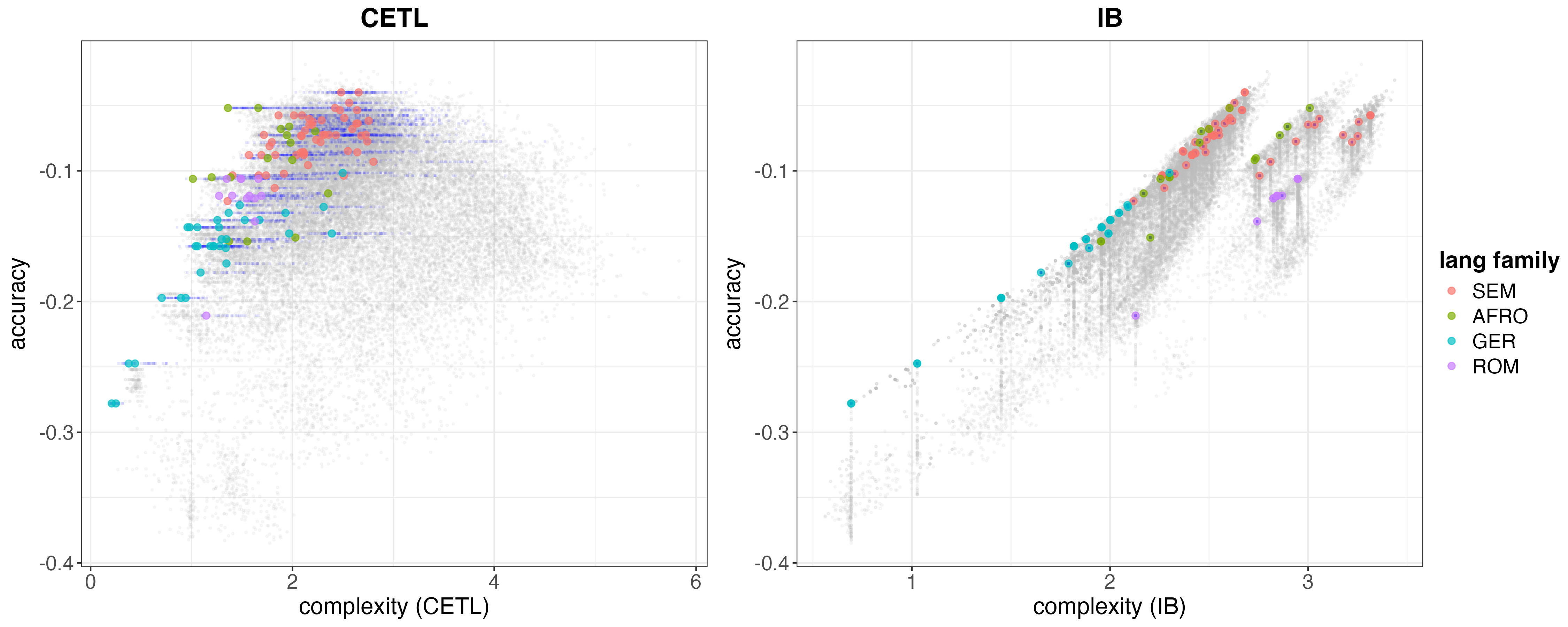}
        \caption{VERB}
    \end{subfigure}
    \caption{Efficiency of permutations: Accuracy plotted against complexity measures for the CETL (left) and original IB model (right). Note that the results in the left column are also shown in \autoref{fig:efficiency}.}
    \label{fig:EFF}
    \reproducibilityNote{
        - Run efficiency_plots.R.
        - The plot is in <OUTPUT_DIR>/EFF_PPD(_PERM_SHUF), <OUTPUT_DIR>/EFF_PRON(_PERM_SHUF), <OUTPUT_DIR>/EFF_VERB(_PERM_SHUF) .
        - The output directory can be set in the first lines (OUTPUT_DIR <- "[...]")
  }
\end{figure*}

\autoref{fig:EFF} shows the efficiency of factual vs. counterfactual paradigms for the CETL vs. the IB model for all three domains.

We performed two-sided one-sample Student's t-tests \citep{ttest} comparing the distribution of relative CETL and relative accuracy to 0 (the relative CETL and accuracy of the baseline distribution) for each domain. \autoref{tab:ttests} lists the results.

\begin{table}[t]
\small
\setlength{\tabcolsep}{3pt}
    \centering
    \begin{tabular}{cc||rr|rr}
     && mean base & mean perms & t-value & p-value \\\hline\hline
    \multirow{2}{*}{\textsc{pron}}&cetl &   7.38  &7.98 &  -8.5632 & \tgr{< 0.001}  \\
    &acc & 0.0075 &0.0134  & -9.5378 &  \tgr{< 0.001} \\
    \hline
    \multirow{2}{*}{\textsc{pr}}&cetl &  2.49   & 3.18  &  -3.6111 &  \tgr{< 0.001} \\
    &acc & 0.0854  & 0.1024 & -4.0620 & \tgr{< 0.001} \\
    \hline
    \multirow{2}{*}{\textsc{verb}}&cetl & 1.79  &  2.49  &  -12.5130 & \tgr{< 0.001}  \\
    &acc & 0.1092  &0.1206  & -2.4713 & \tgr{0.015}\\
\end{tabular}\\
\caption{Statistics for complexity and accuracy of real and counterfactual variants in the three domains.}
    \label{tab:ttests}
  \reproducibilityNote{
        - Run significancetests.R.
        - The table is printed to the console.
        Mean values need to be printed separately from permshuf_PPD_CE_ttest, permshuf_PPD_ACC_ttest etc. 
    }
\end{table}

\begin{table*}[h]
    \centering
    \small
\begin{tabular}{l|r|l|l}
Family & type & description & languages
\\\hline\hline
    AfroSem & Type1 & 4 tenses (3PC, 1SC) &\tgray{\textit{Akkadian:}} 
    Assyrian,
Babylonian,
Sargonic\\\cline{2-4}
     & Type2 & 4 tenses (2PC, 2SC) &
     \tgray{\textit{Ethiopic:}} Amharic,
Geez,
Tigrinya,
Muher\\\cline{2-4}
     & Type3 & 3 tenses (2PC, 1SC) &Proto-Semitic (vers1 + vers2),
\tgray{\textit{South Arabian:}} Jibbali,
Mehri,
Soqotri,\\&&&
\tgray{\textit{Ethiopic:}} Tigre,
\tgray{\textit{Berber:}} General Berber,
\tgray{\textit{Cushitic:}} Beja (PC + SC),\\&&&
\tgray{\textit{Chadic:}} Hausa,
\tgray{\textit{Omotic:}} Yemsa,\\\cline{2-4}
& Type4 & 2 tenses (1PC, 1SC) &
\tgray{\textit{Arabic:}} Classical Arabic,
informal Modern Standard Arabic,
Algerian Arabic,\\&&&
Andalusian Arabic,
Dafur Arabic,
Egyptian Arabic,
Hejazi Arabic,\\&&&
Jewish Baghdadi Arabic,
Muslim Iraqi Arabic,
Beiruti Levantine Arabic,\\&&&
Rural Levantine Arabic,
Jerusalemite Levantine Arabic,
Moroccan Arabic,\\&&&
Maltese,
Omani Arabic,
Saidi Arabic,
Tunesian Arabic,
Yemenite Arabic,\\&&&
\tgray{\textit{Canaanite:}} Biblical Hebrew,
Modern Hebrew,
Standard Phoenician,\\&&&
Punic Phoenician, 
\tgray{\textit{Ugaritic:}} Ugaritic,\\&&&
\tgray{\textit{Aramaic:}} Mlahso Central Neo-Aramaic,
Surayt Central Neo-Aramaic,\\&&&
Turoyo Central Neo-Aramaic,
Imperial Aramaic (vers1 + vers2)\\&&&
Jewish Babylonian,
Jewish Palestinian,
Classical Mandaic,\\&&&
Neo-Mandaic (vers1 + vers2)
Alqosh North-Western Neo-Aramaic,\\&&&
Western Neo-Aramaic,
Samaritan,
Syriac (vers1 + vers2)\\&&&
\tgray{\textit{Ancient South Arabian:}} Sabaic (vers1 + vers2)\\&&&
\tgray{\textit{Ancient Egyptian:}} Earlier Ancient Egyptian,
Middle Ancient Egyptian,
Coptic,\\&&&
\tgray{\textit{Berber:}} Ghadames Berber,
Tamasheq Berber,
Siwa Berber,\\&&&
\tgray{\textit{Cushitic:}} AfarPC (PC + SC),
SomaliPC (PC + SC),
\tgray{\textit{Chadic:}} Moloko,\\
    \hline\hline
    GER & Type1&&Old Dutch (strong + weak),
Middle Dutch (strong + weak),\\&&&
Modern Dutch (strong + weak),
Old English (strong + weak),\\&&&
Middle English (strong + weak),
Modern English (strong + weak),\\&&&
Old High German (strong + weak),
Middle High German (strong + weak),\\&&&
Standard High German (strong + weak),
Old Norse (strong + weak),\\&&&
Faroese (strong + weak),
Icelandic (strong + weak),\\&&&
Swedish (strong + weak),
Gothic (strong + weak),\\&&&
Old Saxon (strong + weak),
Proto-Germanic (strong + weak),
\\
    \hline\hline
    ROM & Type1&&Spanish,
Romanian,
Occitan,
Portuguese,
Latin,
Galician,
Italian,\\&&&
Franco-Provençal,
French,
North Corsican,
South Corsican,
Catalan\\
\end{tabular}

    \caption{Languages belonging to each Paradigm Class for \textsc{verb}. For some languages we used more than one paradigm indicated in brackets, as described in \aref{app:paradigm_representation}.}
    \label{tab:verb_paradigm_classes}
\end{table*}

\begin{table*}[h]
    \centering
    \small
       
\begin{tabular}{l|r|l|l}
Family & type & distinguishing description & languages
\\\hline\hline
    AfroSem & Type1 & &
    \tgray{\textit{Arabic:}} Moroccan Arabic, Classical Arabic, Egyptian Arabic, \\&&&
    North Levantine Arabic, South Levantine Arabic, \\&&&
    \tgray{\textit{Aramaic:}} 
    Babylonian Aramaic,
North-East Neo-Aramaic,\\&&&
Baxa Western Neo-Aramaic, 
Jubbadin Western Neo-Aramaic,\\&&&
Malula Western Neo-Aramaic,
Syriac (vers1 + vers2),\\&&&
Samaritan,
Mandaic,
    \tgray{\textit{Canaanite:}} Modern Hebrew,\\&&&
    Biblical Hebrew (vers1 + vers2), 
    Phoenician, Punic,\\&&&
    \tgray{\textit{Ethiopic:}} Amharic (vers1 + vers2), Geez (vers1 + vers2),\\&&&Tigre,
Tigrinya (vers1 + vers2 + vers3 + vers4), Muher, \\&&&Dahalik, Soddo, Argobba,
\tgray{\textit{Akkadian:}} Akkadian,\\&&&
    \tgray{\textit{South Arabian:}} Mehri,
Soqotri (vers1 + vers2),\\&&&
    \tgray{\textit{Ancient Egyptian:}} Coptic, Middle Ancient Egyptian,\\&&&
    \tgray{\textit{Berber:}} Ghadames Berber, Tashelhiyt Berber,\\&&&
    \tgray{\textit{Cushitic:}} Alaaba,
    Benijamer Beja (vers1 + vers2), \\&&&
    Agaw Bilin, Burunge, Somali, Tsamakko,\\&&&
    \tgray{\textit{Chadic:}} Hdi, Mubi, Hausa,
    \tgray{\textit{Omotic:}} Yemsa, Aari,\\&&&
 \tgray{\textit{Ugaritic:}} Ugaritic,
 \tgray{\textit{Proto-Semitic:}} ProtoSemitic (vers1 + vers2),
\\&&&
 \tgray{\textit{Ancient South Arabian:}} Razihi,
    \\\hline\hline
    GER & Type1&no formality& Middle Dutch,
Old Dutch,
English,
Middle English,\\&&&
Early Middle English,
Old English,
Faroese,\\&&&
Old Frisian,
Middle High German,
Low German,\\&&&
Middle Low German,
Gothic,
Icelandic,
Old Norse,\\&&&
Norwegian-Bokmal,
Norwegian-Nynorsk,
Proto-Germanic,\\&&&
Old Prussian,
Old Saxon\\\cline{2-4}
     & Type2&formality& Swedish,
Old High German,
Standard High German,\\&&&
Bavarian (vers1 + vers2),
Short Dutch,
Archaic English,\\&&&
Dutch,
Afrikaans\\
    \hline\hline
    ROM & Type1&3 cases, reflexive&Rumantsch,
Dalmatian\\\cline{2-4}
     & Type2&5 cases, reflexive&French,
Corsican,
Emilian,
Franco-Provençal,\\&&&
Occitan,
Sicilian\\\cline{2-4}
     & Type3&6 cases&Romanian\\\cline{2-4}
     & Type4&6 cases, formality, reflexive&Castillian Spanish,
Latin American Spanish,Catalan,\\&&&
Archaic Spanish,
Portuguese,
Italian,
Galician\\\cline{2-4}
     & Type5&6 cases, neutral gender, reflexive&Latin\\\hline\hline
    SLAV & Type1 & 6 cases + short vers, animation& Czech,
Kashubian (vers1 + vers2),
Polish, Slovak,\\&&&
Serbo-Croatian (vers1 + vers2),
Slovene (vers1 + vers2),\\&&&
Lower Sorbian,
Upper Sorbian\\\cline{2-4}
     & Type2 &6 cases& Belorussian,
Latvian,
Lithuanian,
Old Church Slavonic,\\&&&
Proto-Slavic,
Russian,
Ukrainian\\\cline{2-4}
     & Type3 &4 cases + short vers& Bulgarian,
Macedonian
\\\hline\hline
    ALTAI & Type Mong &incl 1st, formality, 9 cases& Mongol (vers1 + vers2),\\\cline{2-4}
     & Type Tung1 &incl 1st, 9 cases& Udihe\\\cline{2-4}
     & Type Tung2 &incl 1st, 5 cases& Manchu\\\cline{2-4}
     & Type Turk1 &7 cases& Chuvash,
Southern Altai,
Tuvan\\\cline{2-4}
     & Type Turk2 &7 cases, formality& Kazakh\\\cline{2-4}
     & Type Turk3 &6 cases& Turkmen,
Tartar,
Crimean,
Bashkir,
Azeri\\\cline{2-4}
     & Type Turk4 &6 cases, formality& Uzbek,
Uyghur,
Turkish,
Kyrgyz\\\cline{2-4}
     & Type Turk5 &11 cases& Old Turkic,
Proto-Turkic\\\hline\hline
INDOIRAN& Type1 &&Ossetian,
Sorani Kurdish,
Hewleri Kurdish,
Kurmanji Kurdish\\\cline{2-4}
    & Type2 &&Urdu (vers1 + vers2),
Punjabi,
Kashmiri\\\cline{2-4}
    & Type3 &&Iranian Persian / Farsi,
Afghani Persian / Dari,
Pashto\\\cline{2-4}
    & Type4 &&Assamese\\\cline{2-4}
    & Type5 &&Bengali\\\cline{2-4}
    & Type6 &&Gujurati\\\cline{2-4}
    & Type7 &&Sindhi,
Gilaki\\\cline{2-4}
    & Type8 &&Proto-Indo-European, Sanskrit\\\hline\hline
OTHER& Type1 &Armenian&Eastern Armenian, Classical Armenian\\\cline{2-4}
    & Type2 &Georgian&Georgian\\\cline{2-4}
    & Type3 &Circassian&Kabardian Circassian, Adyghe Circassian\\\cline{2-4}
    & Type4 &Albanian&Albanian\\\cline{2-4}
    & Type5 &Greek&Modern Greek, Ancient Greek\\
\end{tabular}
    \caption{Languages belonging to each Paradigm Class for \textsc{pron}. For some languages we used more than one paradigm indicated in brackets, as described in \aref{app:paradigm_representation}. 
    }
    \label{tab:pron_paradigm_classes}
\end{table*}

\subsection{Naturalness Plots}
\label{app:add_res:NAT}

\begin{figure*}[hbt]
\centering
    \includegraphics[width=0.9\linewidth]{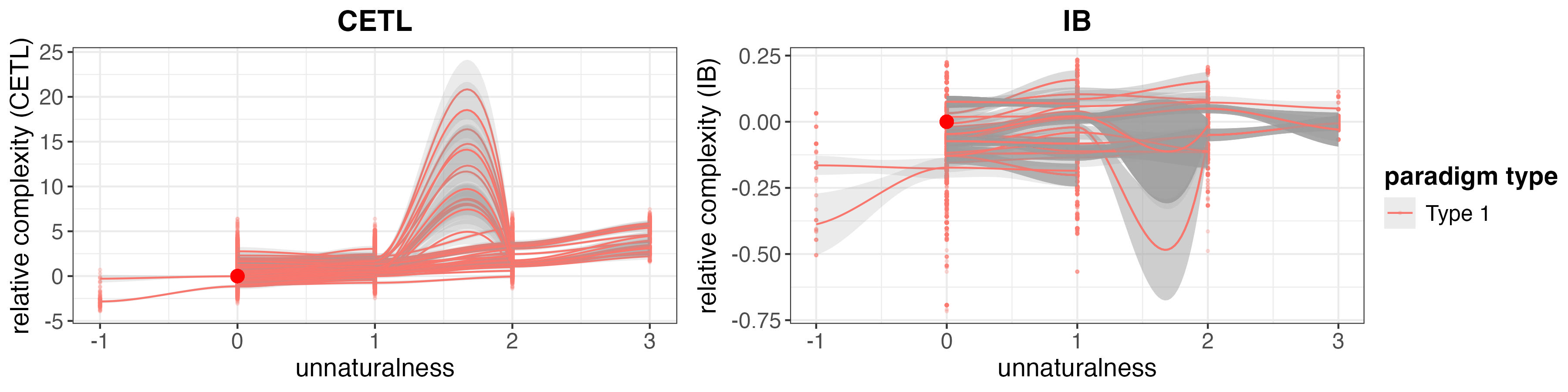}
    \caption{Complexity of Permutations of Different Naturalness on \textsc{ppd}: Complexity measures plotted against naturalness for the PPD data set in our CETL model (left) and the original IB model (right). Each line represents a separate language. 
    }
    \label{app:NAT_plots:PRON}
    \reproducibilityNote{
        - Run naturalness_plots.R.
        - The plot is in <OUTPUT_DIR>/NAT_PPD_langs.pdf.
        - The output directory can be set in the first lines (OUTPUT_DIR <- "[...]")
  }
\end{figure*}

\begin{figure*}[hbt]
\renewcommand{\thesubfigure}{\roman{subfigure}}
\centering
\begin{subfigure}{\textwidth}
\caption{Germanic}
\centering
    \includegraphics[width=0.9\linewidth]{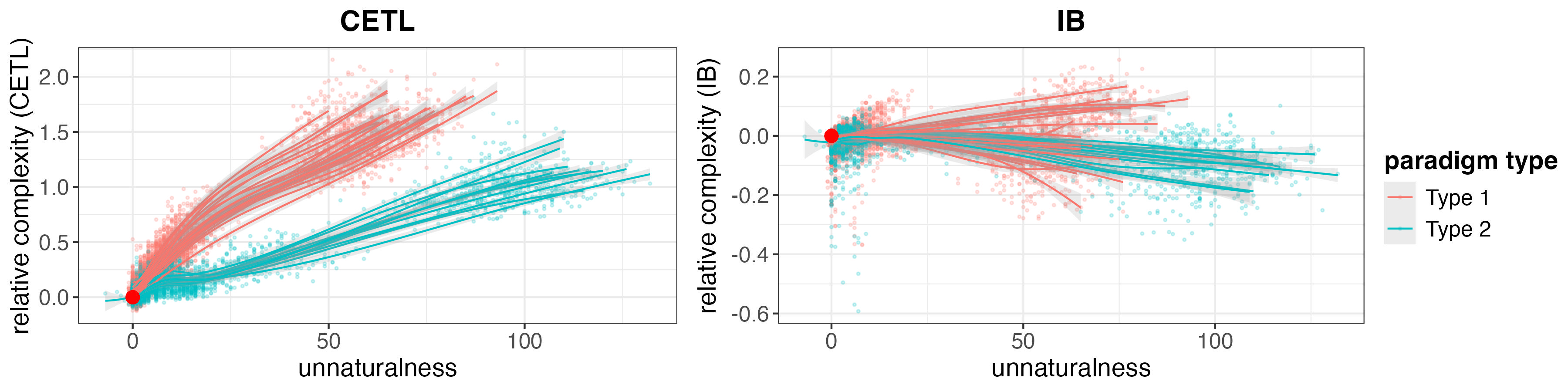}
\end{subfigure}
    \begin{subfigure}{\textwidth}
\caption{Romance}
\centering
    \includegraphics[width=0.9\linewidth]{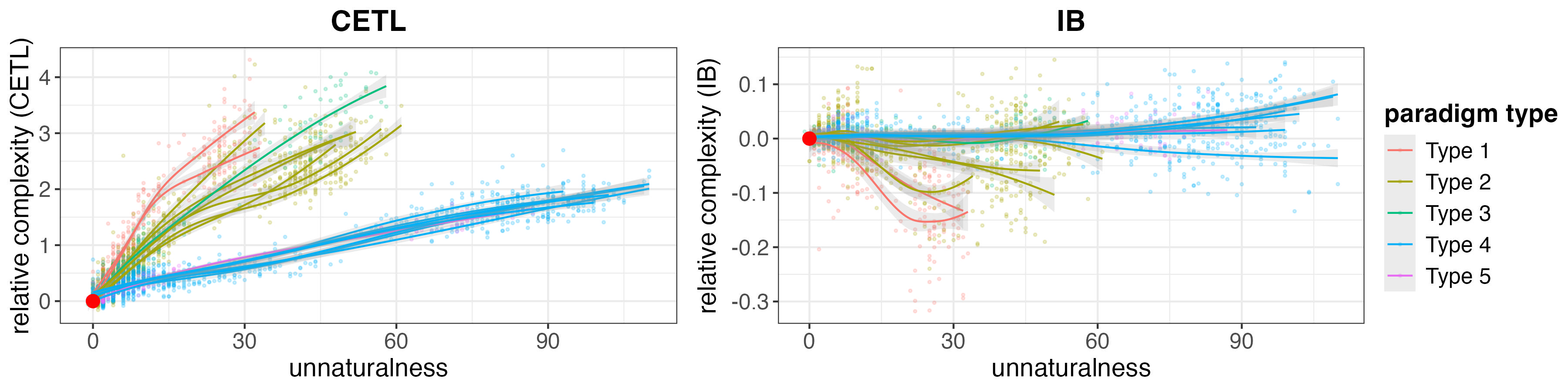}
\end{subfigure}
     \begin{subfigure}{\textwidth}
\caption{Slavic}
\centering
    \includegraphics[width=0.9\linewidth]{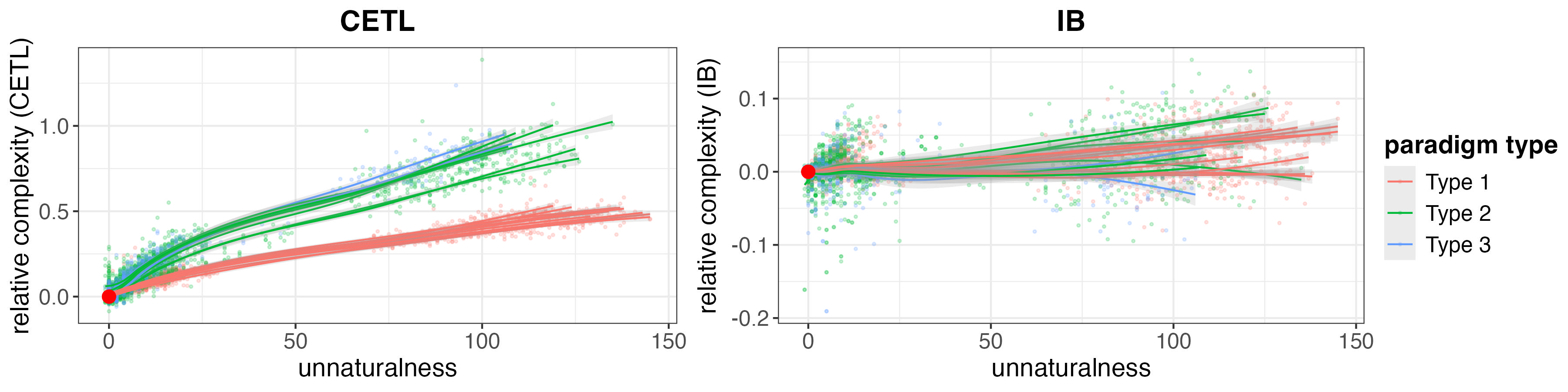}
    \end{subfigure}
    
    \caption{Complexity of Permutations of Different Naturalness on \textsc{pron} (part I): Complexity measures plotted against naturalness for pronouns in our CETL model (left) and the original IB model (right). Each line represents a separate language. Each color represents a different paradigm type.}
    \label{app:NAT_plots:PR_EURO}
    
\reproducibilityNote{
        - Run naturalness_plots.R.
        - The plot is in <OUTPUT_DIR>/NAT_PRON_Romanic_langs.pdf, NAT_PRON_Slavic_langs.pdf,  NAT_PRON_Germanic_langs.pdf.
        - The output directory can be set in the first lines (OUTPUT_DIR <- "[...]")
  }

\end{figure*}

\begin{figure*}[hbt]
\renewcommand{\thesubfigure}{\roman{subfigure}}
\centering
\begin{subfigure}{\textwidth}
\caption{Afroasiatic}
\centering
    \includegraphics[width=0.9\linewidth]{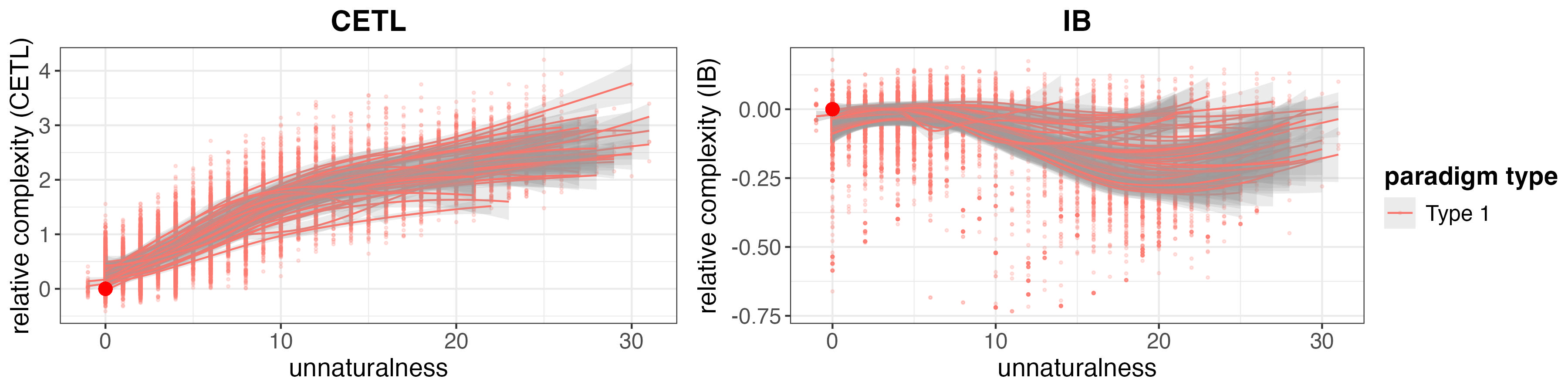}
\end{subfigure}
\begin{subfigure}{\textwidth}
\caption{Altaic (Mongolian, Tungusic, Turkic)}
\centering
    \includegraphics[width=0.9\linewidth]{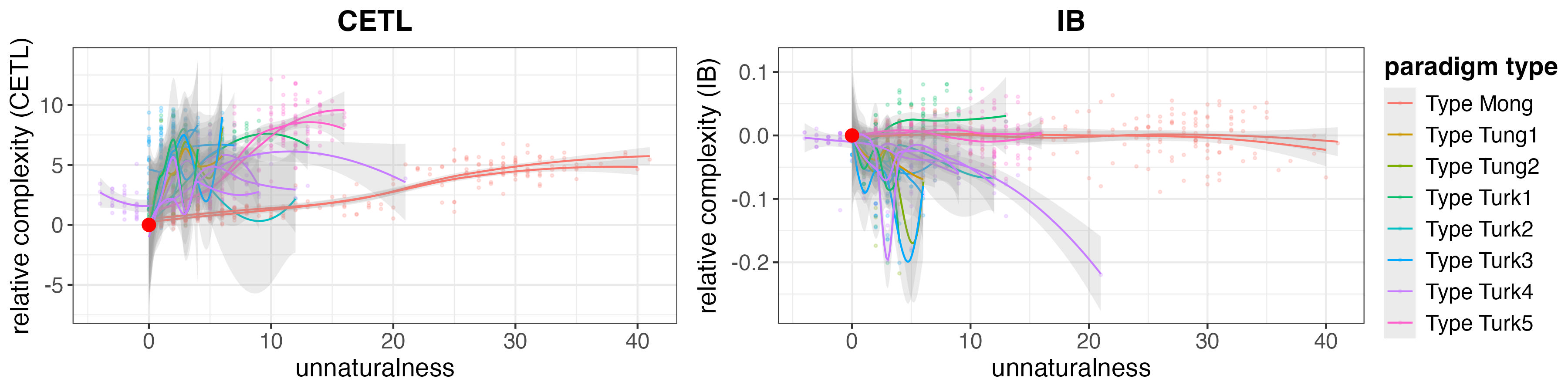}
\end{subfigure}
     \begin{subfigure}{\textwidth}
\caption{Indoiranian, Indoaryan}
\centering
    \includegraphics[width=0.9\linewidth]{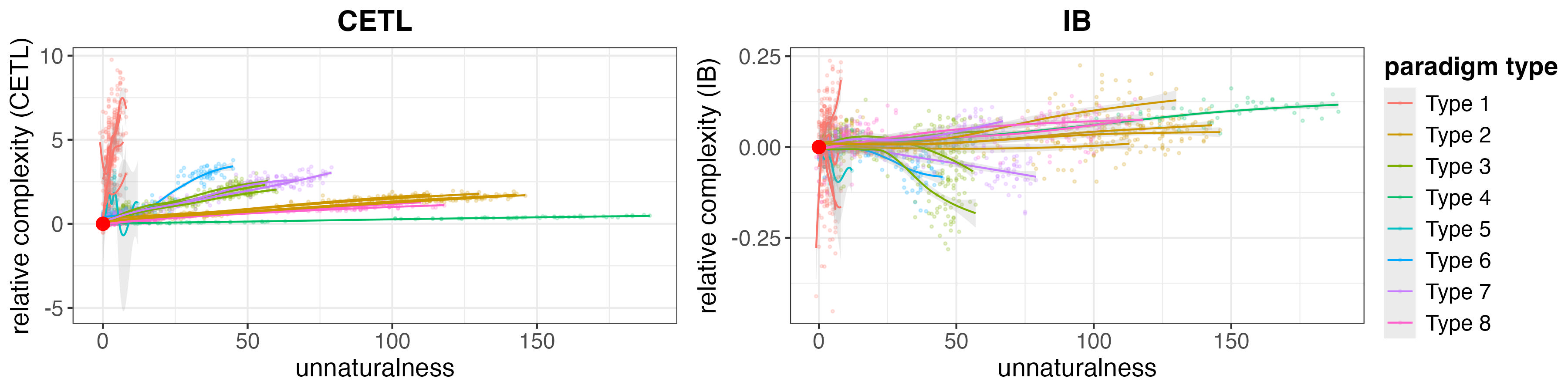}
    \end{subfigure}
     \begin{subfigure}{\textwidth}
\caption{Other (i.a. Greek, Circassian, Albanian)}
\centering
    \includegraphics[width=0.9\linewidth]{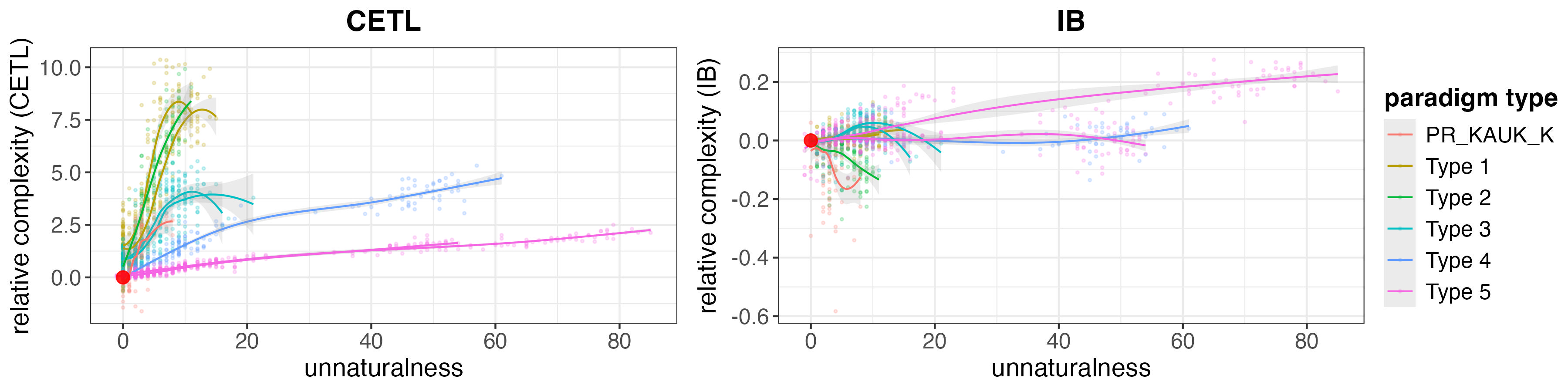}
    \end{subfigure}
    
    \caption{Complexity of Permutations of Different Naturalness on \textsc{pron} (part II): Complexity measures plotted against unnaturalness for pronouns in our CETL model (left) and the original IB model (right). Each line represents a separate language. Each color represents a different paradigm type.}
    \label{app:NAT_plots:PR_nonEURO}
    
\reproducibilityNote{
        - Run naturalness_plots.R.
        - The plot is in <OUTPUT_DIR>/NAT_PRON_AfroSem_langs.pdf, NAT_PRON_Altaic_langs.pdf, NAT_PRON_Indoiranian_langs.pdf,  NAT_PRON_Other_langs.pdf.
        - The output directory can be set in the first lines (OUTPUT_DIR <- "[...]")
  }

\end{figure*}

\begin{figure*}[hbt]
\renewcommand{\thesubfigure}{\roman{subfigure}}
\centering

\begin{subfigure}{\textwidth}
\caption{Afroasiatic}
\centering
    \includegraphics[width=0.8\linewidth]{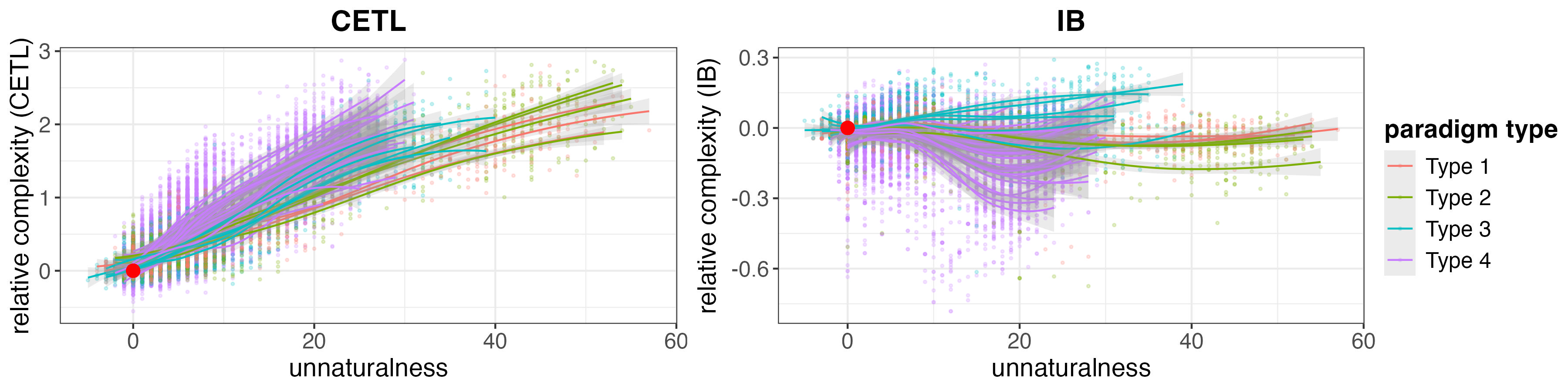}
\end{subfigure}

\begin{subfigure}{\textwidth}
\caption{Germanic}
\centering
    \includegraphics[width=0.8\linewidth]{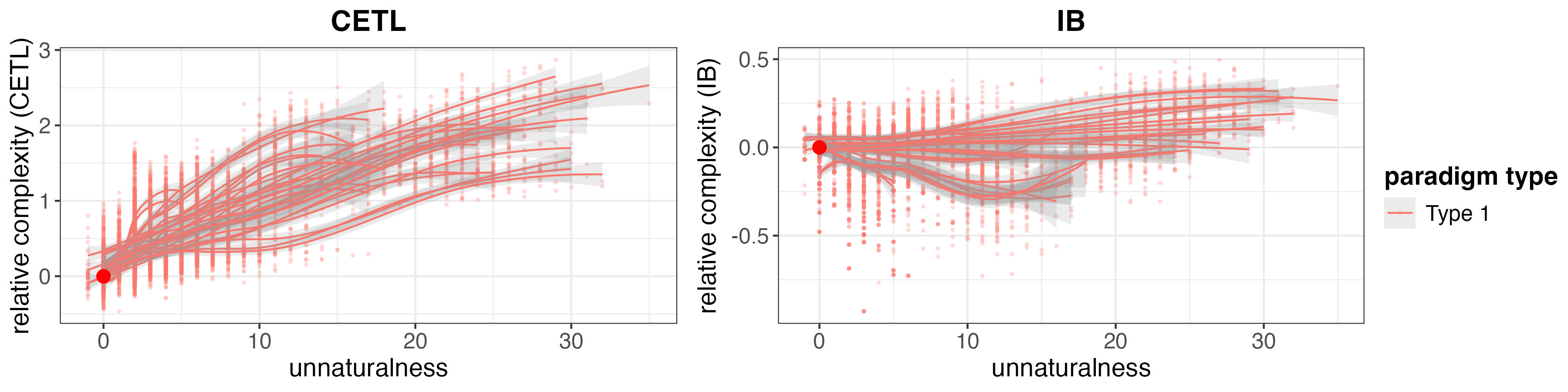}
\end{subfigure}

\begin{subfigure}{\textwidth}
\caption{Romance}
\centering
    \includegraphics[width=0.8\linewidth]{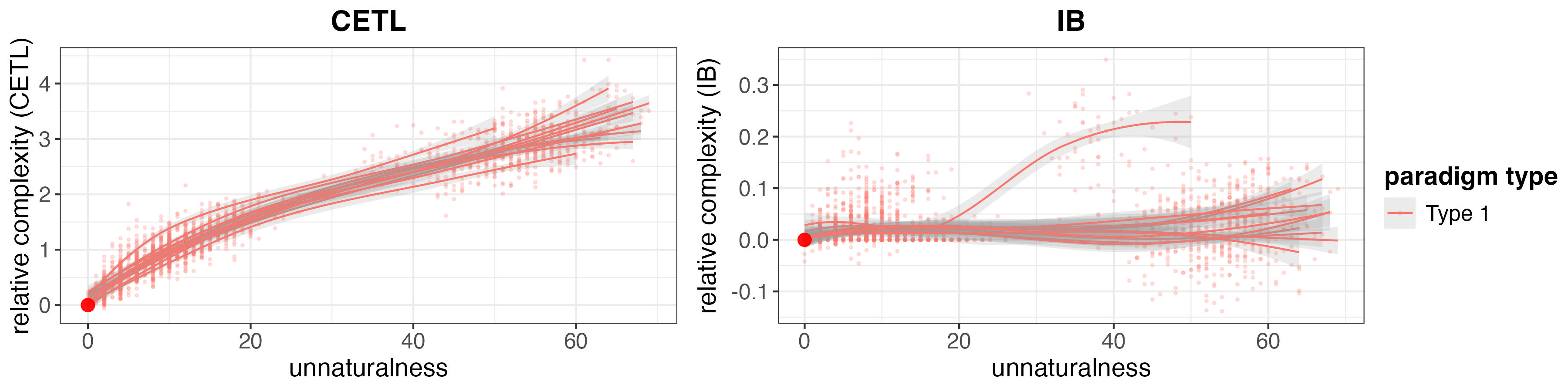}
\end{subfigure}

    \caption{Complexity of Permutations of Different Naturalness on \textsc{verb}: Complexity measures plotted against unnaturalness for verbs in our CETL model (left) and the original IB model (right). Each line represents a separate language. Each color represents a different paradigm type.}
    \label{app:NAT_plots:VERB}
\reproducibilityNote{
        - Run naturalness_plots.R.
        - The plot is in <OUTPUT_DIR>/NAT_VERB_SemAfro_langs.pdf, NAT_VERB_Germanic_langs.pdf, NAT_VERB_Romanic_langs.pdf.
        - The output directory can be set in the first lines (OUTPUT_DIR <- "[...]")
  }
  
\end{figure*}

The figures in this section show the correlation between complexity measure and naturalness for our CETL model vs.~the original IB model for \textsc{ppd} (\autoref{app:NAT_plots:PRON}), \textsc{pron} (\autoref{app:NAT_plots:PR_EURO} and \autoref{app:NAT_plots:PR_nonEURO}), and \textsc{verb} (\autoref{app:NAT_plots:VERB}), separated by language families and individual languages.

\subsection{Correlation Values}
\label{app:add_res:COR}

\begin{figure}
    \includegraphics[width=0.9\linewidth]{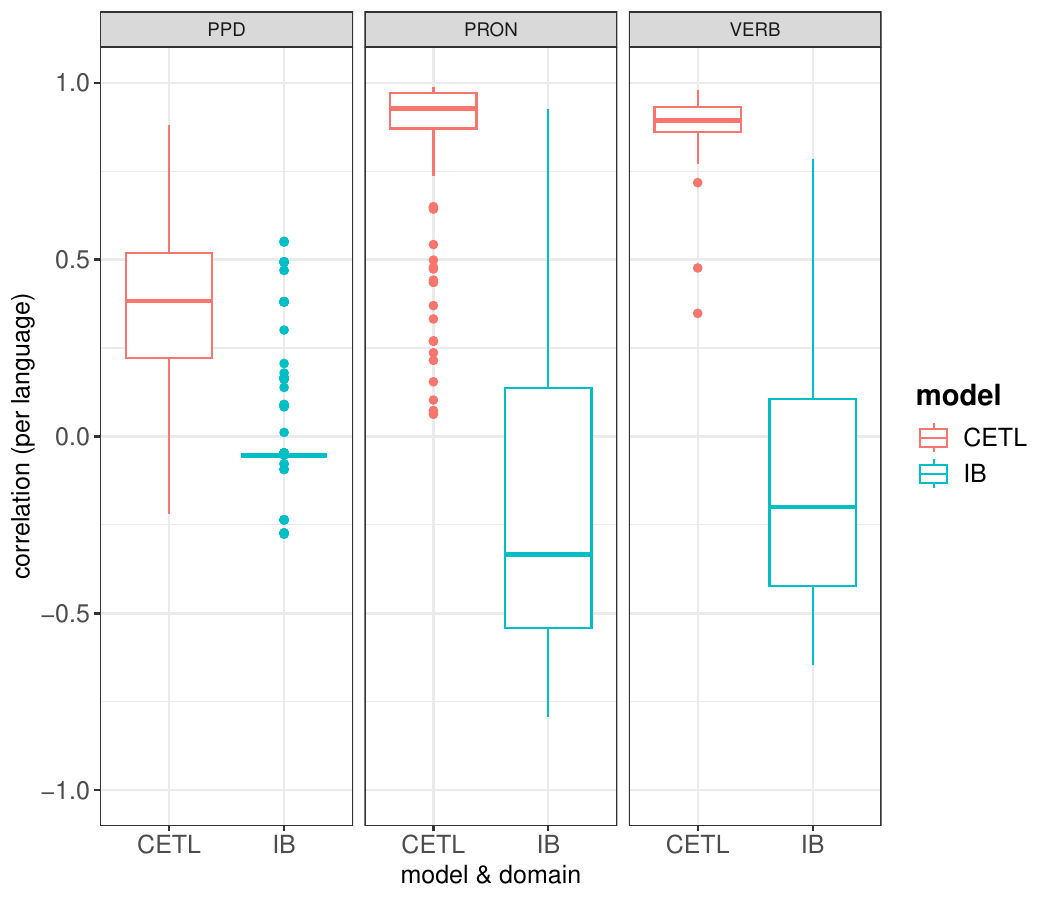}
\caption{Distribution of correlation values for all languages in the CETL model (red) vs.~the IB model (blue) across all language families and paradigm types in a domain.}
    \label{fig:app_COR_all}
\reproducibilityNote{
        - Run correlation_plots.R.
        - The plot is in <OUTPUT_DIR>/COR_all_box.pdf.
        - The output directory can be set in the first lines (OUTPUT_DIR <- "[...]")
  }
\end{figure}

\begin{figure}
    \centering
    \includegraphics[width=0.5\linewidth]{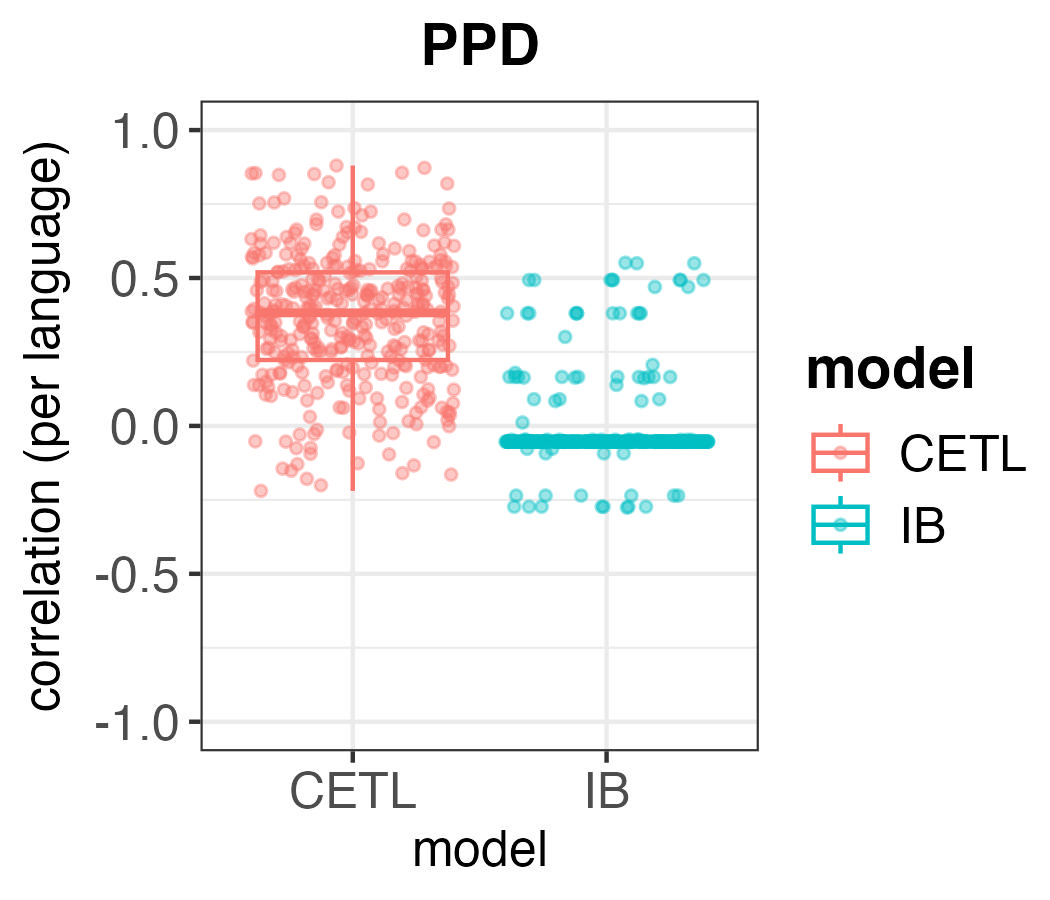}
    \caption{Distribution of correlation values for all languages in the CETL model (red) vs.~the IB model (blue) in a \textsc{ppd}.}
    \label{app:COR_plots:PRON}
\reproducibilityNote{
        - Run correlation_plots.R.
        - The plot is in <OUTPUT_DIR>/COR_PPD.pdf.
        - The output directory can be set in the first lines (OUTPUT_DIR <- "[...]")
  }
\end{figure}

\begin{figure}
\centering
    \includegraphics[width=0.8\linewidth]{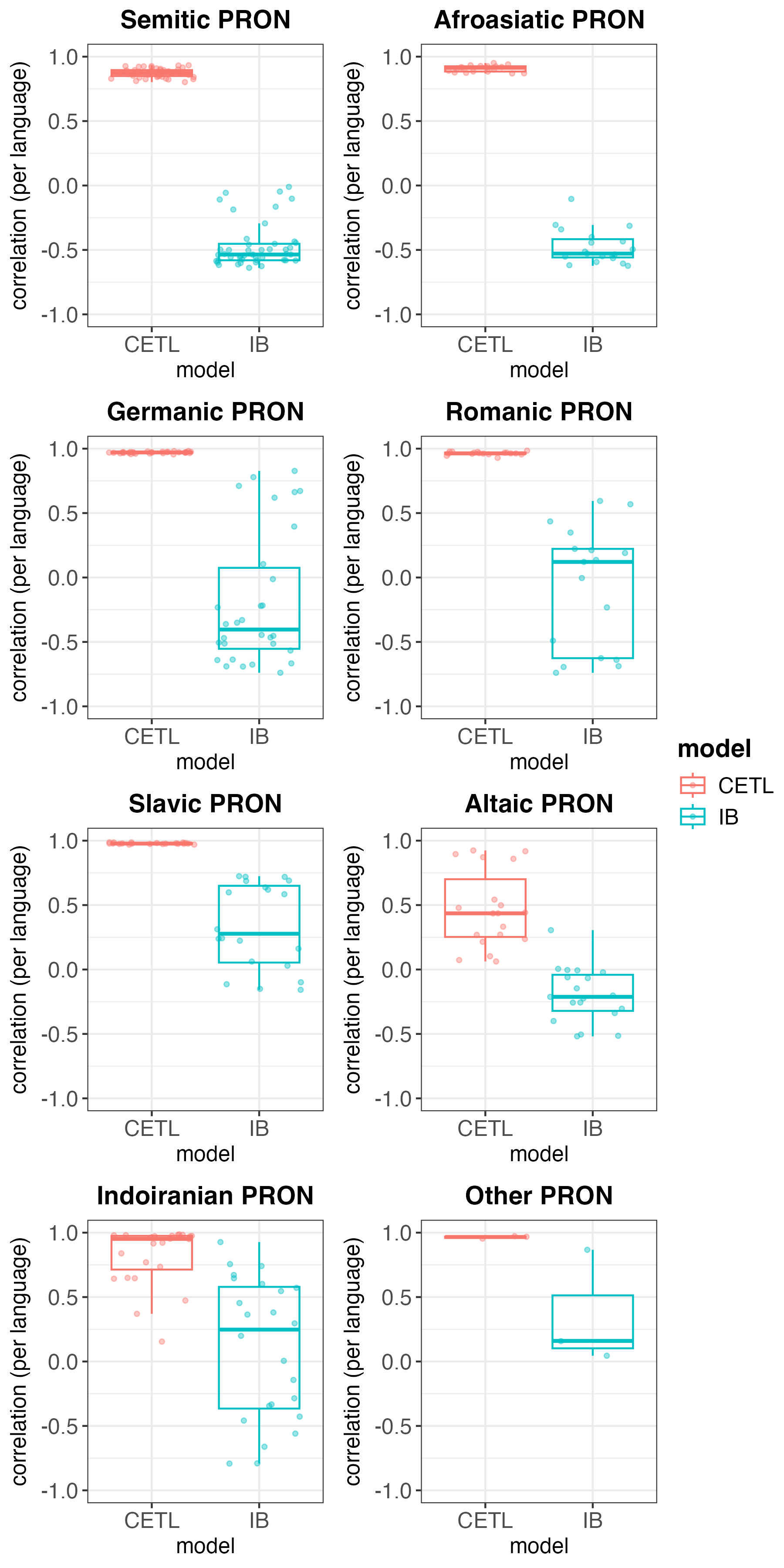}
    \caption{Distribution of correlation values for all languages in the CETL model (red) vs.~the IB model (blue) in \textsc{pron}, separated by language families.}
    \label{app:COR_plots:PR}
\reproducibilityNote{
        - Run correlation_plots.R.
        - The plot is in <OUTPUT_DIR>/COR_PRON.pdf.
        - The output directory can be set in the first lines (OUTPUT_DIR <- "[...]")
  }
\end{figure}

\begin{figure}
\centering
    \includegraphics[width=0.8\linewidth]{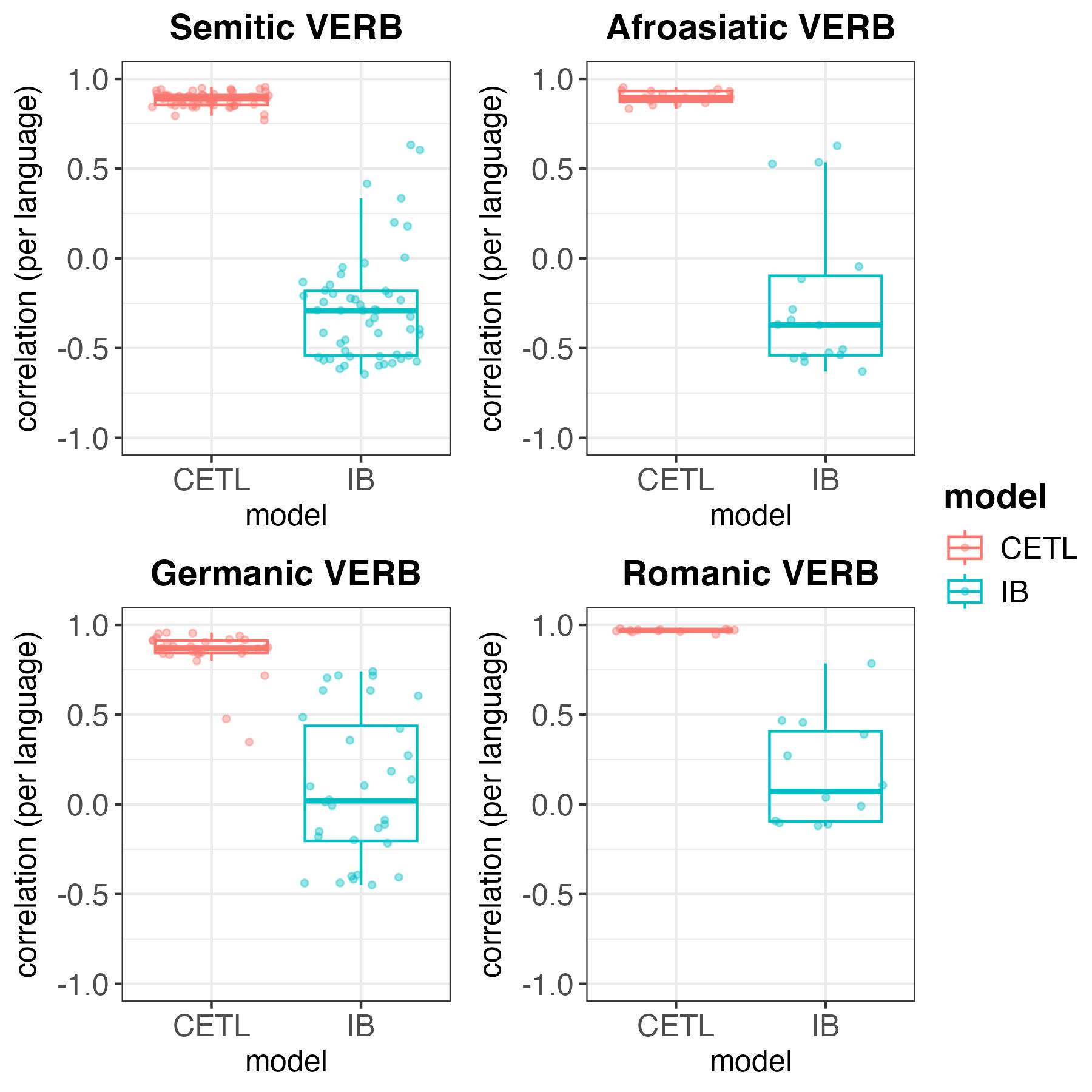}
    \caption{Distribution of correlation values for all languages in the CETL model (red) vs.~the IB model (blue) in \textsc{verb}, separated by language families.}
    \label{app:COR_plots:VERB}
\reproducibilityNote{
        - Run correlation_plots.R.
        - The plot is in <OUTPUT_DIR>/COR_VERB.pdf.
        - The output directory can be set in the first lines (OUTPUT_DIR <- "[...]")
  }
\end{figure}

\autoref{tab:correlation_table_APP} shows the averaged per-langauge correlation values for our CETL model vs.~the IB model, separated by domains.
\autoref{fig:app_COR_all} shows the variance of the per-language correlations for the two models in our three domains.
\autoref{app:COR_plots:PRON}, \autoref{app:COR_plots:PR} and \autoref{app:COR_plots:VERB} show the variance of the per-language correlation separated by language family for \textsc{ppd}, \textsc{pron} and \textsc{verb}, respectively.

\begin{table*}[h]
    \centering
    \small
    \begin{tabular}{c||cc|c}
         &\multicolumn{2}{c|}{correlation (avg)} &support\\
         & \texttt{CETL} & \texttt{IB} &  \\
         \hline\hline
       
       \textsc{ppd} & 
           0.36 &   -0.02   & \tdg{44112}  \\
       \textsc{pron} &
         0.82 &   -0.15 &\tdg{38623} \\
       \textsc{verb} &
         0.88  &  -0.12  &\tdg{32825}  \\
    \end{tabular}
    \caption{Correlation between complexity and unnaturalness, averaged over per-language correlations.}
    \label{tab:correlation_table_APP}
  \reproducibilityNote{
        - Run correlation_table.R.
        - The table is printed to the console.
  }
\end{table*}

\section{Data Sources}
\label{data_sources}

\subsection{Domain 1 (Verbs)}
\paragraph{Semitic Languages}
Proto-Semitic based on \citet{paradigm_proto_semitic, semiticlanguages_lipinski}. 

Classical Arabic based on \citet{paradigm_arabic_classical} and \citet{lehrbuch_arabisch},
Levantine Arabic based on \citet{paradigm_arabic_levantine} (Beiruti and Rural), \citet{jerusalemite} (Jerusalemite).
Nilo-Egyptian Arabic bases on \citet{paradigm_arabic_egyptian} (Egyptian), \citet{saidi} (Saidi), \citet{darfur} (Darfur).
Maghrebi Arabic bases on \citet{paradigm_arabic_moroccan} (Moroccan),
\url{https://en.wikipedia.org/wiki/Tunisian_Arabic_morphology} (Tunesian),
\url{https://en.wikipedia.org/wiki/Algerian_Arabic} (Algerian),
\citet{andalusian_book} (Andalusian),
\citet{maltese_book} (Maltese).
Mesopotamia Arabic based on \citet{iraqi_book_christian} and \citet{semlang_2_semitic_arabic_Meso}
.
Peninsular Arabic based on  \citet{hejazi_book} (Hejazi), \citet{omani} (Omani), \citet{yemeni} (Yemeni).

Hebrew based on 
\citet{paradigm_hebrew_biblical} (Biblical),
\citet{paradigm_hebrew_modern} (Modern),
Phoenician based on \citet{phoenician1} and \url{https://en.wikipedia.org/wiki/Phoenician_language} (Standard),
\url{https://en.wikipedia.org/wiki/Punic_language} (Punic).
Aramaic based on \citet{semlang_2_aramaic_imperial} (Imperial Aramaic),
 \citet{paradigm_aramaic_samaritan} (Samaritan),
\citet{paradigm_aramaic_modernwestern} (Siryon),
\citet{semlang_2_aramaic_jewish_palestinian} (Jewish Palestinian),
\citet{semlang_2_aramaic_classicalmandaic} (Classical Mandaic),
\citet{paradigm_aramaic_mandaic} (Neo-Mandaic),
\citet{paradigm_aramaic_syriac} (Syriac),
\citet{semlang_2_aramaic_jewish_babylonian} (Jewish Babylonian),
\citet{paradigm_aramaic_northwestern_neo} (Suret),
\citet{semlang_2_aramaic_centralneo} (\d{T}uroyo),
\citet{surayt_book} (Surayt),
\citet{semlang_2_aramaic_centralneo} (Mla\d{h}sô).
Ancient South Arabian based on \citet{paradigm_southarabian_sabaic}, \citet{semlang_2_southarabian_ancient} and \url{https://en.wikipedia.org/wiki/Sabaic}.
Ugaritic based on \citet{semlang_2_semitic_ugaritic}, \citet{paradigm_ugaritic}

Ge'ez based on \citet{paradigm_ethiopic_geez},
Tigre based on \citet{paradigm_ethiopic_tigre},
Tigrinya based on \citet{paradigm_ethiopic_tigrinya},
Amharic based on \citet{paradigm_ethiopic_amharic},
Muher based on \citet{paradigm_ethiopic_muher}, \citet{semlang_2_semitic_ethio_gurage}

Sargonic based on \citet{paradigm_old_akk}, \citet{semlang_2_akkadian_ebalaite} and \citet{paradigm_old_sargonic},
Old Assyrian based on \citet{paradigm_old_assyrian} and \citet{paradigm_introduction}, 
Old Babylonian based on \citet{paradigm_akkadian_babylonian}, \citet{semiticlanguages_lipinski} and \citet{semlang_2_akkadian_babylonian}

Mehri based on \citet{paradigm_southarabian_mehri},
Soqo\d{t}ri based on \citet{paradigm_southarabian_soqotri},
Jibbali based on \citet{jibbali_article}, \citet{semlang_2_southarabian}.

\paragraph{Non-Semitic Afroasiatic Languages}

Berber based on 
\url{https://de.wikipedia.org/wiki/Berbersprachen\#Verbalmorphologie} (Tamasheq),  \citet{lehrbuch_berber} (Tamazight), \citet{afro_berber} and \url{https://de.wikipedia.org/wiki/Siwi} (Siwa), \citet{paradigm_afroasiatic} (Standard Berber), \citet{beber_ghadames} (Ghadames).

Cushitic based on \citet{paradigm_afroasiatic} (Afar),
\citet{paradigm_afroasiatic} (Beja),
\citet{somali} (Somali).

Hausa based on \citet{afro_chadic},
Moloko based on \citet{moloko}. 
Yemsa based on \citet{afro_chadic}.

Ancient Egyptian based on 
\citet{afro_intro}, \citet{paradigm_afroasiatic} and \url{https://en.wiktionary.org/wiki/.k\#Egyptian} (Early);
\citet{afro_intro}, \citet{paradigm_afroasiatic} and \url{https://en.wiktionary.org/wiki/.k\#Egyptian} (Middle/Late);
\citet{afro_intro}, \citet{paradigm_afroasiatic} and \url{https://en.wiktionary.org/wiki/Appendix:Coptic_verbs}, \url{https://en.wiktionary.org/wiki/ⲙⲟϣⲓ\#Coptic} (Coptic).

\paragraph{Germanic Languages}
Faroese based on \citet{paradigm_faroese},
Icelandic based on \citet{paradigm_icelandic},
Swedish based on \citet{paradigm_swedish},
Old Norse based on \citet{paradigm_oldnorse},
Proto-Germanic based on \citet{protogermanic},
Gothic based on \citet{paradigm_gothic},
Standard High German based on \citet{paradigm_german},
Middle High German based on \citet{middle_german},
Old High German based on \citet{oldgerman},
Modern English based on \citet{paradigm_english},
Middle English based on \citet{middle_english},
Old English based on \citet{old_english},
Modern Dutch based on \citet{paradigm_dutch},
Middle Dutch based on \citet{middle_dutch},
Old Dutch based on \citet{old_middle_westgermanic}
Old Saxon based on \citet{old_saxon} and \citet{old_middle_westgermanic}

\paragraph{Romance Languages}

Catalan based on \citet{grammar_catalan_1}, \citet{grammar_catalan_2}.
Corsican based on \href{https://en.wiktionary.org/wiki/parlà#Corsican}{https://en.wiktionary.org/wiki/parlà\#Corsican}.
Franco-Provençal based on \href{https://en.wiktionary.org/wiki/chantar#Franco-Provençal}{https://en.wiktionary.org/wiki/chantar\#Franco-Provençal}.
French based on \citet{paradigm_french} and \href{https://fr.wiktionary.org/wiki/Conjugaison:français/regarder}{https://fr.wiktionary.org/wiki/Conjugaison:français\\/regarder}.
Galician based on \url{https://en.wiktionary.org/wiki/falar#Galician}.
Italian based on \citet{paradigm_italian}.
Latin based on \citet{paradigm_latin} and \url{https://verbix.com/webverbix/go.php?&D1=9&T1=canto}.
Occitan based on \citet{paradigm_occitan}.
Portuguese based on \citet{paradigm_portuguese}.
Romanian based on \citet{paradigm_romanian}.
Spanish based on \citet{paradigm_spanish}.

\subsection{Domain 2 (Pronouns)}

\paragraph{Semitic and Afro-Asiatic Languages}

Berber, Chadic, Omotic and Ancient Egyptian based on \citet{paradigm_afroasiatic},
Cushitic based on \citet{semlang_2_cush_omot} (Afar, Oromo, Sidaama) and \citet{paradigm_afroasiatic} (Alaaba, Beja, Bilin, Burunge, Somali, Tsamakko).

Akkadian based on \citet{paradigm_akkadian_babylonian}.

Ge'ez based on \citet{paradigm_ethiopic_geez},
Tigre based on \citet{paradigm_ethiopic_tigre},
Tigrinya based on \citet{paradigm_ethiopic_tigrinya},
Amharic based on \citet{paradigm_ethiopic_amharic},
Muher based on \citet{paradigm_ethiopic_muher}.

Argobba based on \url{https://en.wikipedia.org/wiki/Argobba_language},
Dahalik based on \citet{dahalik},
Razihi based on \citet{razihi}, 
Soddo based on \url{https://en.wikipedia.org/wiki/Soddo_language}.

Mehri based on \citet{paradigm_southarabian_mehri},
Soqo\d{t}ri based on \citet{paradigm_southarabian_soqotri}.

Classical Arabic based on \citet{paradigm_arabic_classical},
Levantine Arabic based on \citet{paradigm_arabic_levantine},
Moroccan Arabic based on \citet{paradigm_arabic_moroccan},
Egyptian Arabic based on \citet{paradigm_arabic_egyptian}.

Ugaritic based on \citet{paradigm_ugaritic}.

Biblical Hebrew based on \citet{paradigm_hebrew_biblical},
Modern Hebrew based on \citet{paradigm_hebrew_modern}.
Phoenician and Punic based on \citet{phoenician_book},  \citet{paradigm_canaanite}, \url{https://en.wikipedia.org/wiki/Punic_language#Personal_pronoun} and \url{https://en.wikipedia.org/wiki/Phoenician_language#Nominal_morphology}.

Aramaic based on \citet{paradigm_aramaic_samaritan} (Samaritan),
\citet{paradigm_aramaic_mandaic} (Mandaic),
\citet{paradigm_aramaic_syriac} (Syriac),
\citet{semlang_2_aramaic_jewish_babylonian} and \citet{jewish_babylonian_2} (Jewish Babylonian),
\citet{paradigm_aramaic_modernwestern} (Modern Western),
\citet{paradigm_aramaic_northwestern_neo} (Modern North-Eastern),
\citet{semlang_2_aramaic_centralneo} (Modern Central).

Proto-Semitic based on \url{https://en.wikipedia.org/wiki/Proto-Semitic_language#Pronouns} and \citet{proto_book}.

\paragraph{Germanic Languages}

High German based on 
\citet{oldgerman} (Old),
\citet{middle_german} (Middle),
\citet{paradigm_german} (Modern).
English based on \citet{paradigm_oldmiddleenglish}  and \url{https://en.wiktionary.org/wiki/git#Old_English} (Old),
\url{https://en.wikipedia.org/wiki/Middle_English} (Middle),\url{https://en.wikipedia.org/wiki/English_personal_pronouns} (Modern), \url{https://en.wikipedia.org/wiki/English_personal_pronouns#Complete_table} (Archaic).
Dutch based on \url{https://en.wiktionary.org/wiki/ik#Old_Dutch} (Old),
\citet{old_middle_westgermanic} (Middle),
\citet{paradigm_dutch} (Modern).
Low German based on \url{https://en.wiktionary.org/wiki/ik#Old_Saxon} (Old), \url{https://en.wiktionary.org/wiki/ik#Middle_Low_German} (Middle), \url{https://de.wikipedia.org/wiki/Niederdeutsche_Sprache#Pronomen} (Modern).
Bavarian based on \url{https://en.wikipedia.org/wiki/Bavarian_language#Pronouns}.
Afrikaans based on \citet{paradigm_afrikaans}.
Faroese on \citet{paradigm_faroese}.
Frisian based on \url{https://en.wiktionary.org/wiki/ik#Old_Frisian} (Old), \citet{paradigm_frisian} (Modern).
Gothic based on \citet{paradigm_gothic}.
Old Norse based on \citet{paradigm_oldnorse}.
Icelandic based on \url{https://en.wikipedia.org/wiki/Icelandic_grammar#Pronouns}.
Swedish based on \citet{paradigm_swedish}.
Norwegian based on \citet{paradigm_norwegian}.
Proto-Germanic based on \url{https://en.wiktionary.org/wiki/Reconstruction:Proto-Germanic/ek#Proto-Germanic}.
Old Prussian based on \url{https://en.wikibooks.org/wiki/Prussian/Personal_Pronouns_Chart}.

\paragraph{Romance Languages}

Dalmatian based on \url{https://en.wikipedia.org/wiki/Dalmatian_grammar#Pronouns}.
Rumantsch based on \url{https://de.wikipedia.org/wiki/Grammatik_des_Rumantsch_Grischun#Personalpronomen}.
Corsican based on \url{https://en.wiktionary.org/wiki/eiu#Corsican}.
Emilian based on \href{https://en.wiktionary.org/wiki/mè#Emilian}{https://en.wiktionary.org/wiki/mè\#Emilian}.
Franco-Provençal based on \href{https://en.wiktionary.org/wiki/o#Franco-Provençal}{https://en.wiktionary.org/wiki/o\#Franco-Provençal}.
French based on \citet{paradigm_french}.
Occitan based on \citet{paradigm_occitan}.
Sicilian based on \citet{grammar_sicilian} and \url{https://it.wikipedia.org/wiki/Lingua_siciliana#Pronomi}.
Latin based on \citet{paradigm_latin}.
Romanian based on \citet{paradigm_romanian}.
Catalan based on \citet{paradigm_catalan}, \citet{grammar_catalan_2}.
Galician based on  \url{https://en.wiktionary.org/wiki/eu#Galician}.
Italian based on \citet{paradigm_italian}.
Portuguese based on \citet{paradigm_portuguese}.
Spanish based on \citet{grammar_spanish}.

\paragraph{Slavic Languages}

Proto-Slavic based on \citet{paradigm_proto_slavic} and \url{https://en.wiktionary.org/wiki/Template:sla-decl-ppron}.
Belorussian based on \citet{paradigm_belorussian}.
Latvian based on \citet{latvian_grammar}.
Lithuanian based on \citet{lithuanian_grammar}.
Old Church Slavonic based on \citet{paradigm_oldchurchslavonic}.
Russian based on \citet{paradigm_russian}.
Ukrainian based on \citet{paradigm_ukrainian}.
Bulgarian based on \citet{paradigm_bulgarian}.
Macedonian based on \citet{paradigm_macedonian}.
Czech based on \citet{paradigm_czech}.
Kashubian based on \citet{paradigm_cassubian}.
Polish based on \citet{paradigm_polish}.
Serbo-Croatian based on \citet{paradigm_serbocroatian}.
Slovak based on \citet{paradigm_slovak}.
Slovene based on \citet{paradigm_slovene}.
Sorbian based on \citet{paradigm_sorbian}, 
\url{https://en.wikibooks.org/wiki/Lower_Sorbian/Grammar/Pronouns},
\url{https://en.wiktionary.org/wiki/ja#Lower_Sorbian} and
\url{https://en.wiktionary.org/wiki/ja#Upper_Sorbian}.

\paragraph{Altaic Languages}

Mongol based on 
\citet{grammar_mongol},  \url{https://en.wikipedia.org/wiki/Mongolian_language#Pronouns} and \url{https://en.wiktionary.org/wiki/Template:mn-personal_pronouns}.
Manchu based on \url{https://en.wikipedia.org/wiki/Manchu_language#Pronouns}
Udihe based on \url{http://www.tufs.ac.jp/ts/personal/kazama/shigen/18/Kazama.pdf}

Southern Altai based on \url{https://en.wikipedia.org/wiki/Altai_languages#Morphology_and_syntax}.
Chuvash  based on \citet{paradigm_chuvash}.
Tuvan  based on \url{https://en.wiktionary.org/wiki/Template:Tuvan_personal_pronouns}.
Kazakh  based on \citet{paradigm_kazakh_karakalpak}.
Azeri based on \citet{paradigm_azeri}.
Bashkir based on \citet{paradigm_tatar_bashkir} and \url{https://en.wikipedia.org/wiki/Bashkir_language#Declension_table}.
Crimean based on \citet{book_crimean}.
Tartar  based on \citet{paradigm_tatar_bashkir} and \citet{tartar_grammar}.
Turkmen based on  \citet{paradigm_turkmen}.
Kyrgyz  based on \citet{paradigm_kyrgyz}.
Turkish  based on \citet{paradigm_turkish}.
Uyghur  based on \citet{paradigm_uyghur}.
Uzbek  based on \citet{paradigm_uzbek}.
Old-Turkic \citet{grammar_old_turkish}.
Proto-Turkic based on 
\href{https://en.wiktionary.org/wiki/Reconstruction:Proto-Turkic/bẹ}{https://en.wiktionary.org/wiki/\\Reconstruction:Proto-Turkic/b\d{e}}, \href{https://en.wiktionary.org/wiki/Reconstruction:Proto-Turkic/sẹ}{https://en.wiktionary.org/wiki/\\Reconstruction:Proto-Turkic/s\d{e}}, \href{https://en.wiktionary.org/wiki/Reconstruction:Proto-Turkic/biŕ}{https://en.wiktionary.org/wiki/\\Reconstruction:Proto-Turkic/biŕ}, \href{https://en.wiktionary.org/wiki/Reconstruction:Proto-Turkic/siŕ}{https://en.wiktionary.org/wiki/\\Reconstruction:Proto-Turkic/siŕ} and  \href{https://en.wiktionary.org/wiki/Reconstruction:Proto-Turkic/ol}{https://en.wiktionary.org/wiki/Reconstruction:Proto-Turkic/ol}.

\paragraph{Indo-Iranian Languages} 
Sanskrit based on \citet{grammar_sanskrit}.
Urdu based on \citet{paradigm_urdu} and \citet{grammar_urdu}.
Assamese based on \citet{paradigm_assamese}.
Bengali based on \citet{paradigm_bengali} and \url{https://en.wikipedia.org/wiki/Bengali_grammar#Pronouns}.
Gujurati based on \citet{paradigm_gujurati}.
Sindhi based on \citet{paradigm_sindhi}.
Kashmiri based on \citet{paradigm_kashmiri} and \citet{grammar_kashmiri}.
Punjabi based on \citet{paradigm_punjabi}.

Gilaki based on \url{https://en.wikipedia.org/wiki/Gilaki_language#Pronouns}.
Pashto based on \citet{grammar_pashto}  and \citet{grammar_pashto_2}.
Persian based on \url{https://en.wiktionary.org/wiki/Template:prs-personal_pronouns} and \url{https://en.wiktionary.org/wiki/Template:fa-personal_pronouns}.
Kurdish based on
\url{https://en.wikipedia.org/wiki/Central_Kurdish#Grammar_and_Syntax} and 
\citet{paradigm_kurdish_kurmanji}.
Ossetian \url{https://en.wikipedia.org/wiki/Ossetian_language#Pronouns}

\paragraph{Other Languages}
Ancient Greek based on \citet{grammateion}.
Modern Greek \citet{grammar_greek_modern}
Albanian based on \citet{grammar_albanian}.
Classical Armenian based on \url{https://en.wiktionary.org/wiki/ես#Old_Armenian}, \url{https://en.wiktionary.org/wiki/դու#Old_Armenian} and \url{https://en.wiktionary.org/wiki/նա#Old_Armenian}.
Eastern Armenian based on \url{https://en.wikipedia.org/wiki/Template:Armenian_personal_pronoun_table}.
Circassian based on \url{https://en.wikipedia.org/wiki/Circassian_pronouns} (Adyghe)
and \url{https://en.wikipedia.org/wiki/Kabardian_grammar#Pronouns} (Kabardian).
Georgian based on \url{https://en.wikibooks.org/wiki/Georgian/Pronouns}.
Proto-Indo-European based on \url{https://en.wikipedia.org/wiki/Proto-Indo-European_pronouns#Personal_pronouns}.

\end{document}